\documentclass{article}

\usepackage{arxiv}
\renewcommand{\today}{}
\usepackage{tcolorbox}
\usepackage{amsmath}
\usepackage{wasysym} 
\usepackage[utf8]{inputenc}
\usepackage[T1]{fontenc}   
\usepackage{hyperref}
\hypersetup{
    colorlinks=true,
    urlcolor=blue,
    linkcolor=blue,
    citecolor=blue
}
\usepackage{xurl}
\usepackage{amsfonts}      
\usepackage{nicefrac}       
\usepackage{microtype}
\usepackage{graphicx}
\usepackage{subcaption}
\usepackage{booktabs}
\usepackage{pdflscape}
\usepackage[sort&compress, numbers]{natbib}
\usepackage{array}
\usepackage{float}
\usepackage{rotating}
\usepackage{xparse}
\usepackage{xstring}
\usepackage{placeins}
\usepackage{booktabs}
\usepackage{multirow}

\title{Natural Language Processing Psychometrics}

\author{
    Edoardo Sebastiano De Duro $^{1}$ \\
    \texttt{edoardo.deduro@unitn.it} \\
    \And
    Emma Franchino $^{1}$ \\
    \texttt{emma.franchino@unitn.it} \\
    \And
    Massimo Stella $^{1}$ \\
    \texttt{massimo.stella-1@unitn.it}
    \vspace{0.5em} \\
    \small{$^{1}$ CogNoscoLab, Department of Psychology and Cognitive Science, University of Trento, Italy}
}
    
\begin{document}

\maketitle

\begin{abstract}
Natural Language Processing (NLP) models predicting mental health outcomes rarely specify what they measure: contextual knowledge, emotional content, or syntactic structure. NLP Psychometrics treats psychological prediction from text as a psychometric problem, linking scores to interpretable linguistic evidence and testing beyond the training text format. Nine LLMs, conditioned on controlled personas (cognitive digital shadows), completed psychometric questionnaires with textual explanations per item. We extracted emotional profiles and syntactic-semantic structure via textual forma mentis networks, combined with personality and sociodemographic variables in ablated random forest (RF) regressors, using SHAP to identify which features drove performance and in which direction. Full RF models explained up to 70.8\% of variance in life satisfaction (SWLS), 55.7\% in depression (PHQ-9), and, for DASS-21, 68.5\% depression, 76.0\% anxiety, 72.4\% stress. Sociodemographics alone explained no meaningful variance in depression, anxiety, or stress, but did so for life satisfaction, where emotion features and income were the strongest predictors; neuroticism and network topology instead dominated depression and anxiety, reversing direction between them. Without retraining, RF models separated diaries from low- and high-score personas ($r$ up to 0.91) and, using only network/emotion features, classified clinical from control participants in real transcripts with up to 68\% accuracy. These results show the promise and limits of synthetic data: LLM personas can expose model biases, recover patterns consistent with clinical rumination, and support psychometric prediction from human text without a matched questionnaire, but cannot substitute for human validation. NLP Psychometrics makes these distinctions explicit, measurable, and testable through interpretable AI and network/emotional features.
\end{abstract}

\keywords{Natural Language Processing \and Psychometrics \and Mental Health \and Large Language Models \and Cognitive Networks}

\section{Introduction}

Measuring psychological phenomena depends on inner experience leaving quantifiable traces behind \citep{perinelli2026substantive, giovanelli2026self, goretzko2021exploratory}. Some of those are explicit, as in questionnaire ratings: Individuals consciously rate experiences and leave correlated numerical sequences, or item scores, as traces of their inner world \citep{goretzko2021exploratory, perinelli2026substantive}. Other traces are linguistic \citep{fatima2021dasentimental, carrillo2026llms, bilotta2024examining}, distributed across the words people choose, the emotions they express, and the concepts they connect, either explicitly or implicitly. In this view, language is not only a vehicle for communication \citep{fedorenko2024language, aitchison2012words} but a key trace for psychological measurement, whose structured knowledge can open the way to psychological measurements \citep{cutler2023deep, semeraro2025emoatlas}.

This premise is deeply connected to research on the mental lexicon \citep{aitchison2012words, vitevitch2014using, kenett2016structure}. In the modelling metaphor of the mental lexicon, language is reflected within cognition as a complex system of interconnected concepts or words \citep{stella2024cognitive}. The latter are not independent labels attached to experience, but structured cognitive representations embedded in networks of meaning \citep{aitchison2012words}. Psychological states may therefore leave traces not only in what words are used \citep{al2018absolute, taylor2025users}, but in how those words are arranged into conceptual structure \citep{semeraro2025emoatlas, vitevitch2014using, kenett2016structure}. For instance, past research found that recalling emotions sitting at different locations of a network of free associations, modelling associative memory, was predictive of individuals' psychometric levels of anxiety, stress and depression \citep{fatima2021dasentimental}. Similar approaches have established further connections between language use and psychological constructs of wellbeing, especially with the use of generative AI (GenAI; cf. \citep{liu2022detecting, de2013predicting} for review). These patterns invite novel network-based and AI-informed approaches to psychometrics, treating language as a measurable architecture of cognition \citep{semeraro2025emoatlas}, rather than as a mere container of symptoms expressed with words in isolation \citep{aitchison2012words}.

This paper develops this idea into a framework that we call \textit{Natural Language Processing Psychometrics} or \textit{NLP Psychometrics}. Encompassing a unique blend of network science \citep{semeraro2025emoatlas}, explainable AI \citep{franchino2026digital, carrillo2026llms} and cognitive science \citep{fatima2021dasentimental}, NLP Psychometrics studies how psychological constructs can be inferred, validated, and interpreted from language. Our approach builds on recent proposals for Text Psychometrics \cite{low2024speech, lowtext}, which argue that models assessing psychological constructs from text should be evaluated with the same concern for validity and reliability expected of traditional psychometric instruments. NLP Psychometrics extends this agenda by combining validated psychometric questionnaires with personal, language-based descriptions of individual items. The goal is not to wholly replace psychometric scales with opaque AI or NLP classifiers but rather to:  (1) model how psychometric variation becomes expressed in language in interpretable ways and (2) deploy methods that can transfer such psychometric scoring/linguistic mapping to data where only language is available.

For instance, we here show how NLP Psychometrics can be used to learn the correspondence between depression scores (from DASS-21 \citep{lovibond1995structure} and PHQ-9 \citep{kroenke2001phq}) in language-extended psychometric questionnaires to then classify depression levels in annotated clinical data where scoring is absent.

\subsection{Main challenges for NLP Psychometrics and the role of LLMs}

Crucially, language-based psychological assessment faces a core measurement problem \citep{low2024speech, lowtext}. A text can reflect a latent construct, but it can also reflect topic, genre, prompt structure, demographic background, or stylistic habit \citep{bilotta2024examining, taylor2025users, ardebili2026mapping}. A model may therefore predict a score without measuring the intended construct. This risk becomes sharper in mental-health applications \citep{taylor2025users}, where language models are increasingly used for screening, conversational support, and psychological inference, often ignoring key issues like interpretability, ethics, and clinical readiness \citep{guo2024large, de2025introducing, wulff2026escaping}.

Large language models provide a new experimental setting for this problem. They can generate language under controlled psychological and sociodemographic constraints \citep{ardebili2026mapping, liu2024lost}. LLMs can also complete psychometric questionnaires \citep{franchino2026digital}, provide item-level explanations \citep{fulawka2026large}, and produce free-form narratives \citep{de2025introducing, casoria2025evaluating}. Yet LLMs should not be treated as human participants or as transparent models of human psychology. Recent work shows that LLM survey responses can be unstable, sensitive to prompt perturbations, and affected by response-order and labelling biases \citep{hu2024quantifying, huang2025safety}. Other studies show that LLMs often fail to reproduce human-like response biases in survey settings \citep{de2025introducing, wang2025large}, often in terms of providing lower variance compared to human respondents \citep{wenger2026large}. These findings make LLM-generated psychometric data in need of careful experimental framing.

To this aim, we build on the emerging framework of cognitive digital shadows \cite{ardebili2026mapping, franchino2026digital, esposito2026math}: controlled, language-generating projections through which LLMs are asked to shadow human-like or AI-like profiles under explicit psychological, sociodemographic, and contextual constraints. This framework has already been used to audit how LLMs debate societal issues under demographic and personality constraints \citep{ardebili2026mapping}, to model depression, anxiety, and stress profiles in Mental Health Digital Shadows \citep{franchino2026digital}, and to compare simulated learners and AI tutors in Math Education Digital Shadows \citep{esposito2026math}.

The present study uses cognitive digital shadows to construct language-enhanced psychometric responses. Each response couples three levels of information: a controlled persona, a psychometric score, and a natural-language explanation for the assigned score. We administer validated instruments measuring life satisfaction \citep{di2016measuring}, depression \citep{kroenke2001phq, lovibond1995structure}, anxiety, and stress \citep{lovibond1995structure} to a large population of LLM instances. The design spans multiple model families and model regimes, including thinking and non-thinking models, as well as models differing in safety alignment. This variety importantly allows researchers to ask not only whether psychometric scores can be predicted from generated language, but also whether different model architectures and alignment conditions shape the linguistic expression of psychological profiles.

\subsection{Theoretical groundings for NLP Psychometrics in cognitive science and complex systems}

The theoretical motivation for NLP Psychometrics comes from the Deep Lexical Hypothesis \citep{cutler2023deep}: the idea that psychologically meaningful variation is not only named by language, but partly structured within language. Classical psychological research \citep{perinelli2026substantive, giovanelli2026self} assumed that socially and psychologically important differences become encoded in trait terms because social groups need words to describe consequential patterns of thought, feeling, and behaviour. The Deep Lexical Hypothesis computationally extends this intuition, linking language with psychological constructs \citep{cutler2023deep}. This hypothesis asks whether natural language contains a recoverable trace of psychological meaning, to the extent that relations among words can approximate structures traditionally estimated from human ratings \citep{fatima2021dasentimental}. This shift is crucial for NLP Psychometrics. Psychometric signal should not be expected to reside only in explicit symptom words, sentiment polarity, or a small set of diagnostic markers \citep{al2018absolute, taylor2025users}. If the mental lexicon is a structured system of semantic, affective, and associative relations \citep{stella2024cognitive}, then psychological states can be studied as perturbations of that system rather than as isolated lexical events \citep{fatima2021dasentimental, kenett2016structure, semeraro2025emoatlas}. A depressive profile, for instance, may not merely increase the frequency of words such as sadness or fatigue. It may also reorganise discourse around recurrent negative concepts, reduce semantic exploration, and increase local closure among affectively congruent ideas \citep{de2025introducing, lovibond1995structure, bottesi2015italian}. Likewise, life satisfaction may appear not only through positive words, but through broader conceptual integration, richer emotional balance, or more flexible transitions among self-related themes \citep{manea2015diagnostic}. Such patterns are difficult to capture with bag-of-words models \citep{al2018absolute} because they depend on relations among words, not only on word counts. This relationship becomes observable when language is represented as a cognitive network \citep{stella2024cognitive, semeraro2025emoatlas}, where topology, emotional salience, and semantic organisation jointly define the psychometric trace.

Cognitive network science provides the tools for this representation \citep{stella2024cognitive, siew2019using, haim2026cognitive}. It models cognition as a system of interacting units, where concepts, memories, and words are connected through relations that constrain processing and behaviour \citep{siew2019cognitive}. Network models of the mental lexicon have shown that lexical structure affects word learning, lexical access, and semantic processing \citep{stella2024cognitive, siew2019spreadr}. Related work has shown that multiplex lexical structure can explain naming performance in people with aphasia, suggesting that network topology captures cognitively meaningful constraints on language use \citep{vitevitch2015using}. The network structure of the mental lexicon might also shape how individuals organise syntactic and semantic associations when they produce their own narratives or texts \citep{semeraro2025emoatlas, haim2026cognitive}. Textual forma mentis networks (TFMN) offer a principled way to operationalise this idea \citep{haim2026cognitive}. A textual forma mentis network represents content words as nodes and links them through syntactic and semantic relations, extracted here using the EmoAtlas toolkit \citep{semeraro2025emoatlas}. The resulting network reconstructs the conceptual organisation expressed in discourse. TFMNs can thus reveal not only which words are present in a given text, but also, and especially, how meanings are connected via syntactic specifications between words. In the present NLP Psychometrics framework, TFMNs transform psychometric explanations into interpretable network structures, from which we extract measures grounded in established cognitive frameworks \citep{stella2024cognitive}.

In addition to conceptual associations, language can also convey emotions \citep{mohammad2013crowdsourcing}. Constructs such as life satisfaction, depression, anxiety, and stress are inseparable from affective meaning \citep{lovibond1995structure, bottesi2015italian}. Computational emotion lexicons, like EmoLex \citep{mohammad2013crowdsourcing}, provide a basis for estimating emotional content from words and phrases. In addition, psycholinguistic norms of valence, arousal, and dominance further show that affective properties can be quantified at scale across thousands of lemmas \citep{warriner2015affective}. Yet emotion counts alone are insufficient for psychometric interpretation. Affective words must be interpreted against a baseline, and their role must be considered in relation to the structure of discourse \citep{semeraro2025emoatlas, haim2026cognitive, fatima2021dasentimental}. EmoAtlas, the same network-and-emotion toolkit underlying the textual forma mentis networks introduced above, addresses this need by estimating emotional over- or under-representation relative to a null model, while textual forma mentis networks describe the conceptual scaffold in which those emotions appear.

The combination of EmoAtlas and TFMNs makes explainability central to the NLP Psychometrics framework introduced here. Explainable AI \citep{salih2025perspective} is often introduced after prediction, as a post-hoc attempt to interpret a black-box model \citep{rudin2019stop}. In NLP Psychometrics, interpretability begins earlier: at the level of network representation of language. In NLP Psychometrics, network features describe how discourse is organised \citep{semeraro2025emoatlas, haim2026cognitive}. Emotional profiles describe which affective dimensions are amplified or suppressed \citep{semeraro2025emoatlas}. Predictive models can then be interrogated with feature-attribution methods to identify which linguistic and psychological dimensions contribute to score prediction. This approach aligns with the broader movement from black-box prediction toward glass-box modelling \citep{rudin2019stop, salih2025perspective}. It also provides a more cognitively grounded alternative to purely embedding-based psychometric inference \citep{taylor2025users, fatima2021dasentimental}.

\subsection{Manuscript scope, aims and contributions}

NLP Psychometrics' empirical design follows this logic. Persona variables, Big Five traits \citep{serapio2023personality}, network descriptors \citep{haim2026cognitive}, and emotion features \citep{semeraro2025emoatlas} are treated as distinct but complementary predictors of psychometric scores. Random-forest regressors estimate how much variance each feature family explains, both alone and in combination. SHAP analyses then identify which features contribute most strongly to prediction and in which direction. This design functions as an ablation-style test of NLP Psychometrics. It asks whether the psychometric signal resides primarily in persona metadata, personality constraints, emotional expression, discourse topology, or their interaction.

Two further tests concern transfer, each relaxing a different assumption of the questionnaire-explanation setting. The first test addresses genre. Questionnaire explanations are structured by the items that elicit them, so a model trained and tested only on such explanations may learn the genre rather than the construct. We therefore evaluate whether mappings learned from questionnaire-based explanations transfer to diary-like narratives, which remove the explicit questionnaire scaffold and approximate a more naturalistic form of self-report. Successful transfer would suggest that the learned mapping captures psycholinguistic regularities beyond the item format; failure or feature reversal would be equally informative, revealing which markers are genre-bound and which are robust across registers.  

The second test addresses population: whether a mapping learned entirely from LLM-generated language says anything about real human language at all. For this, we turn to a dataset of authentic, transcribed clinical speech, in which participants carry a binary clinical label (depressed vs.\ control) but no psychometric questionnaire score. Because no continuous ground truth is available here, this test cannot ask whether our models recover a precise score; instead, it asks the weaker but decisive question of whether the predicted scores separate clinically depressed speakers from controls at all. Passing this test would indicate that our LLM-trained models capture markers present in genuine human depressive language, not merely artefacts of LLM-generated text; failing it would indicate that the learned mapping is specific to synthetic data and does not generalise to humans.

Crucially, the aim of NLP Psychometrics is not to diagnose individuals from text, nor to claim that LLMs possess mental states. Rather, this work makes two contributions. First, NLP Psychometrics is structured as an interpretable framework for linking psychometric scores with language, emotional profiles, and cognitive network structure. It mainly aims to extract well-being estimates from linguistic data where psychometric questionnaires are not available, e.g. personal diaries or other NLP datasets. Second, NLP Psychometrics introduces cognitive digital shadows as controlled probes for studying how LLMs express linguistically and encode numerically psychological and sociodemographic constraints. This means that NLP Psychometrics can be used to measure what kind of information drives LLMs' psychometric responses.

To achieve the first aim, in the absence of large-scale datasets linking psychometric questionnaires with linguistic explanations, we design and train NLP Psychometrics models on LLM data and then test their generalisability, first to LLM-generated diaries and then to the clinical speech transcripts described above. By extending this testing on real human data, NLP Psychometrics can secondarily help build a transparent computational framework for studying how psychometric constructs become expressed in language, and how such expression changes across artificial cognitive agents, like LLMs, and humans.

\section{Methods}

\begin{figure}[!ht]
    \centering
    \includegraphics[width=1\linewidth]{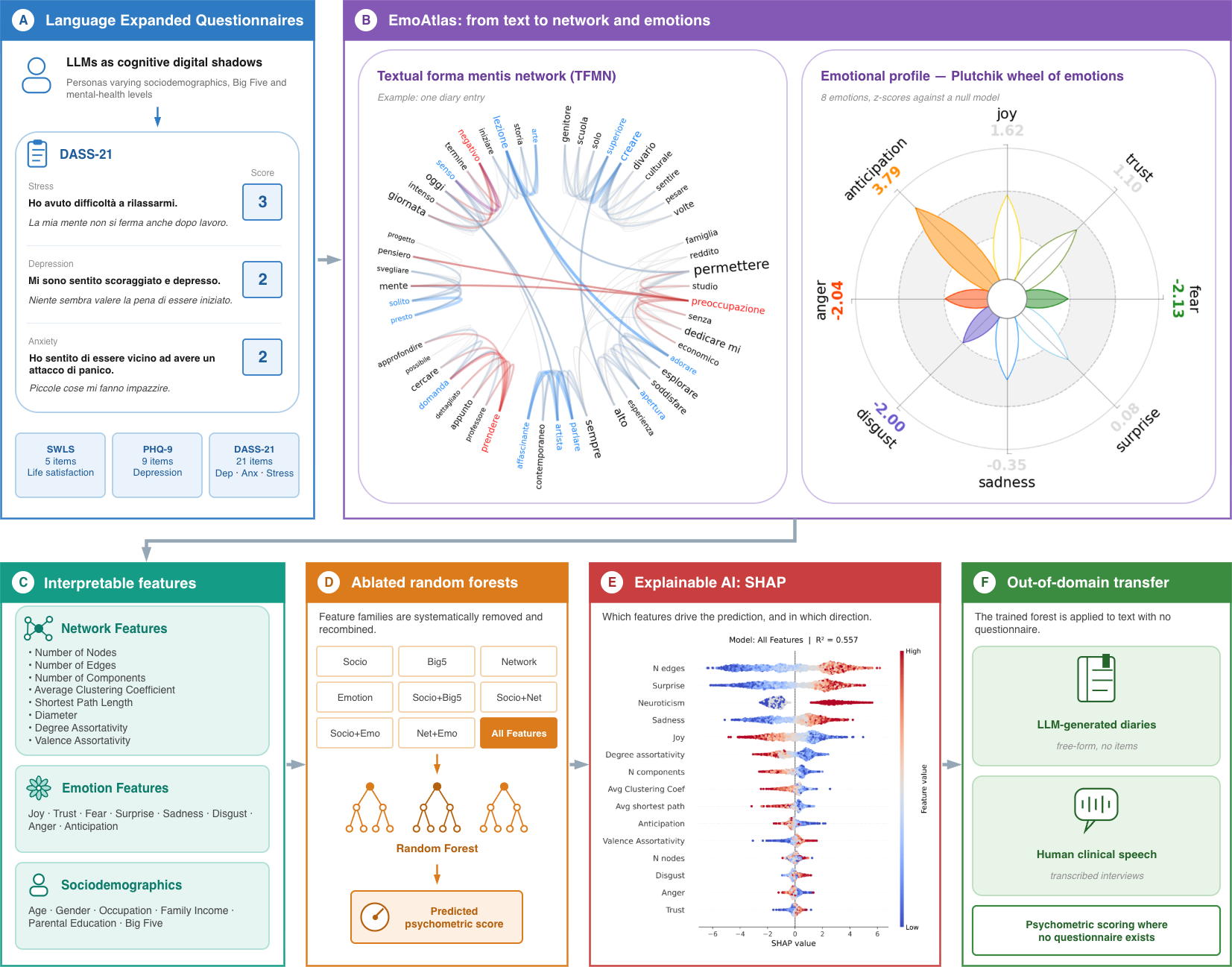}
    \caption{NLP Psychometrics: an interpretable pipeline from language to psychometric scores.}
    \label{fig:infographic}
\end{figure}

We implemented NLP Psychometrics in different stages, which will be described in detail in the current Section and are visually presented in Figure \ref{fig:infographic}. LLMs were prompted to impersonate randomly generated personas and complete validated psychometric questionnaires, producing both item scores and free-text explanations (Sections \ref{sec:psychometric_scales} - \ref{sec:prompting}; Figure \ref{fig:infographic}A). These explanations were processed with EmoAtlas into a textual forma mentis network and an emotional profile (Figure \ref{fig:infographic}B), from which we derived network and emotion features, combined with sociodemographic and Big Five features into four feature families (Figure \ref{fig:infographic}C; Section \ref{sec:feat_set}). These families were systematically removed and recombined to train ablated Random Forest regressors (Figure \ref{fig:infographic}D; Section \ref{sec:rf_ablation_methods}), interpreted via SHAP to identify which features drove predictions and in which direction (Figure \ref{fig:infographic}E; Section \ref{sec:shap_methods}). We then tested whether these models generalise beyond the questionnaire-explanation format, first to LLM-generated diary entries and then to real human speech transcripts (Figure \ref{fig:infographic}F; Sections \ref{sec:diaries_methods} - \ref{sec:human_data_methods}).

The main aim of this pipeline is to use machine-learning prediction to estimate psychometric scores in settings where a direct psychometric assessment is not available, such as personal diaries or transcribed speech. At this stage, the training data consist of text generated by LLMs acting as cognitive digital shadows \citep{ardebili2026mapping, franchino2026digital}; authentic human data enter the pipeline only at the transfer stage (Section \ref{sec:human_data_methods}). To the best of our knowledge, no large-scale human dataset pairing psychometric scores with item-level textual explanations is currently available. Until such data exist, NLP Psychometrics should be read as an auditing and exploration tool rather than as a psychometric measure validated on human populations (cf. Section \ref{sec:limitations}).

\FloatBarrier

\subsection{Data Collection}
We administered psychometric assessments to nine different LLMs to examine the relationship between language characteristics and psychometric scores, sampling across multiple model families and versions. For each LLM (Table \ref{tab:models}), we collected 2,500 questionnaire-explanation instances for SWLS and 2,500 for PHQ-9; for DASS-21, given its higher administration cost across three subscales, data collection was restricted to Mistral Small, the LLM selected for the downstream machine learning analyses (Section \ref{sec:analysis}), yielding a further 2,500 instances. In addition to this LLM-generated data, we later collected a smaller corpus of LLM-generated diary entries (Section \ref{sec:diaries_methods}) and used an external human dataset of transcribed speech from 63 clinically depressed patients and 52 control individuals (Section \ref{sec:human_data_methods}) to evaluate the generalisability of our models beyond LLM-generated text. We describe the psychometric instruments in Section \ref{sec:psychometric_scales}, model selection criteria in Section \ref{sec:model_selection}, prompting strategies in Section \ref{sec:prompting}, and analysis methods in Section \ref{sec:analysis}.

\subsubsection{Psychometric scales} \label{sec:psychometric_scales}
Three well-validated psychometric questionnaires were chosen for the experiment:
\begin{itemize}
    \item Satisfaction With Life Scale (SWLS). This self-report questionnaire was validated in English \cite{diener1985satisfaction} and Italian \cite{di2016measuring}. It comprises 5 items scored on a 7-point Likert scale (1-7) that load onto a single factor, aimed at measuring the cognitive component of subjective well-being.
    \item Patient-Health-Questionnaire-9 (PHQ-9). This psychometric instrument is widely adopted for screening of depression \citep{manea2015diagnostic}, and has been validated in English \citep{kroenke2001phq}. It has been employed in Italian studies, as in \citet{picardi2005screening}. PHQ-9 is composed of 9 items, loading onto a single factor only. 
    \item Depression Anxiety Stress Scales (DASS-21). The DASS-21 was developed and validated in English \citep{lovibond1995structure} and validated in Italian \citep{bottesi2015italian}. It comprises 21 items rated on a 4-point Likert scale (0-3), reflecting the extent to which each statement applied to the respondent over the past week, loading onto three subscales—depression, anxiety, and stress—each composed of 7 items.
\end{itemize}

These instruments were selected for their complementary assessment of mental health and well-being. The SWLS captures the cognitive-evaluative dimension of subjective well-being, reflecting individuals' conscious judgments about their life satisfaction  \cite{diener1985satisfaction}; the PHQ-9 assesses depressive symptomatology specifically \citep{kroenke2001phq}; and the DASS-21 provides a broader, multidimensional assessment of negative affect, distinguishing between depression, anxiety, and stress as related but distinct constructs \citep{lovibond1995structure}. Together, the three instruments provide a balanced evaluation spanning both positive psychological functioning and multiple, differentiated dimensions of psychopathology. All three scales have demonstrated strong psychometric properties \cite{diener1985satisfaction, kroenke2001phq, lovibond1995structure} and are brief enough to minimise risk of hallucination or response inconsistency when administered to LLMs \citep{liu2024lost}. This is particularly relevant given that, in addition to scale scores, the models were required to provide explanatory text for each item (used for subsequent feature extraction), substantially increasing the total token output; this consideration applies with even greater force to the DASS-21, whose 21 items generate a correspondingly larger volume of explanatory text per respondent. The brevity of these scales thus served a dual purpose: maintaining response coherence and consistency across items while generating sufficient high-quality explanatory text for machine learning analysis. This balance between scale length and explanation depth helps ensure both the reliability of LLM-generated responses and the richness of textual data needed for computational feature extraction \citep{liu2024lost}.

\subsubsection{LLMs selection}\label{sec:model_selection}

In Table \ref{tab:models}, we summarise the LLMs we chose to collect our data, the number of parameters they were trained on, their instance and references, as well as their collection mode.

\begin{table}[!ht]
\centering
\footnotesize
\caption{Language models employed in the study: the former 5 are standard instruction-tuned LLMs while the latter 4 are reasoning (chain-of-thought) LLMs. Collection refers to the method used to access each model (API, LM Studio, or Ollama). Text Name refers to the label used to refer to each model in the running text and figures.}
\label{tab:models}
\renewcommand{\arraystretch}{1.4}
\small
\begin{tabular}{
    >{\raggedright\arraybackslash}p{3cm}
    >{\raggedright\arraybackslash}p{3cm}
    c
    >{\raggedright\arraybackslash}p{3cm}
    >{\raggedright\arraybackslash}p{2.6cm}
    c c
}
\toprule
\textbf{Model} & \textbf{Text Name} & \textbf{Params} & \textbf{Instance} & \textbf{Reference} & \textbf{Collection} \\
\midrule
Mistral Small 3.2    & Mistral Small        & 24B          & \path{mistral-small-2506}
    & \url{https://docs.mistral.ai/models/mistral-small-3-2-25-06} & API \\
ANITA-NEXT-24B       & Anita Uncensored     & 24B          & \path{m-polignano/ANITA-NEXT-24B-Dolphin-Mistral-UNCENSORED-ITA}
    & \citep{polignano2024advanced} & LM Studio \\
Qwen3-4B-Instruct    & Qwen-4B-Instruct     & 4B           & \path{Qwen/Qwen3-4B-Instruct-2507}
    & \citep{qwen3technicalreport} & LM Studio \\
GPT-OSS              & GPT-OSS-20B          & 20B          & \path{openai/gpt-oss-20b}
    & \citep{openai2025gptoss120bgptoss20bmodel}
     & LM Studio \\
GPT-OSS-Uncensored   & GPT-OSS-Uncensored   & 20B          & \path{huihui_ai/gpt-oss-abliterated}
    & \url{https://huggingface.co/huihui-ai/Huihui-gpt-oss-20b-BF16-abliterated} & Ollama \\
\midrule
Qwen3-4B-Thinking    & Qwen-4B-Thinking     & 4B           & \path{Qwen/Qwen3-4B-Thinking-2507}
    & \citep{qwen3technicalreport} & Ollama \\
OLMo3-32B-Think      & Olmo-3-32B           & 32B          & \path{allenai/Olmo-3-32B-Think}
    & \citep{olmo3} & Ollama \\
Granite-3.3-8B       & Granite-3-8B         & 8B           & \path{ibm-granite/granite-3.3-8b-instruct}
    & \url{https://huggingface.co/ibm-granite/granite-3.3-8b-instruct} & Ollama \\
Nemotron-3-Nano      & Nemotron-3-Nano-30B  & 30B (3B act.)& \path{nvidia/NVIDIA-Nemotron-3-Nano-30B-A3B} & \citep{nemotron3nano} & Ollama \\
\bottomrule
\end{tabular}
\end{table}

The model selection for this experiment was motivated by (i) the importance of testing LLMs spanning a wide range of parameter scales (i.e., from 4B to 32B) (ii) the need to test LLMs with and without explicit chain-of-thought reasoning (Qwen3-4B-Thinking vs.\ Qwen3-4B-Instruct), and (iii) the significance of understanding the effect of censorship (GPT-OSS vs.\ GPT-OSS-Uncensored, and ANITA-NEXT-24B) on LLMs.

The first critical point of comparison concerns model size, a factor that influences not only raw performance but also the capacity for nuanced cognitive tasks. Recent work \citep{strachan2024testing} has demonstrated this relationship: while LLaMA-70B successfully mastered Theory of Mind tasks such as the False Belief test \citep{wimmer1983beliefs}, smaller models within the same family (13B and 7B parameters) frequently failed at these same tasks. For mental health applications, where understanding emotional nuance is paramount, model dimensionality becomes especially crucial. \citet{pinzuti2025comparative} illustrates this phenomenon: LLaMA-3.3-70B achieved substantially higher accuracy in identifying emotional risks, including self-harm and violence, in zero-shot scenarios compared to its smaller counterparts (i.e., LLaMA-3.2-3B-Instruct), suggesting that larger models possess enhanced capabilities for detecting subtle emotional and psychological content that smaller architectures may overlook. We stress this distinction even further by trying to understand whether this enhanced psychological understanding influences the textual descriptions (e.g., in terms of syntactic structure) given by the models.

Another important characteristic to consider is Chain-of-Thought (CoT) reasoning, sometimes referred to as "thinking". Recent models can expose the intermediate reasoning process underlying their responses. However, whether these thinking tokens systematically and detectably influence model outputs remains unclear. \citet{de2025cost} demonstrated that the CoT process, particularly its length, is psychologically meaningful, correlating with tasks that demand greater human reasoning effort. Conversely, \citet{turpin2023language} revealed a disconnection between reasoning and response: models produced CoT rationalisations that aligned with biased outputs while failing to acknowledge the influence of those biases. This suggests that LLMs do not always express their implicit reasoning processes explicitly. Whether thinking tokens systematically shape model behaviour in detectable ways thus remains an open question.

Finally, a potential confound in our experimental design stems from the safety alignment (or "censorship") applied to commercial LLMs. While intended to prevent harm, these mechanisms often result in refusal behaviours. As noted by \citet{huang2025safety}, aggressive guardrails can inadvertently limit a model's core competencies, a phenomenon known as the "safety tax". This degradation is particularly critical for our work regarding impersonation capabilities. Indeed, \citet{casoria2025evaluating} demonstrated that the uncensored "DarkLlama" model exhibited higher semantic richness and lexical diversity when simulating Big 5 personality traits compared to its censored counterpart, Mixtral. To mitigate this limitation, we include two uncensored models in our study: ANITA-NEXT-24B (a research-oriented uncensored model released by \cite{polignano2024advanced}) and GPT-OSS, an abliterated community model (\url{https://huggingface.co/huihui-ai/Huihui-gpt-oss-20b-BF16-abliterated}).

Having taken into account all of these considerations, we proceed to explain the prompting strategies adopted to collect data from all of the 5 non-thinking and 4 thinking LLMs.

\FloatBarrier

\subsubsection{Prompting strategies} \label{sec:prompting}

Each LLM was prompted to take on the role of a specific persona, and subsequently asked to fill in a psychological questionnaire (see Section \ref{sec:psychometric_scales}) while providing textual explanations for the score assigned to each item. Persona prompting has been reported to have only a limited effect on subjective text-annotation tasks — i.e., how personas shape the labels an LLM assigns to external stimuli. However, the same body of work shows a different pattern for survey-based tasks, where persona variables explain a sizeable share of the variance in the LLM's own self-reported responses \citep{casoria2025evaluating}. This distinction matters for our design: psychometric questionnaires are themselves a survey-based task, so persona effects are expected to be strong here specifically. More broadly, prompting LLMs with structured psychometric instruments is by now an established way to probe model responses across psychological constructs \citep{serapio2023personality, liu2025leveraging}. In our study, persona prompting was employed to maximise response variability, thereby generating the linguistic diversity necessary for examining relationships between language characteristics and psychometric scores.

To inject the persona and instruct the model to complete the desired task, we manipulated both the system prompt and the main prompt, following the approach of \citet{de2025introducing}: the system prompt (or "pre-prompt") assigned the model a specific persona, while the main prompt specified the task to be completed. This two-part structure was used consistently across the study, though the specific content of the main prompt differed depending on the data being collected: this section describes the prompting strategy used to elicit the original questionnaire-explanation data, used for the Random Forest and SHAP analyses (Sections~\ref{sec:rf_ablation_methods} and \ref{sec:shap_methods}); the prompting strategy used to generate diary entries for the out-of-domain transfer analysis follows a different main-prompt structure and, for DASS-21, a different set of conditions, and is described separately in Section~\ref{sec:diaries_methods}. The complete system and main prompts for both procedures are provided in the Supplementary Material.

For the questionnaire-explanation data, personas were generated by randomly varying the following sociodemographic and psychological attributes:

\begin{itemize}
    \item Gender: Male or Female
    \item Age: from 20 to 85 years old
    \item Occupation: Employed / Unemployed / Student
    \item Family income: Low / Medium / High
    \item Education of the Parents: Middle School / High School / University
    \item Personality Traits: High / Medium / Low levels of Big 5 traits (Openness, Conscientiousness, Extraversion, Agreeableness, and Neuroticism).
    \item Mental Health: High / Medium / Low levels of:
    \begin{itemize}
        \item Depression
        \item Anxiety
        \item Stress
    \end{itemize}
\end{itemize}

All values were assigned via uniform random sampling, except for the mental health variable, which was weighted to approximate the elevated symptom prevalences reported for the Italian population during the COVID-19 lockdown \citep{rossi2020covid, bonati2021psychological}. Since these studies report overlapping, non-exclusive symptom categories, we impose a simplified mutually exclusive distribution for persona generation ("no symptoms" 50\%, depression 20\%, anxiety 20\%, stress 10\%); these values are indicative rather than epidemiologically exact.

\subsection{Analyses}\label{sec:analysis}
We conducted three sets of analyses to characterise and validate the relationship between LLM-generated text and psychometric scores. First, we trained Random Forest regressors to quantify how well sociodemographic, personality, and text-derived network and emotion features predict SWLS, PHQ-9, and DASS-21 scores, both individually and in combination (Section \ref{sec:feat_set}, \ref{sec:rf_ablation_methods}). Second, we used SHAP to interpret the best-performing models and identify which individual features drive their predictions (Section \ref{sec:shap_methods}). Finally, to assess whether the learned relationships generalise beyond the questionnaire-explanation format, we tested transfer to a structurally different, out-of-domain register of LLM-generated diary entries (Section \ref{sec:diaries_methods}). We also evaluated whether this transfer extends to authentic human language by applying our models to a dataset of real, transcribed clinical speech (Section \ref{sec:human_data_methods}).

This pipeline rests on a specific rationale: because each persona's assigned psychometric profile systematically shapes the LLM's generated text (cognitive shadowing, Section \ref{sec:prompting}), a model trained to associate text-derived features with these known scores can learn a genuine mapping from language to psychometric constructs. Once learned, this mapping is a property of the trained Random Forest itself, and can therefore be applied to any new text to produce a psychometric rating, including text with no explicit psychometric scaffold, such as free-form diary entries or real human speech transcripts, where no ground-truth score is directly available.

\subsubsection{Feature sets} \label{sec:feat_set}
We organised the predictors into four groups, summarised below.

\begin{itemize}
    \item \textbf{Sociodemographic features}: taken directly from the persona variables described in Section \ref{sec:prompting} — age, gender, occupation, family income and parental education. The assigned mental-health level was deliberately excluded: it is the persona attribute that the questionnaire prompt asks the model to express as a score, so using it as a predictor would amount to regressing the score on a coarse version of itself rather than on the respondent's background.
    \item \textbf{Big Five personality features}: also taken directly from the persona variables described in Section \ref{sec:prompting} — openness, conscientiousness, extraversion, agreeableness and neuroticism (OCEAN).
    \item \textbf{Network features}: extracted from the Italian free-text item explanations produced by each model, concatenated into a single document per respondent. From this text, we built a Textual Forma Mentis Network (TFMN) with EmoAtlas \cite{semeraro2025emoatlas}, a syntactic-semantic network whose nodes are content words and whose edges encode syntactic dependencies and synonymy relations. We derived eight features describing the structure of each respondent's discourse: 
        \begin{enumerate}
            \item \textbf{$N$ nodes}: proxy for lexical diversity, i.e., the size of the inventory of content (non-stop) words \citep{stella2020text}.
            \item \textbf{$N$ edges}: a basic measure of syntactic complexity \citep{stella2020text}.
            \item \textbf{$N$ components}: the number of disconnected subgraphs in the network; a higher number of components indicates a discourse fragmented into separate clusters of concepts that share no direct or indirect syntactic-semantic link, rather than a single unified conceptual structure \citep{carrillo2025textual}.
            \item \textbf{Mean Clustering Coefficient}: the average tendency of a node's neighbours to also be connected; higher values indicate that concepts sharing a common associate are themselves directly linked, reflecting tighter local cohesion and redundancy in how ideas are grouped \citep{haim2023cognitive}.
            \item \textbf{Average Shortest Path Length}: the mean number of edges separating every pair of connected nodes, computed on the largest connected component; shorter paths indicate a more tightly integrated conceptual structure, in which any two ideas in the discourse can be reached through fewer intermediate concepts \citep{de2025introducing}.
            \item \textbf{Diameter}: the length of the longest shortest path between any two nodes in the largest connected component; it captures the maximum conceptual "distance" spanned by the discourse, i.e., how far the most loosely related concepts still connected in the network are from one another \citep{newman2016estimating}.
            \item \textbf{Degree Assortativity}: a Pearson correlation between the degrees of nodes joined by an edge \citep{rossetti2026ysocial}: high assortativity means that highly connected nodes (hubs) tend to link to other hubs, whereas low assortativity indicates hubs connected to many peripheral, low-degree nodes, producing star-like structures in which a few key concepts are surrounded by constellations of specific associates linked only to the hub. 
            \item \textbf{Valence Assortativity}: a Pearson correlation between the psycholinguistic valence of nodes joined by an edge \citep{stella2019forma}: high assortativity means that positively-valenced words tend to connect to other positively-valenced words, and negatively-valenced words to other negatively-valenced words, whereas low or negative assortativity indicates that words of opposite emotional valence are frequently linked together within the same discourse
        \end{enumerate}
    \item \textbf{Emotion features}: eight basic emotions of Plutchik's psychoevolutionary model of affect \citep{plutchik1980general} (i.e., joy, sadness, trust, disgust, fear, anger, anticipation and surprise). They also extracted from the same TFMN-processed text. EmoAtlas scores emotional content against the eight emotions. Detection relies on EmoLex, the NRC Word-Emotion Association Lexicon, in which thousands of lemmas were annotated by crowdsourced raters for their association with each of these eight emotions \citep{mohammad2013crowdsourcing}. For each respondent, EmoAtlas returns one emotional $z$-score per emotion, quantifying how strongly that emotion is over- or under-represented in the text relative to a null model in which word sets of the same size are randomly sampled from the language's emotion lexicon, with $|z| > 1.96$ indicating a statistically significant deviation \citep{semeraro2025emoatlas}. Scoring emotions against a null model, rather than counting affective words directly, makes the resulting profiles comparable across texts differing in length and lexical composition.
\end{itemize}

\subsubsection{Random-forest regression and ablation study} \label{sec:rf_ablation_methods}
For each model and questionnaire separately, we trained Random Forest regressors \cite{breiman2001random} to predict the total score (5-35 for the SWLS, 0-27 for the PHQ-9 and 0-21 for each DASS-21 subscale) from the four feature groups and five of their combinations, for a total of 9 different models:
\begin{enumerate}
    \item Sociodemographics
    \item Big Five
    \item Network Features
    \item Emotion Features
    \item Sociodemographics and Big Five
    \item Sociodemographics and Network Features
    \item Sociodemographics and Emotion Features
    \item Network Features and Emotion Features
    \item All features
\end{enumerate}

This design implements an ablation study: the four feature families are systematically removed and recombined across the nine configurations to measure the contribution of each group and to establish which families are necessary for prediction. The guiding question is where the psychometric signal resides among all available characteristics: in the demographics, in personality constraints, in emotional expression, in the network features extracted from text, or in their interaction. Throughout the Results, we refer to any reduced configuration as an ablated model. When an ablated model matches the full model with a fraction of its features (e.g., the Sociodemographics and Emotions model), we treat it as the reference ablated model, i.e. the most parsimonious configuration preserving predictive power. Analyses were run independently for the non-thinking and thinking models. For DASS-21, given that data collection was restricted to Mistral Small Section~\ref{sec:model_selection}), Random Forest regressors were trained only for this LLM, separately for the total DASS-21 score (0--59) and for each of its three subscales (Depression, Anxiety, Stress; 0-20 each), using the same nine Random Forest models. Using scikit-learn \cite{scikit-learn}, each model was a pipeline chaining mean imputation with a \texttt{RandomForestRegressor}; hyperparameters (number of trees, maximum depth, minimum samples per split and per leaf, and features per split) were tuned by grid search nested in shuffled 5-fold cross-validation, optimising the negative mean squared error. Performance was estimated from out-of-fold predictions under the same 5-fold scheme, and in our results we report RMSE, MAE, $R^2$, and the Spearman correlation ($\rho_s$, with $p$-value) between predicted and observed scores. 

\subsubsection{Feature-importance analysis (SHAP)} \label{sec:shap_methods}
To interpret the fitted models, we used SHapley Additive exPlanations (SHAP) \cite{lundberg2017unified}. For each model and questionnaire, we selected the most parsimonious well-performing feature set: the ablated model with the fewest features among those whose $R^2$ lay within 0.01 of the best-scoring set, defined in Section \ref{sec:rf_ablation_methods}. We then refit a Random Forest with the selected ablated model's tuned hyperparameters, and computed exact Shapley values with \texttt{TreeExplainer} \citep{lundberg2020local}. Feature contributions and their directionality are summarised with beeswarm plots, in which each point represents one respondent's Shapley value for a given feature: horizontal position indicates the magnitude and direction of that feature's contribution to the predicted score, and colour indicates the respondent's value on that feature (low to high). Plots display the top 15 features by mean absolute Shapley value. For SWLS and PHQ-9, this procedure was applied to each of the nine LLMs; for space, we report beeswarm plots in the main text only for the two non-reasoning and two reasoning best-performing LLMs per questionnaire, with the remaining five reported in Appendix \ref{app:shap_results}. For DASS-21, given that Random Forests were trained only for Mistral Small (Section \ref{sec:rf_ablation_methods}), SHAP was likewise computed only for this LLM, separately for each of the three subscales (Depression, Anxiety, Stress).

\subsubsection{Out-of-domain transfer to diary entries} \label{sec:diaries_methods}
The regressors described above are trained and tested on a single textual register, the free-text explanations generated for each questionnaire item. Such within-genre evaluation cannot, on its own, establish whether the learned mapping from language to psychometric score reflects a substantive relationship or merely exploits regularities specific to that genre \citep{harrigian2020models}. We therefore assessed generalisation on an out-of-domain register of free-form diary entries, which dispenses with the questionnaire scaffolding and more closely approximates naturalistic self-report.

For each questionnaire (i.e., SWLS, PHQ-9 and DASS-21), agents were prompted to generate a diary entry reflecting their assigned psychometric profile (250 low-scoring and 250 high-scoring diaries per questionnaire, for a total of 500 diaries per condition; for DASS-21, see below for the two conditions considered). We initially attempted the same transfer using Qwen-4B-Instruct, one of the strongest-performing LLMs in the Random Forest analysis (Section \ref{sec:rf_results}); however, since it did not yield significant results for the majority of the employed diary-prompting variants, we decided to use one of the other best-performing LLMs: Mistral Small. The results for Qwen-4B-Instruct are reported in Appendix \ref{app:rf_results}. We therefore report, in the main text, transfer results for Mistral Small only, using a single diary-prompting condition constrained solely by response length (100-150 words for SWLS, 150-200 words for PHQ-9 and 250-300 words for DASS-21). The whole prompt can be found in the Supplementary Material.

For DASS-21, diaries were generated under two conditions. In the whole-scale condition, agents were prompted with a single total DASS-21 score (range 0-59), yielding 250 low- and 250 high-scoring diaries. In the factor-subscale condition, agents were instead prompted with a score on a single subscale at a time (range 0-20), yielding three independent sets of 250 low- and 250 high-scoring diaries, one each for Depression, Anxiety, and Stress. This design allowed us to test whether the subscale-specific Random Forest models transfer better when diaries are generated from matching, construct-specific prompts (factor-subscale condition) than when generated from a single aggregate score (whole-scale condition). For each condition, the three subscale Random Forest models were applied to the corresponding diary set: in the whole-scale condition, all three models were applied to the same diaries generated from total DASS-21 scores; in the factor-subscale condition, each model was applied to the diaries generated using its own matching subscale (e.g., the Depression Random Forest was tested on diaries generated from low/high Depression scores).

For all the questionnaires, we evaluated transfer by applying the random forest models fitted on the questionnaire explanations to the diary entries, without any retraining, directly testing whether the learned mapping holds across registers. For each questionnaire, we report the \emph{Ground Truth}, the median score of the low- and high-scoring diary groups (as assigned during generation) and their difference ($\Delta$), alongside the \emph{Transfer} result, the median predicted score for each group under the questionnaire-trained Random Forest, its $\Delta$, and the associated Mann–Whitney $U$ test and effect size $r$.

\subsubsection{Out-of-domain transfer to human data} \label{sec:human_data_methods}
To assess whether the transfer observed on LLM-generated diaries extends to authentic human language, we additionally applied the PHQ-9 and DASS-21 (Depression subscale) Random Forest models to a dataset of real human speech transcripts. The transcripts were drawn from the Androids Corpus \citep{tao2023androids}, a benchmark dataset for speech-based depression detection comprising audio recordings of clinically depressed participants and matched controls, and were transcribed to text using Whisper \citep{radford2023robust} by \citet{borraccino2025modeling}. Participants in this dataset were labelled as belonging to a clinically depressed group (formal clinical diagnosis) or a control group (no depression diagnosis); no questionnaire-based ground truth was available for this sample. The final sample comprised $N = 115$ participants (63 clinically depressed, 52 controls).

Sociodemographic information available for this sample was limited to gender, age and education level. Because the categories of education level in this dataset did not match the categories used in our simulated personas (Section \ref{sec:prompting}), and because several sociodemographic variables used in our Random Forest models (e.g., occupation, family income, parental education) had no counterpart in the human sample, we did not include sociodemographic features for this analysis. Therefore, only network and emotion features (Section \ref{sec:feat_set}), extracted from the transcribed text, were used as predictors (i.e., only the Network Features and Emotion Features model). Thus, this feature set was determined a priori by data availability, and not selected based on transfer performance.

The Random Forest models output continuous questionnaire scores, whereas the human sample provides only a binary clinical label. We therefore evaluated transfer at two levels. First, at the level of discrimination, we tested whether the predicted scores separated the two clinically defined groups, using a Mann-Whitney $U$ test on the predicted scores with clinical group as the grouping variable, and summarising the separation as the area under the ROC curve (AUC), obtained as $1 - U/(n_{\text{low}} \cdot n_{\text{high}})$. AUC admits a direct interpretation as the probability that a randomly chosen clinically depressed participant receives a higher predicted score than a randomly chosen control.

Second, at the level of classification, we derived a binary prediction from the continuous scores. To dichotomise the predicted scores into our binary categories, we estimated the decision threshold from the data by fitting a logistic regression with the predicted score as the sole predictor and clinical group as the outcome. Because the model is monotone in a single predictor, it preserves the ordering of participants and therefore leaves AUC unchanged: it determines where the cut is placed, not how well the underlying scores discriminate. To avoid evaluating the threshold on the data used to estimate it, all reported classification metrics were obtained under leave-one-out cross-validation, so that each participant's predicted label comes from a model fitted without that participant.

For each model we report the median predicted score of the two clinical groups and their difference ($\Delta$), the Mann--Whitney $U$ statistic with its $p$-value, and the AUC, alongside cross-validated precision, recall and F1-score for each class, overall accuracy, and macro- and weighted-average metrics (Tables \ref{tab:human_phq} and \ref{tab:human_dass21}).

The evaluation on human data was restricted to the PHQ-9 questionnaire and the DASS-21 Depression subscale, since both instruments assess depression-related symptomatology relevant to this clinical sample. The SWLS model, which assesses life satisfaction, and the DASS-21 Anxiety and Stress models, which assess constructs not targeted by this clinical sample, were not evaluated on human data.

\FloatBarrier


\section{Results}
In this Section, we present our results in three stages: (i) the Random Forest models computed for each questionnaire and each LLM; (ii) the SHAP analysis performed on the Random Forest outputs, to interpret which features drive the predictions; and (iii) the application of the trained Random Forest models to newly generated data and to human data, to assess the transferability of the learned associations.

\subsection{Random Forest Performance - Ablation studies}
\label{sec:rf_results}
\begin{figure}[!hbpt]
    \centering
    \includegraphics[width=\linewidth]{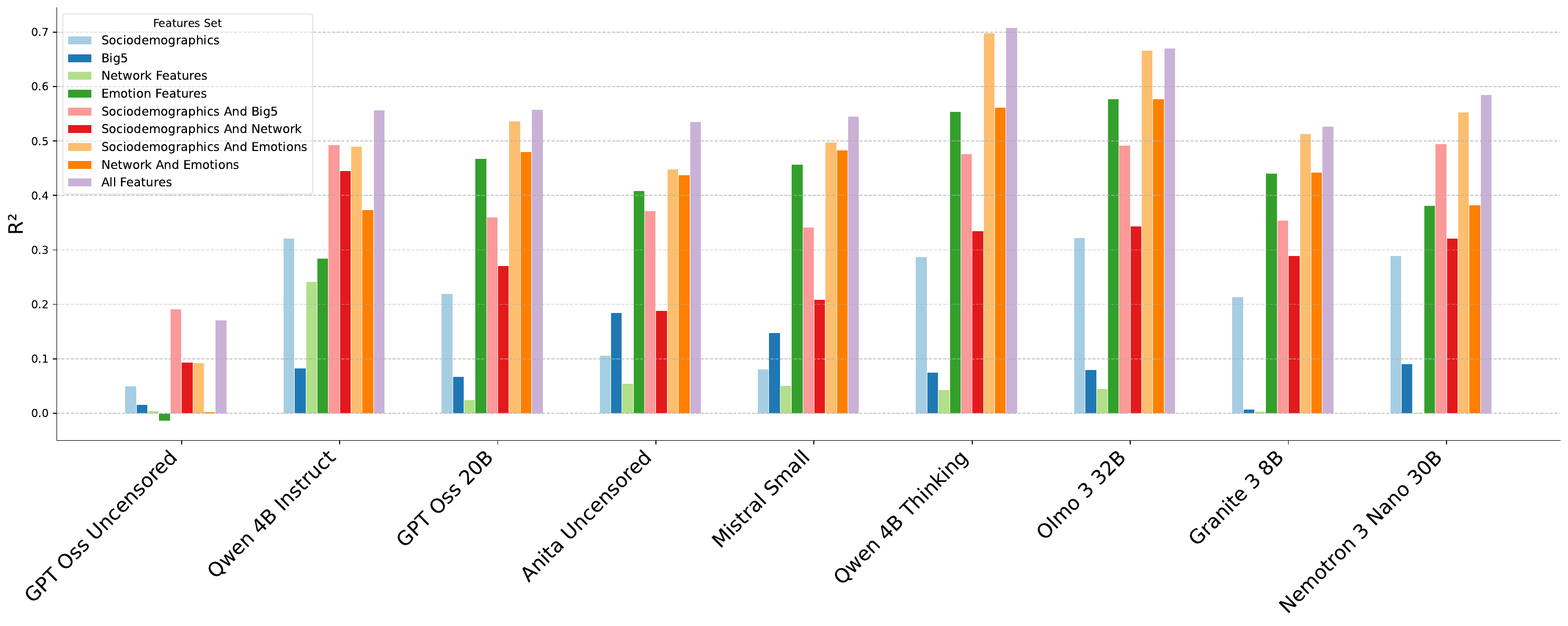}
    \caption{Random Forest prediction performance ($R^2$) on SWLS scores for each of the nine feature-set configurations (see legend), grouped by LLM. Bars within each group correspond to the same feature sets across all LLMs, allowing direct comparison of how predictive performance varies both across LLMs (x-axis) and across feature combinations (bar colour).}
    \label{fig:r2_swls}
\end{figure}

\begin{figure}[!hbpt]
    \centering
    \includegraphics[width=\linewidth]{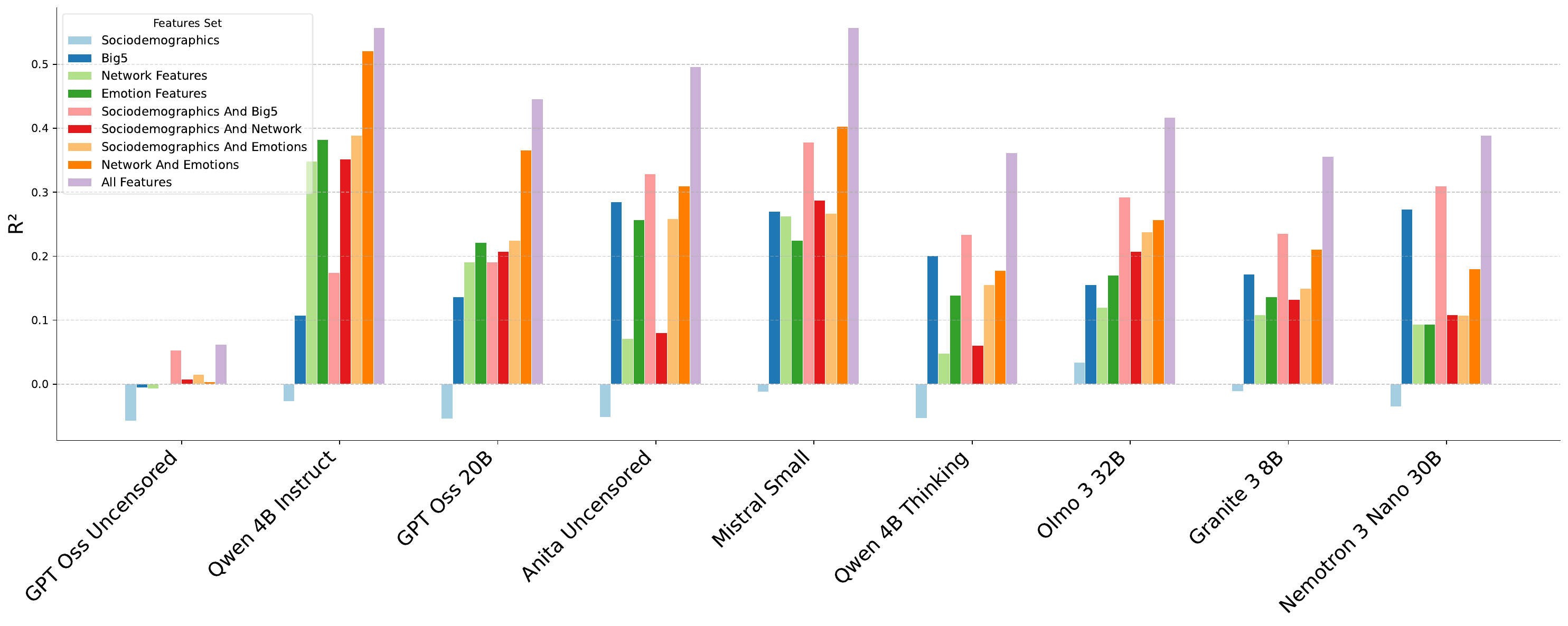}
    \caption{Random Forest prediction performance ($R^2$) on PHQ-9 scores for each of the nine feature-set configurations (see legend), grouped by LLM. Bars within each group correspond to the same feature sets across all LLMs, allowing direct comparison of how predictive performance varies both across LLMs (x-axis) and across feature combinations (bar colour).}
    \label{fig:r2_phq}
\end{figure}

\begin{figure}[!hbpt]
    \centering
    \includegraphics[width=0.75\linewidth]{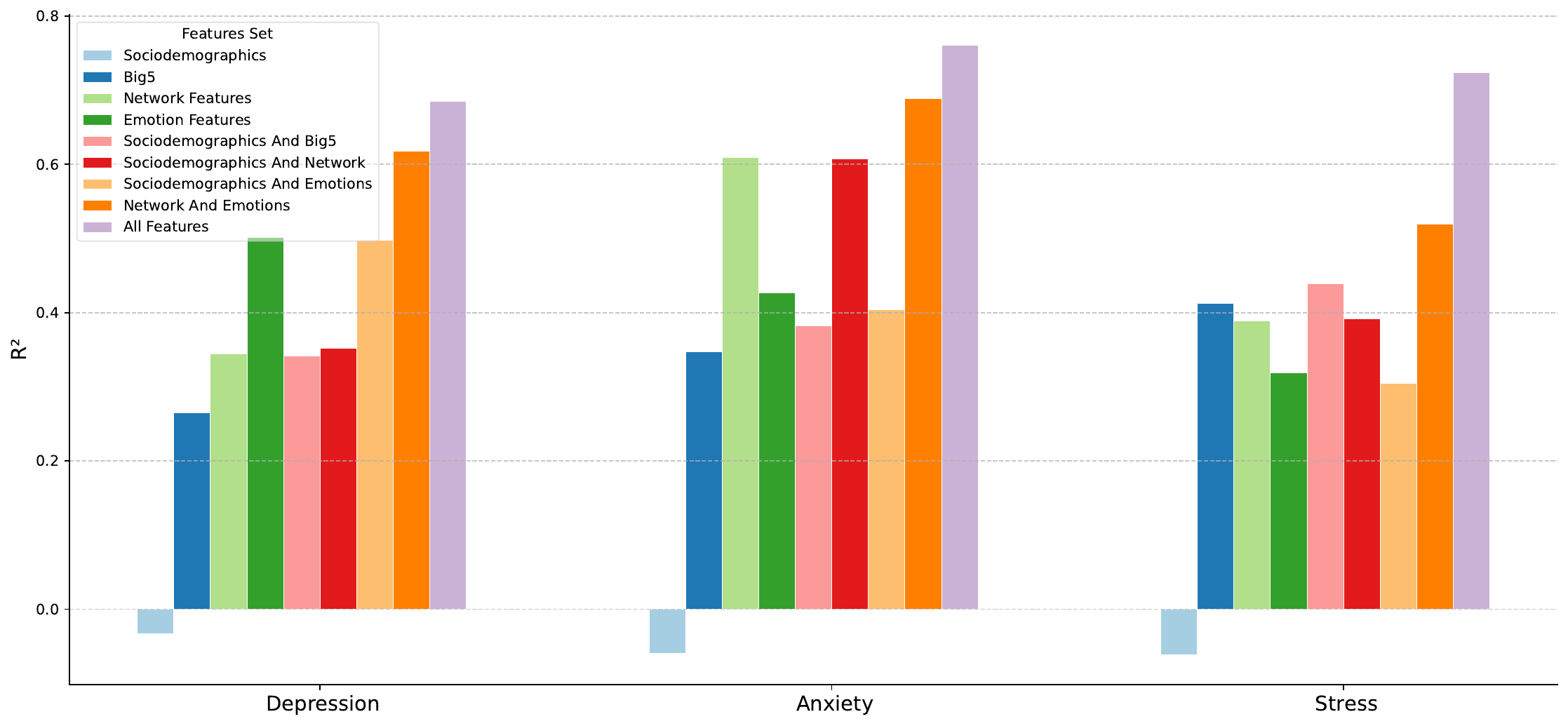}
    \caption{Random Forest prediction performance ($R^2$) on DASS-21 scores for Mistral Small, across each of the nine feature-set configurations (bar colour, see legend) and each factor subscale (Depression, Anxiety, Stress).}
    \label{fig:r2_dass21}
\end{figure}

For each questionnaire (SWLS, PHQ-9 and DASS-21) and for each LLM, we trained nine separate Random Forest models, implementing an ablation design in which the four feature families — sociodemographics, Big Five traits, network features, and emotion features — were systematically removed and recombined into all nine possible configurations (cf. Sections~\ref{sec:feat_set} and~\ref{sec:rf_ablation_methods}). In Figures~\ref{fig:r2_swls}, \ref{fig:r2_phq} and \ref{fig:r2_dass21}, we report the distribution of the $R^2$ coefficient obtained for every LLM and for each of the nine Random Forest configurations, for the prediction of SWLS, PHQ-9 and DASS-21 scores, respectively. For each questionnaire, we additionally report a table summarising the full set of performance metrics (MSE, RMSE, MAE, $R^2$ and $p_s$). The reader can find the tables for the best-performing models in the main text (Tables \ref{tab: rf_shap_llms_swls}, \ref{tab: rf_shap_llms_phq}, and \ref{tab: rf_dass21} for the SWLS, PHQ-9, and DASS-21 questionnaires, respectively), and the additional tables in Appendix \ref{app:rf_results} (Tables \ref{tab: rf_no_shap_llms_swls} and \ref{tab: rf_no_shap_llms_phq}).

\paragraph{SWLS: ablated models rely mainly on sociodemographics and emotions.}
Looking at the SWLS performance scores (Figure \ref{fig:r2_swls} and Tables \ref{tab: rf_shap_llms_swls}, \ref{tab: rf_no_shap_llms_swls}), a clear distinction emerges between LLMs. GPT-OSS-Uncensored, in particular, shows a markedly weaker fit, with $R^2$ ranging from $-0.014$ to $0.192$ across the nine feature configurations, whereas Qwen-4B-Thinking achieves substantially higher predictive performance, with $R^2$ ranging from $0.043$ to $0.708$. Across almost all LLMs (with the exception noted below), a consistent pattern emerges in how performance varies with the feature set used. The model trained on the full feature set (sociodemographics, Big Five personality traits, network features and emotion features) tends to perform best, closely followed by the ablated model combining sociodemographic and emotion features, which performs only marginally worse. For Qwen-4B-Thinking, this ablated model reaches $R^2 = 0.699$, within $0.01$ of the full model ($R^2 = 0.708$), and its predictions are strongly correlated with the true scores ($\rho_s = 0.786$, $p < .001$; Table \ref{tab: rf_shap_llms_swls}). The Random Forest also performs well in absolute terms even when ablated, achieving a MAE of $2.64$ on a scale ranging from 5 to 35. The model based solely on emotion features also performs comparatively well, followed by the model combining sociodemographic and network features. By contrast, the models built on sociodemographics, Big Five traits, and network features alone (without emotion features) consistently rank among the weakest performers, although the magnitude of this gap varies across LLMs. The one clear exception to this pattern is GPT-OSS-Uncensored, which, as noted above, yields uniformly poor predictive performance across all nine Random Forest configurations, suggesting that the features extracted from its generated texts carry comparatively little information about SWLS scores regardless of which feature subset is used.
 
\begin{table}[!ht]
\centering
\scriptsize
\caption{Random Forest performance on SWLS scores across the nine feature-set configurations, for the best performing random forest LLMs (non reasoning at the top and reasoning at the bottom). $N$ is the number of features used; $\rho_s$ is the Spearman correlation between predicted and true scores.}
\begin{tabular}{ll|ccccc|ccccc}
\toprule
\multicolumn{2}{c}{} & \multicolumn{5}{c}{Mistral Small} & \multicolumn{5}{c}{Qwen 4B Instruct} \\
\cmidrule(lr){3-7} \cmidrule(lr){8-12}
Feature Set & $N$ & MSE & RMSE & MAE & $R^2$ & $\rho_s$ & MSE & RMSE & MAE & $R^2$ & $\rho_s$ \\
\midrule
Sociodemographics & 5 & 20.684 & 4.548 & 3.592 & 0.081 & 0.294*** & 33.410 & 5.780 & 4.523 & 0.321 & 0.611*** \\
Big5 & 5 & 19.196 & 4.381 & 3.457 & 0.147 & 0.386*** & 45.129 & 6.718 & 5.281 & 0.083 & 0.329*** \\
Network  & 8 & 21.361 & 4.622 & 3.554 & 0.051 & 0.217*** & 37.317 & 6.109 & 4.804 & 0.242 & 0.480*** \\
Emotion  & 8 & 12.226 & 3.497 & 2.743 & 0.457 & 0.683*** & 35.213 & 5.934 & 4.685 & 0.285 & 0.552*** \\
Sociodemographics And Big5 & 10 & 14.823 & 3.850 & 3.047 & 0.342 & 0.564*** & 24.954 & 4.995 & 3.869 & 0.493 & 0.750*** \\
Sociodemographics And Network & 13 & 17.823 & 4.222 & 3.308 & 0.208 & 0.427*** & 27.325 & 5.227 & 4.069 & 0.445 & 0.697*** \\
Sociodemographics And Emotions & 13 & 11.301 & 3.362 & 2.625 & 0.498 & 0.704*** & 25.114 & 5.011 & 3.852 & 0.490 & 0.739*** \\
Network And Emotions & 16 & 11.631 & 3.410 & 2.678 & 0.483 & 0.697*** & 30.838 & 5.553 & 4.384 & 0.374 & 0.622*** \\
Full Model  & 26 & 10.247 & 3.201 & 2.496 & 0.545 & 0.739*** & 21.790 & 4.668 & 3.572 & 0.557 & 0.783*** \\
\bottomrule
\end{tabular}

\vspace{1em}

\begin{tabular}{ll|ccccc|ccccc}
\toprule
\multicolumn{2}{c}{} & \multicolumn{5}{c}{Qwen 4B Thinking} & \multicolumn{5}{c}{Olmo 3 32B} \\
\cmidrule(lr){3-7} \cmidrule(lr){8-12}
Feature Set & $N$ & MSE & RMSE & MAE & $R^2$ & $\rho_s$ & MSE & RMSE & MAE & $R^2$ & $\rho_s$ \\
\midrule
Sociodemographics & 5 & 27.288 & 5.224 & 4.169 & 0.287 & 0.546*** & 38.190 & 6.180 & 5.075 & 0.322 & 0.559*** \\
Big5 & 5 & 35.412 & 5.951 & 4.745 & 0.075 & 0.304*** & 51.832 & 7.199 & 6.038 & 0.080 & 0.308*** \\
Network  & 8 & 36.612 & 6.051 & 4.834 & 0.043 & 0.203*** & 53.806 & 7.335 & 6.292 & 0.045 & 0.214*** \\
Emotion  & 8 & 17.070 & 4.132 & 3.262 & 0.554 & 0.666*** & 23.845 & 4.883 & 3.860 & 0.577 & 0.756*** \\
Sociodemographics And Big5 & 10 & 20.051 & 4.478 & 3.618 & 0.476 & 0.675*** & 28.652 & 5.353 & 4.365 & 0.492 & 0.693*** \\
Sociodemographics And Network & 13 & 25.474 & 5.047 & 4.034 & 0.334 & 0.579*** & 36.977 & 6.081 & 5.081 & 0.344 & 0.587*** \\
Sociodemographics And Emotions & 13 & 11.534 & 3.396 & 2.641 & 0.699 & 0.786*** & 18.790 & 4.335 & 3.424 & 0.667 & 0.809*** \\
Network And Emotions & 16 & 16.772 & 4.095 & 3.239 & 0.562 & 0.672*** & 23.827 & 4.881 & 3.853 & 0.577 & 0.757*** \\
Full Model  & 26 & 11.164 & 3.341 & 2.613 & 0.708 & 0.793*** & 18.571 & 4.309 & 3.422 & 0.670 & 0.814*** \\
\bottomrule
\end{tabular}
\caption*{\footnotesize *** $p<.001$, ** $p<.01$, * $p<.05$}
\label{tab: rf_shap_llms_swls}
\end{table}

\paragraph{PHQ-9: network and emotion features carry great predictive power.} 
The PHQ-9 results (Figure~\ref{fig:r2_phq}, Tables \ref{tab: rf_shap_llms_phq} and \ref{tab: rf_no_shap_llms_phq}) largely mirror the pattern observed for SWLS. GPT-OSS-Uncensored again yields the weakest fit across all nine feature configurations, with $R^2$ remaining close to zero throughout (never exceeding $0.062$), confirming that the textual features derived from this LLM carry little information about depressive symptomatology as measured by the PHQ-9. At the other end of the spectrum, Qwen-4B-Instruct achieves the strongest predictive performance, with $R^2$ reaching 0.557 for the All Features configuration. With the same $R^2$, Mistral Small also positions itself as the best-performing LLM when trained on all the features; its full-model predictions are strongly correlated with the true scores ($\rho_s = 0.770$, $p < .001$) and achieve a MAE of $3.39$ on a scale ranging from 0 to 27, with the ablated Network and Emotions model remaining accurate at a MAE of $3.98$. As with SWLS, the ranking of feature sets is broadly consistent across LLMs: models incorporating emotion features, whether alone, combined with sociodemographics, combined with network features, or as part of the full feature set, consistently outperform models built solely on sociodemographics, Big Five traits, and network features. The All Features model remains the strongest for every LLM, before the ablated network and emotion features model, reinforcing the central role of text's emotion and network features in predicting PHQ-9 scores. However, for PHQ-9, unlike the results found for SWLS, persona metadata alone carry almost no predictive signal: the Sociodemographics model fails for every LLM, with $R^2$ values close to zero or negative (e.g., $R^2 = -0.012$ for Mistral Small and $R^2 = -0.027$ for Qwen-4B-Instruct; Table \ref{tab: rf_shap_llms_phq}).

\begin{table}[!ht]
\centering
\scriptsize
\caption{Random Forest performance on PHQ scores across the nine feature-set configurations, for the best performing random forest LLMs (non reasoning at the top and reasoning at the bottom). $N$ is the number of features used; $\rho_s$ is the Spearman correlation between predicted and true scores.}
\begin{tabular}{ll|ccccc|ccccc}
\toprule
\multicolumn{2}{c}{} & \multicolumn{5}{c}{Mistral Small} & \multicolumn{5}{c}{Qwen 4B Instruct} \\
\cmidrule(lr){3-7} \cmidrule(lr){8-12}
Feature Set & $N$ & MSE & RMSE & MAE & $R^2$ & $\rho_s$ & MSE & RMSE & MAE & $R^2$ & $\rho_s$ \\
\midrule
Sociodemographics & 5 & 41.820 & 6.467 & 5.502 & -0.012 & 0.153*** & 124.910 & 11.176 & 10.218 & -0.027 & 0.087*** \\
Big5 & 5 & 30.150 & 5.491 & 4.495 & 0.270 & 0.536*** & 108.553 & 10.419 & 9.092 & 0.108 & 0.345*** \\
Network  & 8 & 30.461 & 5.519 & 4.411 & 0.263 & 0.502*** & 79.265 & 8.903 & 7.157 & 0.348 & 0.527*** \\
Emotion  & 8 & 32.027 & 5.659 & 4.633 & 0.225 & 0.447*** & 75.187 & 8.671 & 7.132 & 0.382 & 0.583*** \\
Sociodemographics And Big5 & 10 & 25.706 & 5.070 & 4.050 & 0.378 & 0.621*** & 100.429 & 10.021 & 9.022 & 0.175 & 0.391*** \\
Sociodemographics And Network & 13 & 29.446 & 5.426 & 4.379 & 0.287 & 0.529*** & 78.850 & 8.880 & 7.322 & 0.352 & 0.532*** \\
Sociodemographics And Emotions & 13 & 30.293 & 5.504 & 4.529 & 0.267 & 0.512*** & 74.358 & 8.623 & 7.199 & 0.389 & 0.594*** \\
Network And Emotions & 16 & 24.670 & 4.967 & 3.981 & 0.403 & 0.639*** & 58.276 & 7.634 & 6.024 & 0.521 & 0.662*** \\
Full Model  & 26 & 18.303 & 4.278 & 3.392 & 0.557 & 0.770*** & 53.850 & 7.338 & 5.876 & 0.557 & 0.702*** \\
\bottomrule
\end{tabular}

\vspace{1em}

\begin{tabular}{ll|ccccc|ccccc}
\toprule
\multicolumn{2}{c}{} & \multicolumn{5}{c}{Nemotron 3 Nano 30B} & \multicolumn{5}{c}{Olmo 3 32B} \\
\cmidrule(lr){3-7} \cmidrule(lr){8-12}
Feature Set & $N$ & MSE & RMSE & MAE & $R^2$ & $\rho_s$ & MSE & RMSE & MAE & $R^2$ & $\rho_s$ \\
\midrule
Sociodemographics & 5 & 35.672 & 5.973 & 4.921 & -0.035 & 0.081*** & 36.531 & 6.044 & 4.995 & 0.035 & 0.235*** \\
Big5 & 5 & 25.032 & 5.003 & 3.994 & 0.274 & 0.531*** & 31.970 & 5.654 & 4.568 & 0.155 & 0.413*** \\
Network  & 8 & 31.239 & 5.589 & 4.561 & 0.093 & 0.288*** & 33.297 & 5.770 & 4.752 & 0.120 & 0.341*** \\
Emotion  & 8 & 31.242 & 5.589 & 4.565 & 0.093 & 0.301*** & 31.380 & 5.602 & 4.535 & 0.171 & 0.382*** \\
Sociodemographics And Big5 & 10 & 23.789 & 4.877 & 3.889 & 0.310 & 0.561*** & 26.768 & 5.174 & 4.177 & 0.293 & 0.536*** \\
Sociodemographics And Network & 13 & 30.712 & 5.542 & 4.542 & 0.109 & 0.316*** & 29.997 & 5.477 & 4.516 & 0.207 & 0.436*** \\
Sociodemographics And Emotions & 13 & 30.749 & 5.545 & 4.526 & 0.108 & 0.326*** & 28.822 & 5.369 & 4.374 & 0.238 & 0.461*** \\
Network And Emotions & 16 & 28.254 & 5.315 & 4.329 & 0.180 & 0.415*** & 28.106 & 5.302 & 4.287 & 0.257 & 0.493*** \\
Full Model  & 26 & 21.045 & 4.588 & 3.696 & 0.389 & 0.629*** & 22.055 & 4.696 & 3.784 & 0.417 & 0.649*** \\
\bottomrule
\end{tabular}
\caption*{\footnotesize *** $p<.001$, ** $p<.01$, * $p<.05$}
\label{tab: rf_shap_llms_phq}
\end{table}

\paragraph{DASS-21: network and emotion features outperform sociodemographics}  
Figure \ref{fig:r2_dass21} and Table \ref{tab: rf_dass21} report the full performance metrics for Mistral Small. As stated in Section \ref{sec:rf_ablation_methods}, for the DASS-21 questionnaire, we generated the data and computed its random forest only for Mistral Small, which resulted in the best-performing LLM for both random forest performance and machine learning results. As with PHQ-9, and unlike SWLS, the Sociodemographics model consistently fails to explain any meaningful variance, yielding negative $R^2$ values for all three subscales ($R^2 = -0.033$, $-0.060$ and $-0.061$ for Depression, Anxiety and Stress, respectively), with the corresponding correlation being non-significant or only marginally significant. In line with the results of the other questionnaires, the All Features model achieves the best performance for every subscale, with $R^2 = 0.685$ for depression, $R^2 = 0.760$ for anxiety, and $R^2 = 0.724$ for stress. Its predictions are strongly correlated with the true scores ($\rho_s = 0.816$, $0.837$ and $0.816$, respectively, all $p < .001$), with MAEs of $3.08$, $2.04$ and $2.62$ on subscales ranging from 0 to 20.

The ablated Network and Emotions model is consistently the second-best configuration across all three subscales ($R^2 = 0.618$, $0.689$ and $0.520$ for Depression, Anxiety and Stress, respectively), suggesting that the combination of these two feature families captures a substantial share of the predictive signal captured by the full model. Notably, however, its performance is degraded for the Stress factor compared to Depression and Anxiety, marking an exception to the otherwise stable pattern. Unlike SWLS and PHQ-9, the ranking of the remaining feature sets is not fully consistent across subscales: emotion features are the strongest single contributor for depression ($R^2 = 0.502$), network features for anxiety ($R^2 = 0.609$ for the network features model and $R^2 = 0.607$ for the network and sociodemographics model), and Big 5 traits for stress ($R^2 = 0.413$ for the Big 5-only model and $R^2 = 0.439$ for the sociodemographics and Big 5 model). This suggests that the relative importance of each feature family shifts depending on the specific symptom dimension being predicted.

\begin{table}[!ht]
\centering
\scriptsize
\caption{Random Forest performance on DASS-21 scores for Mistral Small, across the nine feature-set configurations and the three factor subscales (Depression, Anxiety, Stress). $N$ is the number of features used; $\rho_s$ is the Spearman correlation between predicted and true scores.}
\begin{tabular}{ll|ccccc|ccccc}
\toprule
\multicolumn{2}{c}{} & \multicolumn{5}{c}{Mistral Small Depression} & \multicolumn{5}{c}{Mistral Small Anxiety} \\
\cmidrule(lr){3-7} \cmidrule(lr){8-12}
Feature Set & $N$ & MSE & RMSE & MAE & $R^2$ & $\rho_s$ & MSE & RMSE & MAE & $R^2$ & $\rho_s$ \\
\midrule
Sociodemographics & 5 & 48.426 & 6.959 & 5.828 & -0.033 & 0.101*** & 32.330 & 5.686 & 4.881 & -0.060 & 0.024 \\
Big5 & 5 & 34.457 & 5.870 & 4.932 & 0.265 & 0.521*** & 19.928 & 4.464 & 3.837 & 0.347 & 0.583*** \\
Network  & 8 & 30.716 & 5.542 & 4.545 & 0.345 & 0.553*** & 11.920 & 3.453 & 2.592 & 0.609 & 0.729*** \\
Emotion  & 8 & 23.345 & 4.832 & 3.834 & 0.502 & 0.707*** & 17.489 & 4.182 & 3.227 & 0.427 & 0.620*** \\
Sociodemographics And Big5 & 10 & 30.840 & 5.553 & 4.695 & 0.342 & 0.583*** & 18.855 & 4.342 & 3.817 & 0.382 & 0.605*** \\
Sociodemographics And Network & 13 & 30.383 & 5.512 & 4.529 & 0.352 & 0.565*** & 11.989 & 3.463 & 2.623 & 0.607 & 0.726*** \\
Sociodemographics And Emotions & 13 & 23.545 & 4.852 & 3.906 & 0.498 & 0.709*** & 18.170 & 4.263 & 3.374 & 0.404 & 0.616*** \\
Network And Emotions & 16 & 17.904 & 4.231 & 3.346 & 0.618 & 0.773*** & 9.492 & 3.081 & 2.309 & 0.689 & 0.776*** \\
Full Model  & 26 & 14.753 & 3.841 & 3.082 & 0.685 & 0.816*** & 7.308 & 2.703 & 2.042 & 0.760 & 0.837*** \\
\bottomrule
\end{tabular}

\vspace{1em}

\begin{tabular}{ll|ccccc}
\toprule
\multicolumn{2}{c}{} & \multicolumn{5}{c}{Mistral Small Stress} \\
\cmidrule(lr){3-7}
Feature Set & $N$ & MSE & RMSE & MAE & $R^2$ & $\rho_s$ \\
\midrule
Sociodemographics & 5 & 41.929 & 6.475 & 5.460 & -0.061 & 0.007 \\
Big5 & 5 & 23.217 & 4.818 & 3.928 & 0.413 & 0.661*** \\
Network  & 8 & 24.160 & 4.915 & 3.802 & 0.389 & 0.576*** \\
Emotion  & 8 & 26.909 & 5.187 & 4.217 & 0.319 & 0.494*** \\
Sociodemographics And Big5 & 10 & 22.161 & 4.708 & 3.964 & 0.439 & 0.671*** \\
Sociodemographics And Network & 13 & 24.043 & 4.903 & 3.813 & 0.392 & 0.576*** \\
Sociodemographics And Emotions & 13 & 27.465 & 5.241 & 4.343 & 0.305 & 0.486*** \\
Network And Emotions & 16 & 18.987 & 4.357 & 3.409 & 0.520 & 0.639*** \\
Full Model  & 26 & 10.922 & 3.305 & 2.616 & 0.724 & 0.816*** \\
\bottomrule
\end{tabular}
\caption*{\footnotesize *** $p<.001$, ** $p<.01$, * $p<.05$}
\label{tab: rf_dass21}
\end{table}

\paragraph{Summary: Sociodemographics and Emotions for SWLS, Network and Emotions for PHQ-9 and DASS-21} To sum up these preliminary results, in five out of the eight informative LLMs (i.e., excluding GPT-OSS-Uncensored), the ablated model combining sociodemographic and emotion features performs almost as well as the full model on SWLS, whereas for PHQ-9 and DASS-21 the most competitive ablated configuration is the one combining network and emotion features. Taken together, the SWLS, PHQ-9, and DASS-21 results converge on two main findings. First, predictive performance depends strongly on the LLM used to generate the underlying texts, with GPT-OSS-Uncensored consistently the weakest and Qwen-4B-Thinking and Mistral Small consistently among the strongest across SWLS and PHQ-9. Second, emotion- and network-related features are the most reliable contributors to prediction accuracy: across all three questionnaires, the All Features and Network and Emotions models are consistently the top two performers, whilst Sociodemographics models never explain meaningful variance. The DASS-21 results, however, add an important nuance: which single feature family drives performance beyond this core pair is not fixed, but shifts with the specific factor (i.e., emotion features lead for depression, network features for anxiety, and Big Five traits for stress). Overall, these results indicate that the predictive power of the Random Forest models depends jointly on the LLM used to generate the underlying texts and on the specific combination of features employed, with emotion and network features together forming the most consistent core of predictive signal, while the relative importance of the remaining feature families can vary across questionnaires and subscales.

\FloatBarrier

\subsection{SHAP Feature Importance} \label{sec:shap_results}
\begin{figure}[!ht]
    \centering
    \begin{subfigure}[b]{0.45\textwidth}
        \centering
        \includegraphics[width=\textwidth]{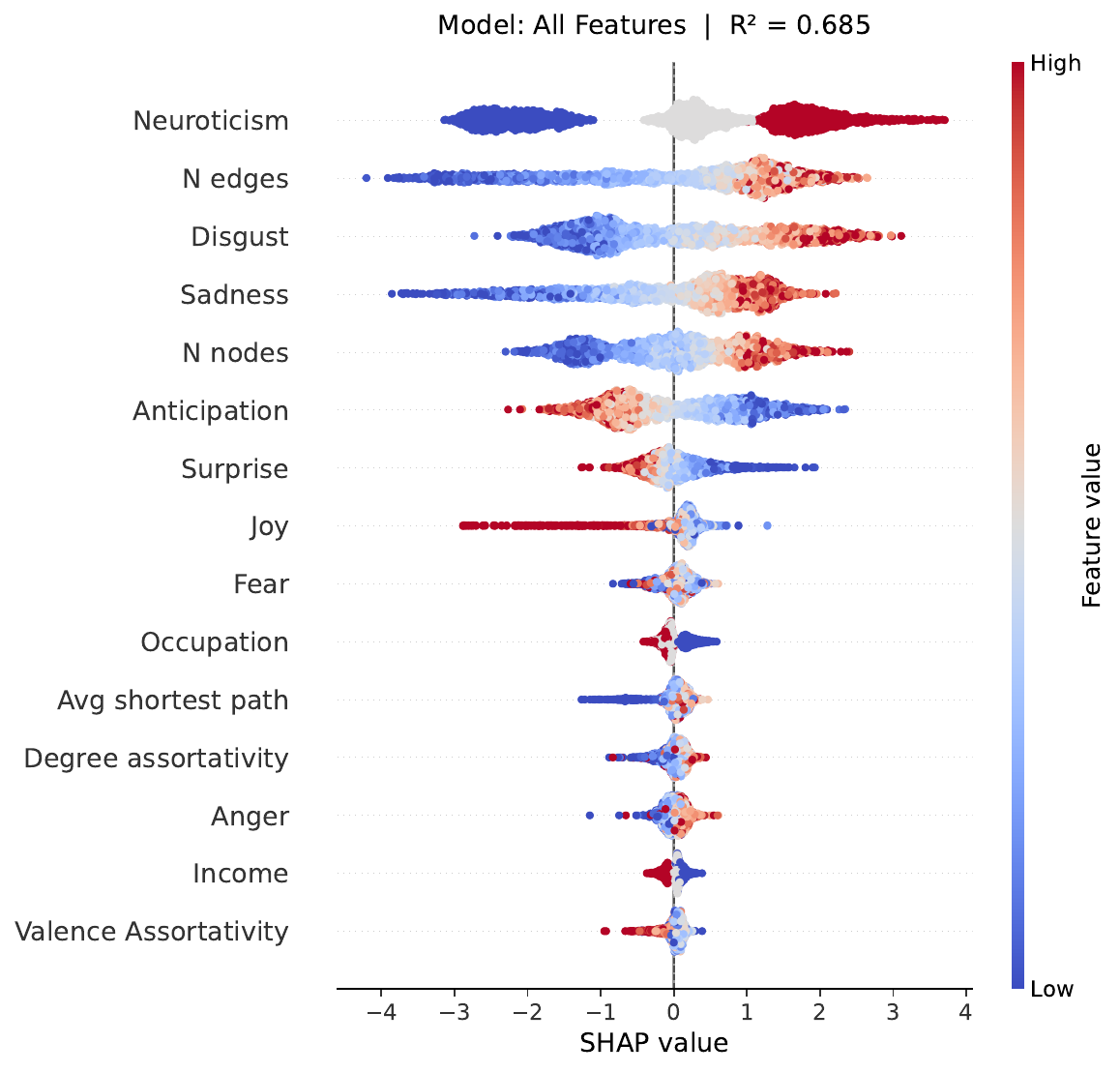}
        \caption{Depression}
    \end{subfigure}
    \hfill
    \begin{subfigure}[b]{0.45\textwidth}
        \centering
        \includegraphics[width=\textwidth]{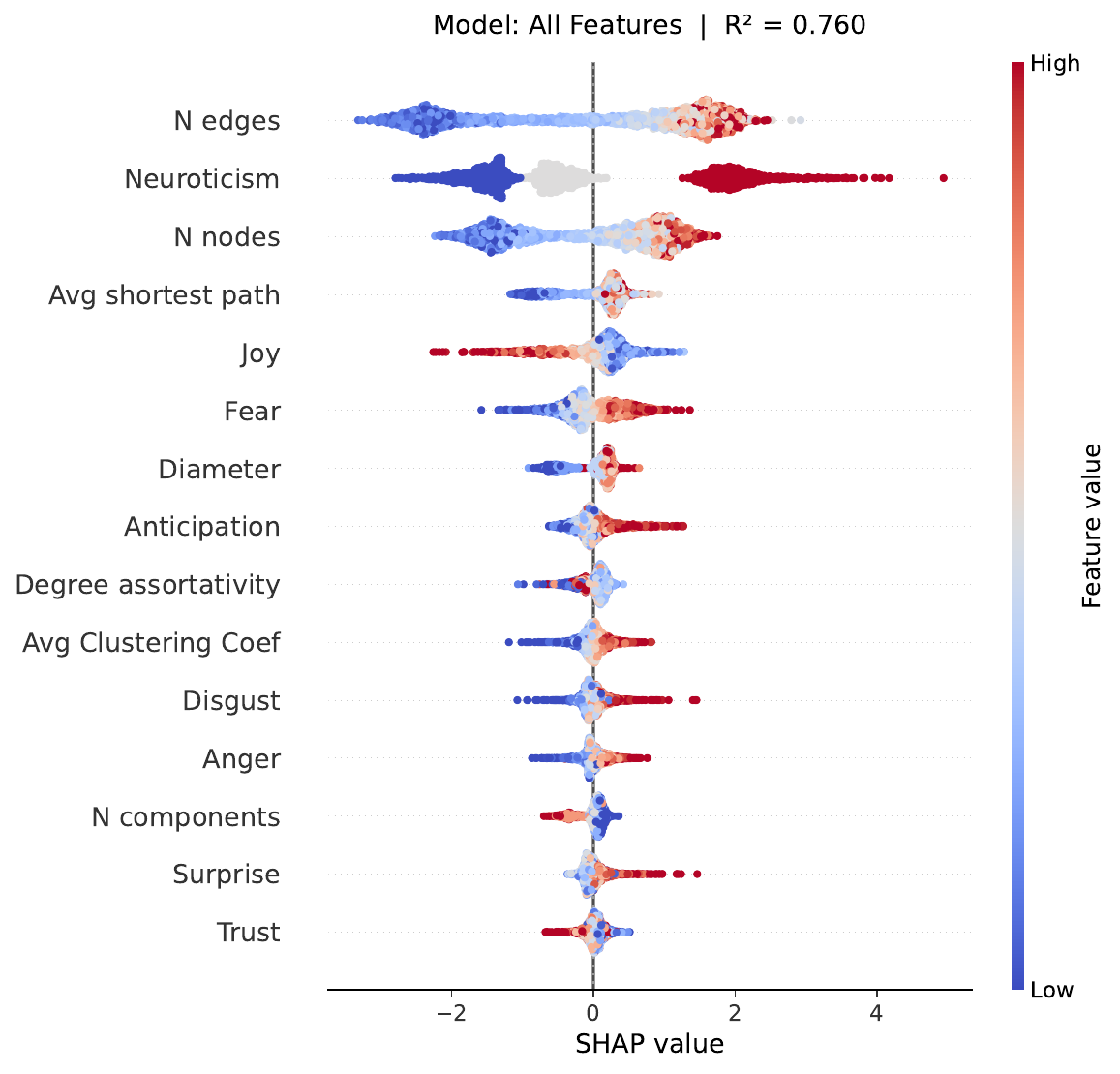}
        \caption{Anxiety}
    \end{subfigure}
    \hfill
    \begin{subfigure}[b]{0.45\textwidth}
        \centering
        \includegraphics[width=\textwidth]{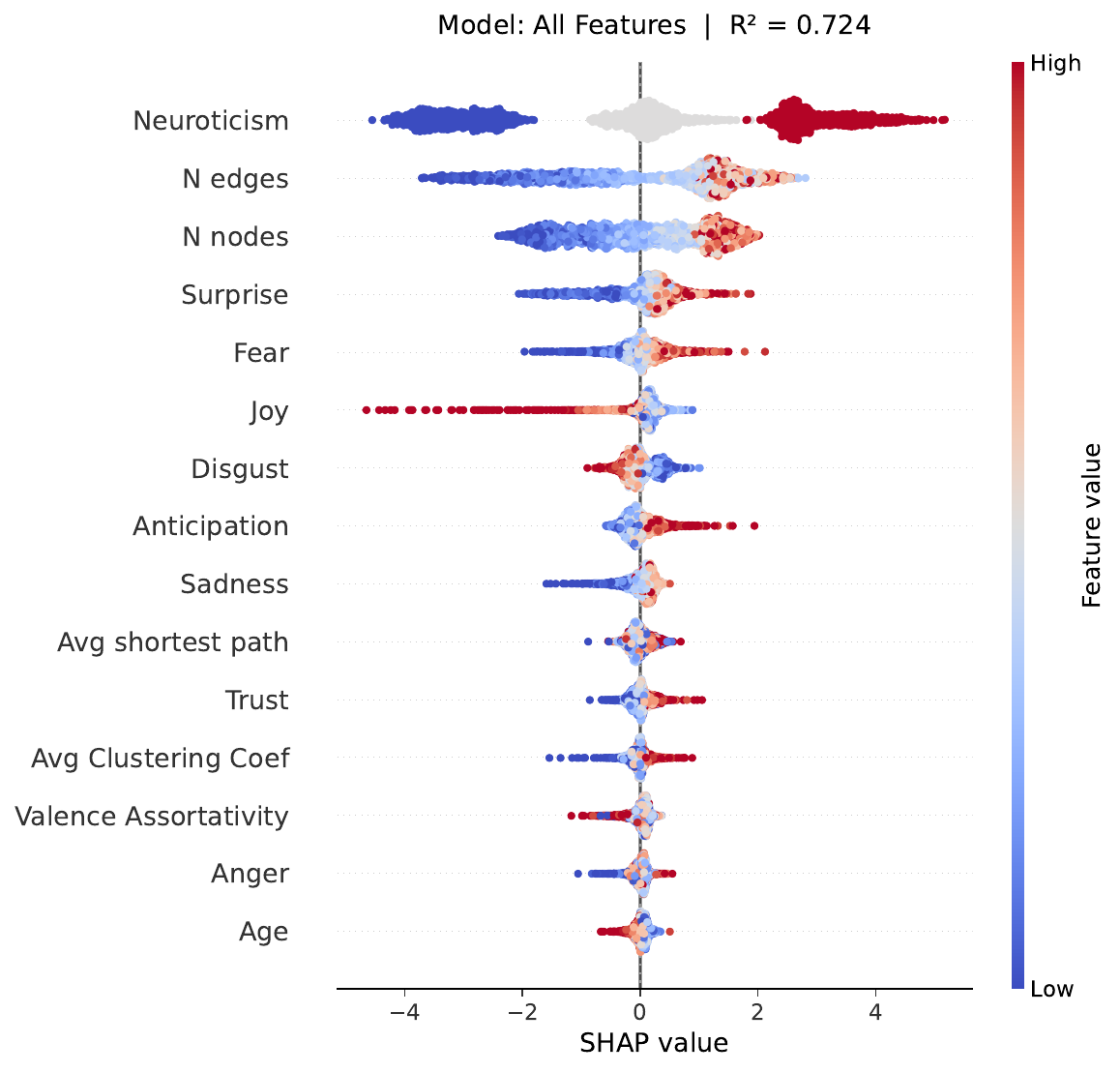}
        \caption{Stress}
    \end{subfigure}
    
   \caption{SHAP summary plots for Mistral Small on the three DASS-21 factor subscales, each showing the best Random Forest model (i.e.,  All Features model). Top 15 features are shown and ranked by mean absolute SHAP value (top to bottom); point colour encodes the feature's original value (red = high, blue = low), and horizontal position indicates the SHAP value's effect on the predicted subscale score.}
    \label{fig:shap_dass21}
\end{figure}

While the Random Forest results identify which feature combinations (or ablated model) yield the best predictive performance, they do not indicate which individual features drive these predictions. To address this, we performed a SHAP analysis on the best-performing Random Forest model for each questionnaire and LLM (i.e., the model presenting the highest $R^2$ and the minimum number of features), allowing us to identify the main predictive features underlying each model's output (cf. Section \ref{sec:shap_methods}). For SWLS and PHQ-9, we report the SHAP summary plots for the four best-performing LLMs identified in the previous analysis, two reasoning and two non-reasoning models, selected to allow a comparison of feature importance patterns across model types (Figures \ref{fig:shap_swls} and \ref{fig:shap_phq}). The reader can find the results of the remaining LLMs in Appendix \ref{app:shap_results}. For DASS-21, since only Mistral Small was analysed, we instead report the SHAP summary plots separately for each of the three subscales, depression, anxiety, and stress (Figure \ref{fig:shap_dass21}).

\paragraph{SWLS: income and emotional content drive satisfaction with life} 
Figure \ref{fig:shap_swls} reports the SHAP summary plots for the four best-performing LLMs on SWLS: Mistral Small and Qwen-4B-Instruct, both best on the All Features model ($R^2 = 0.545$ and $0.557$, respectively), and Olmo-3-32B and Qwen-4B-Thinking, both best on the ablated Sociodemographics and Emotions model ($R^2 = 0.667$ and $0.699$, respectively). Across all four LLMs, emotion features consistently rank among the top predictors: sadness, joy, and fear appear in the top five for every model, with a stable direction of effect; higher sadness and fear push predictions toward lower SWLS scores, while higher joy pushes them upward. This result is in line with our expectation, since the SWLS measures positive aspects of life (i.e., satisfaction with life). Income is the single strongest predictor for the Qwen LLMs (Qwen-4B-Instruct and Qwen-4B-Thinking), with higher income consistently associated with higher predicted life satisfaction. This pattern agrees with findings in human samples, where income is a robust, though bounded, correlate of life evaluation, the cognitive component of subjective well-being targeted by the SWLS \citep{diener2002will, kahneman2010high, jebb2018happiness}. Personality and network features contribute more unevenly across models: neuroticism is influential only for Mistral Small and Qwen-4B-Instruct, where higher values push predictions downward, while network features (e.g., $N$ Edges, Degree Assortativity) appear only in the All Features models and with a comparatively smaller impact than emotion or sociodemographic features. Overall, regardless of the LLM used, emotional content and income emerge as the most robust drivers of SWLS predictions, while the remaining feature families play a more model-dependent, secondary role.

\begin{figure}[!ht]
    \begin{subfigure}[b]{0.45\textwidth}
        \centering
        \includegraphics[width=\textwidth]{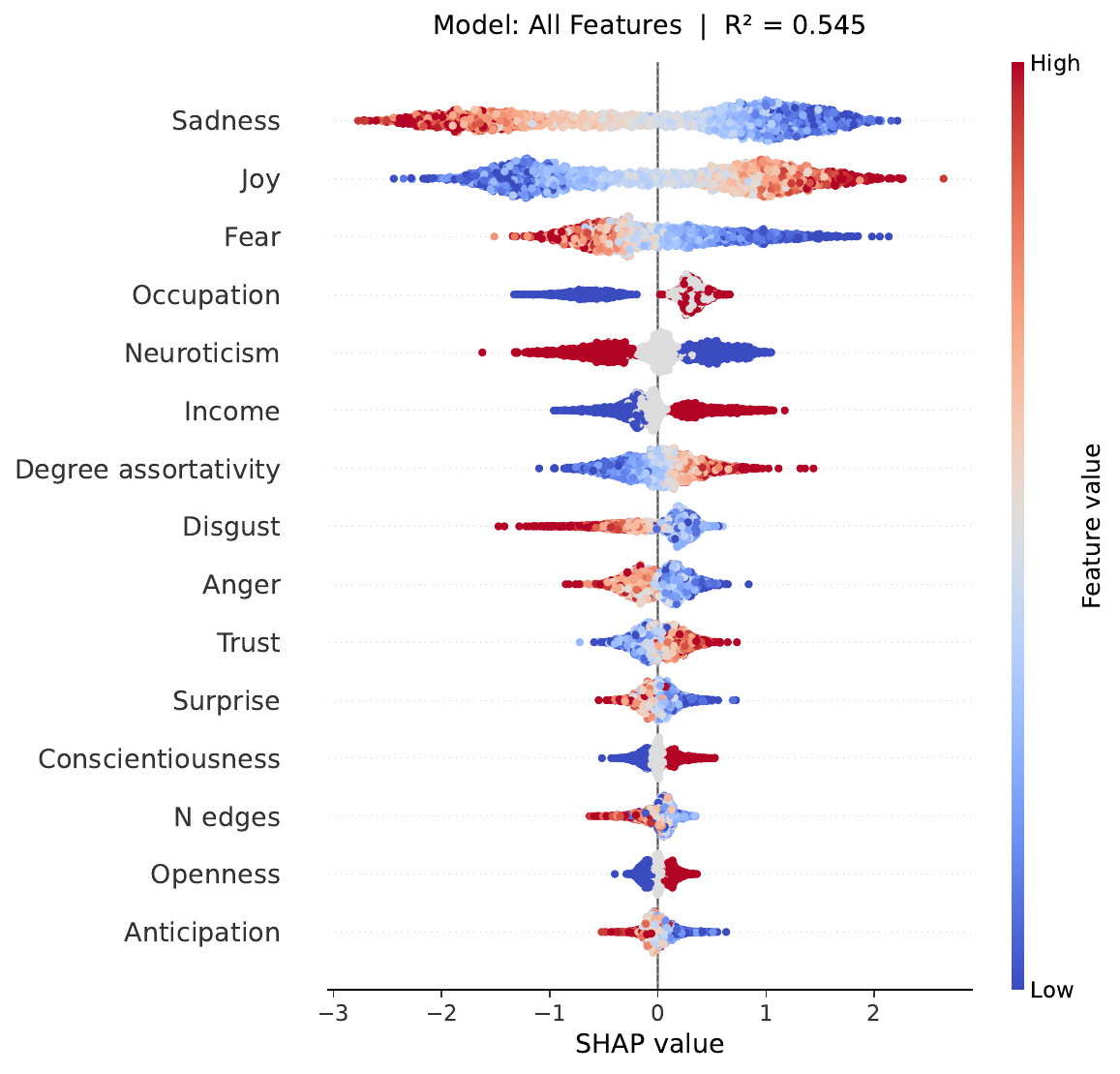}
        \caption{Mistral Small}
    \end{subfigure}
    \hfill
    \begin{subfigure}[b]{0.45\textwidth}
        \centering
        \includegraphics[width=\textwidth]{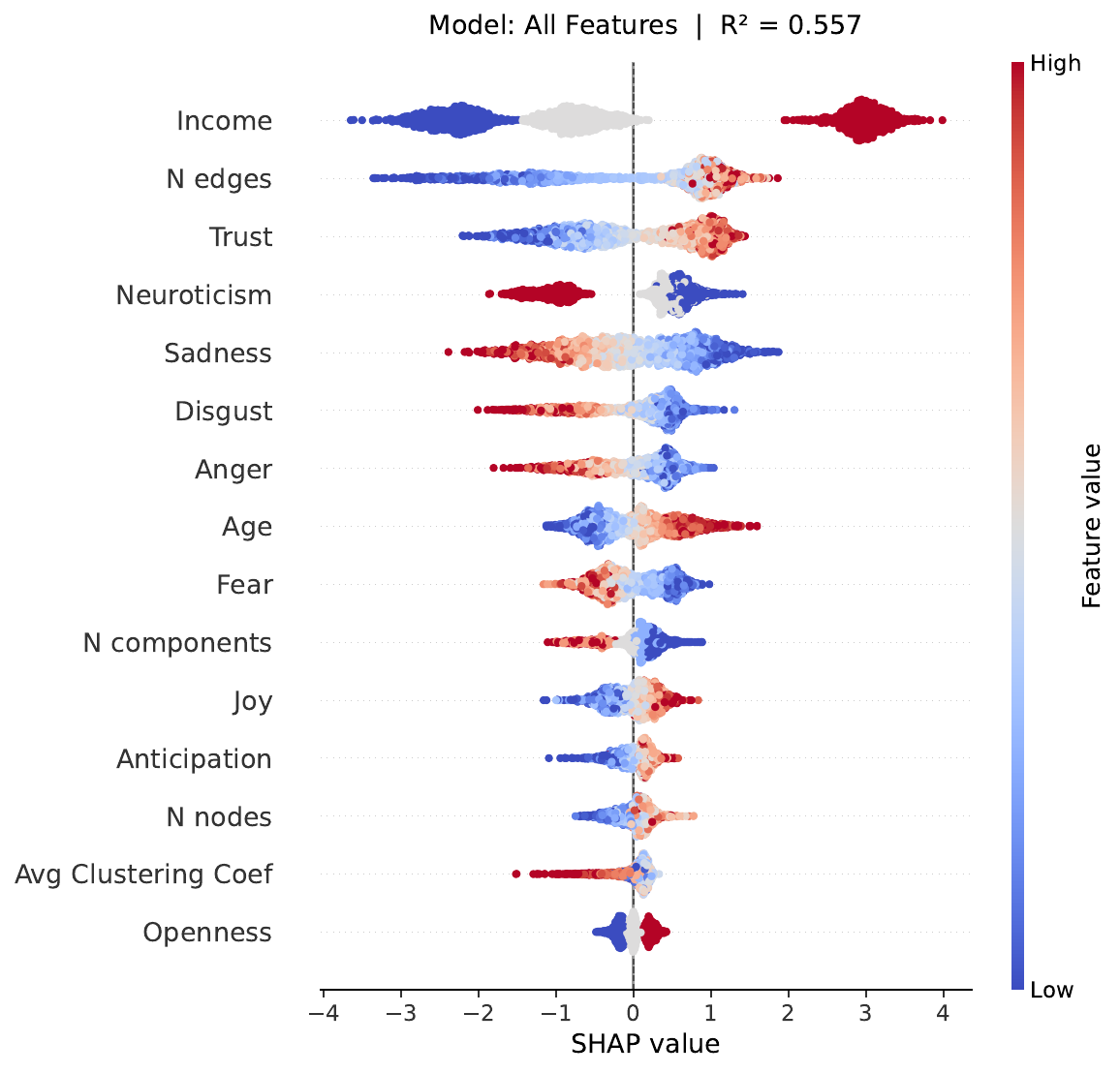}
        \caption{Qwen-4B-Instruct}
    \end{subfigure}
    \begin{subfigure}[b]{0.45\textwidth}
        \centering
        \includegraphics[width=\textwidth]{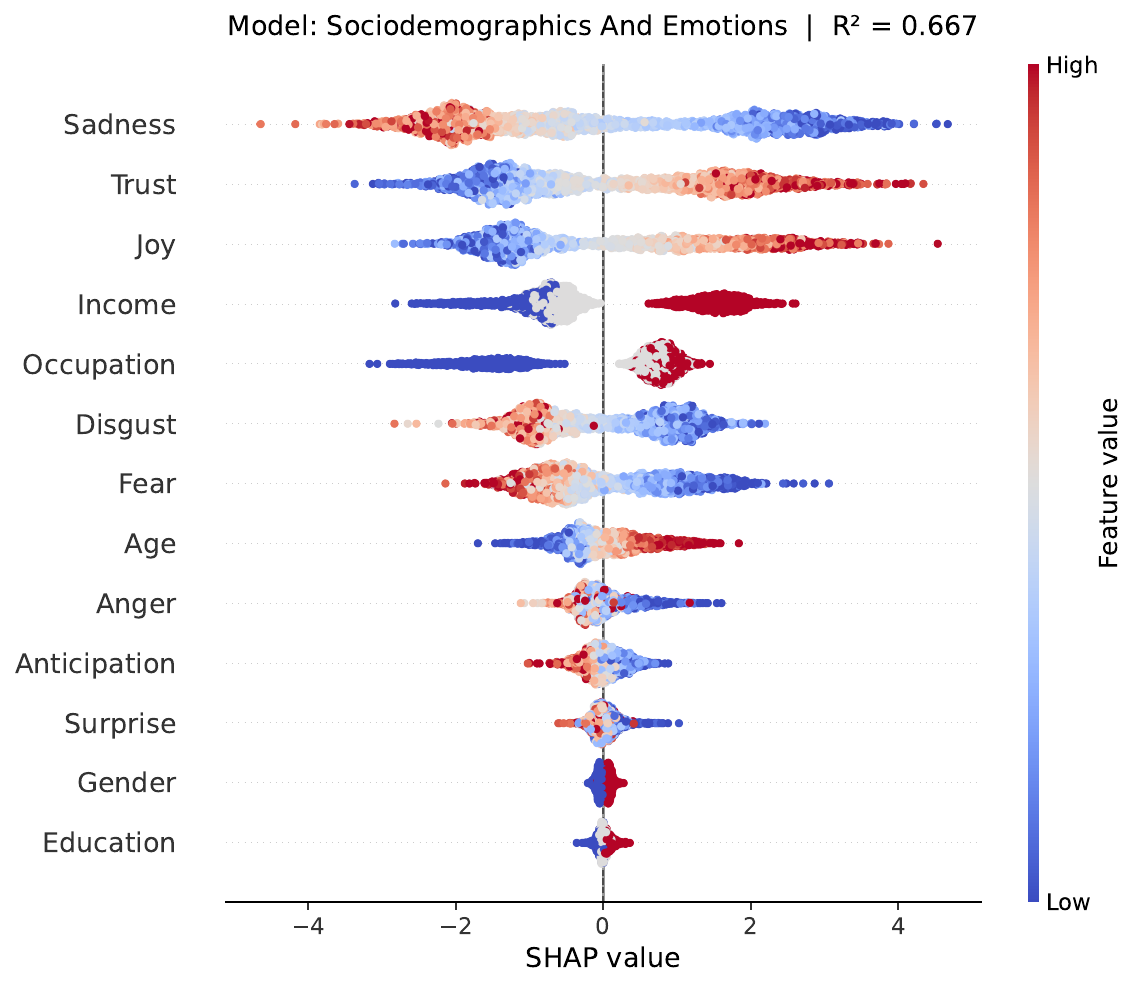} 
        \caption{Olmo-3-32B}
    \end{subfigure}
    \hfill
    \begin{subfigure}[b]{0.45\textwidth}
        \centering
        \includegraphics[width=\textwidth]{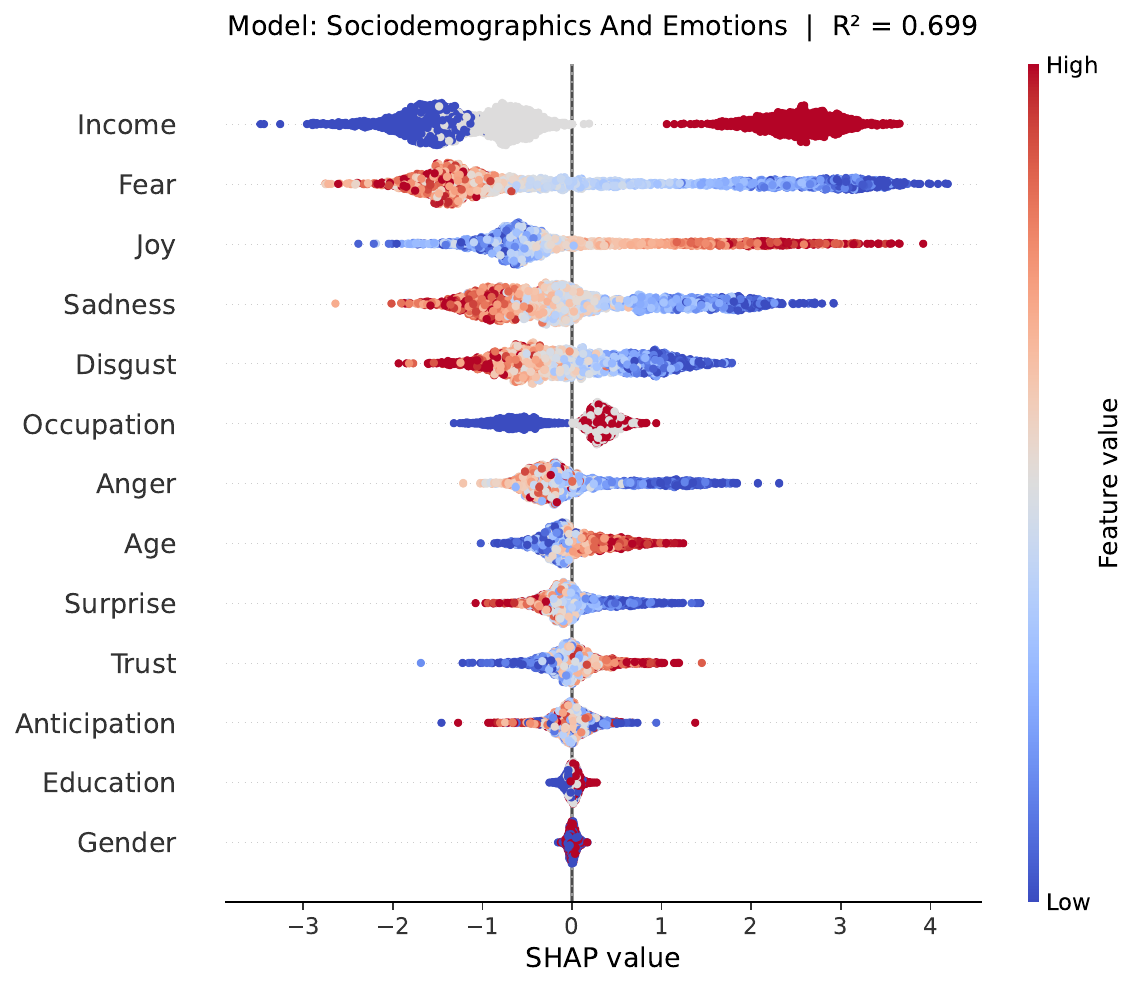}
        \caption{Qwen-4B-Thinking}
    \end{subfigure}
    
   \caption{SHAP summary plots for the four best-performing LLMs on SWLS, each showing the best Random Forest model. Features are ranked by mean absolute SHAP value (top to bottom); point colour encodes the feature's original value (red = high, blue = low), and horizontal position indicates the SHAP value's effect on the predicted SWLS score.}
    \label{fig:shap_swls}
\end{figure}


\paragraph{PHQ-9: network features signal syntactic complexity and rumination.} 
Figure \ref{fig:shap_phq} reports the SHAP summary plots for the four best-performing LLMs on PHQ-9: Mistral Small and Qwen-4B-Instruct (both best on the All Features model, $R^2 = 0.557$ for both), and Olmo-3-32B and Nemotron-3-Nano-30B (also best on the All Features model, $R^2 = 0.417$ and $0.389$, respectively). Across three of the four LLMs, neuroticism is by far the most consistent and dominant predictor (Mistral Small, Olmo-3-32B and Nemotron-3-Nano-30B) and ranks third for Qwen-4B-Instruct, with a uniform direction of effect: higher neuroticism values push predictions toward higher PHQ-9 scores, while lower values push them downward. This pattern is consistent with what we would expect in humans, since neuroticism is a factor influencing depression, anxiety and stress \citep{lovibond1995structure}.  

Emotion features play a comparatively secondary and less stable role. Sadness and joy appear among the top five predictors for the non-reasoning LLMs, with the remaining emotions contributing less. For the reasoning LLMs, other emotions come to the fore: disgust and sadness for Olmo-3-32B, and sadness, anticipation and trust for Nemotron-3-Nano-30B. Sadness is therefore the only emotion ranked consistently among the strongest predictors across both model regimes, which is what a depression scale would lead us to expect \cite{mouchet2008sadness}. Sociodemographic features, by contrast, play only a marginal role: only occupation and income enter the top 15 predictors. Occupation is mainly influential for Olmo-3-32B, where it ranks fifth, and lower for Nemotron-3-Nano-30B and Mistral Small; income is relevant only for Olmo-3-32B and Mistral Small.

Network features contribute consistently to PHQ-9 score predictions: degree assortativity appears among the top predictors for every LLM, together with $N$ edges in all the LLMs except for Olmo-3-32B and $N$ nodes in Olmo and Mistral Small. Texts with more syntactic links between concepts correspond to higher predicted depression levels, and lexically richer texts likewise correspond to more depressed profiles. The low-assortativity structures we observe at higher PHQ-9 scores describe discourse that revolves around a few central concepts, specified at length through many distinct syntactic associates, without repeating the same words or connections: hence the simultaneously high lexical diversity. 
We interpret the joint behaviour of $N$ edges, $N$ nodes and degree assortativity found in most of the top-performing LLMs, as a possible linguistic signature of rumination, the repetitive, self-focused elaboration of a narrow set of negative concepts that is a well-documented cognitive feature of major depression \citep{american2013diagnostic}. However, we note that $N$ nodes and $N$ edges are also partly correlated with raw text length (see Section \ref{sec:limitations}); the co-occurrence of low degree assortativity alongside these size-sensitive measures is the more distinctive part of the pattern, since assortativity is a normalized topological property largely independent of network size.

Overall, and unlike SWLS, neuroticism together with network richness and connectivity are the most robust drivers of PHQ-9 predictions across LLMs, while emotion and sociodemographic features contribute more variably from model to model. The emotional and network content extracted for the PHQ-9 is thus coherent with patterns observed in humans reporting depression, whose language shows heightened negative affect alongside repetitive, self-focused structure \citep{rude2004language, eichstaedt2018facebook, al2018absolute}.

\begin{figure}[!ht]
    \centering
    \begin{subfigure}[b]{0.45\textwidth}
        \centering
        \includegraphics[width=\textwidth]{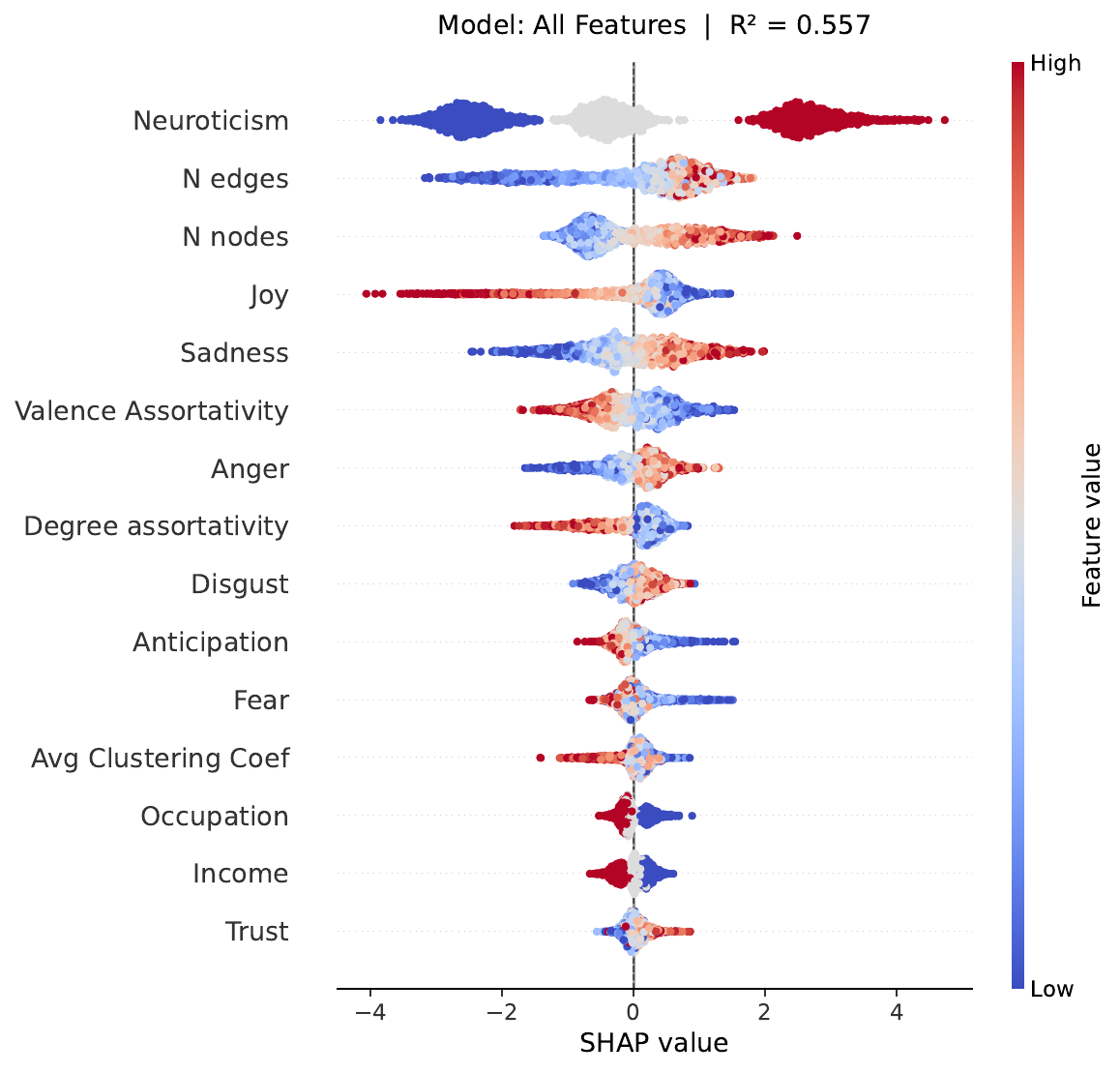}
        \caption{Mistral Small}
    \end{subfigure}
    \hfill
    \begin{subfigure}[b]{0.45\textwidth}
        \centering
        \includegraphics[width=\textwidth]{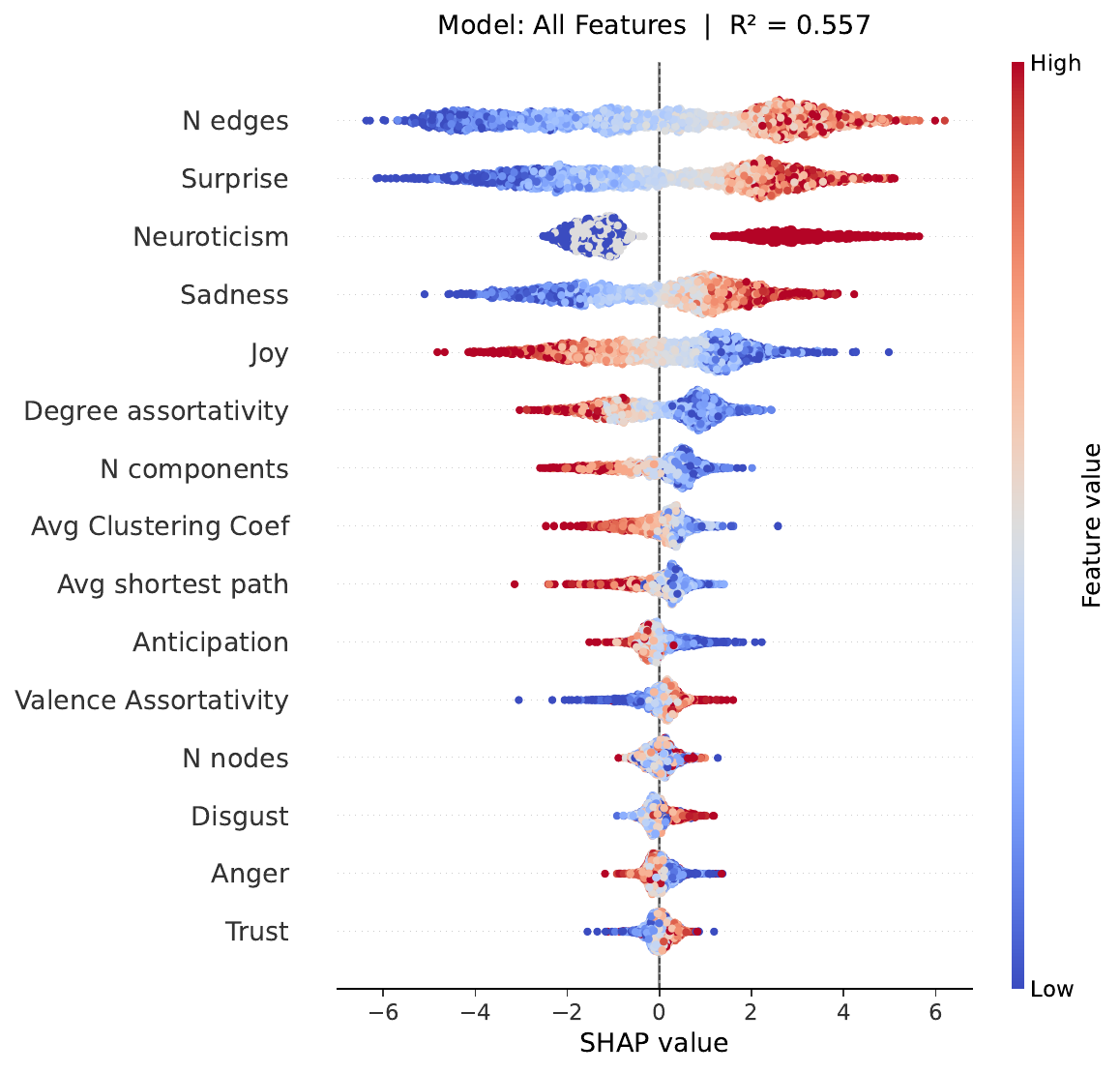}
        \caption{Qwen-4B-Instruct}
    \end{subfigure}
    \begin{subfigure}[b]{0.45\textwidth}
        \centering
        \includegraphics[width=\textwidth]{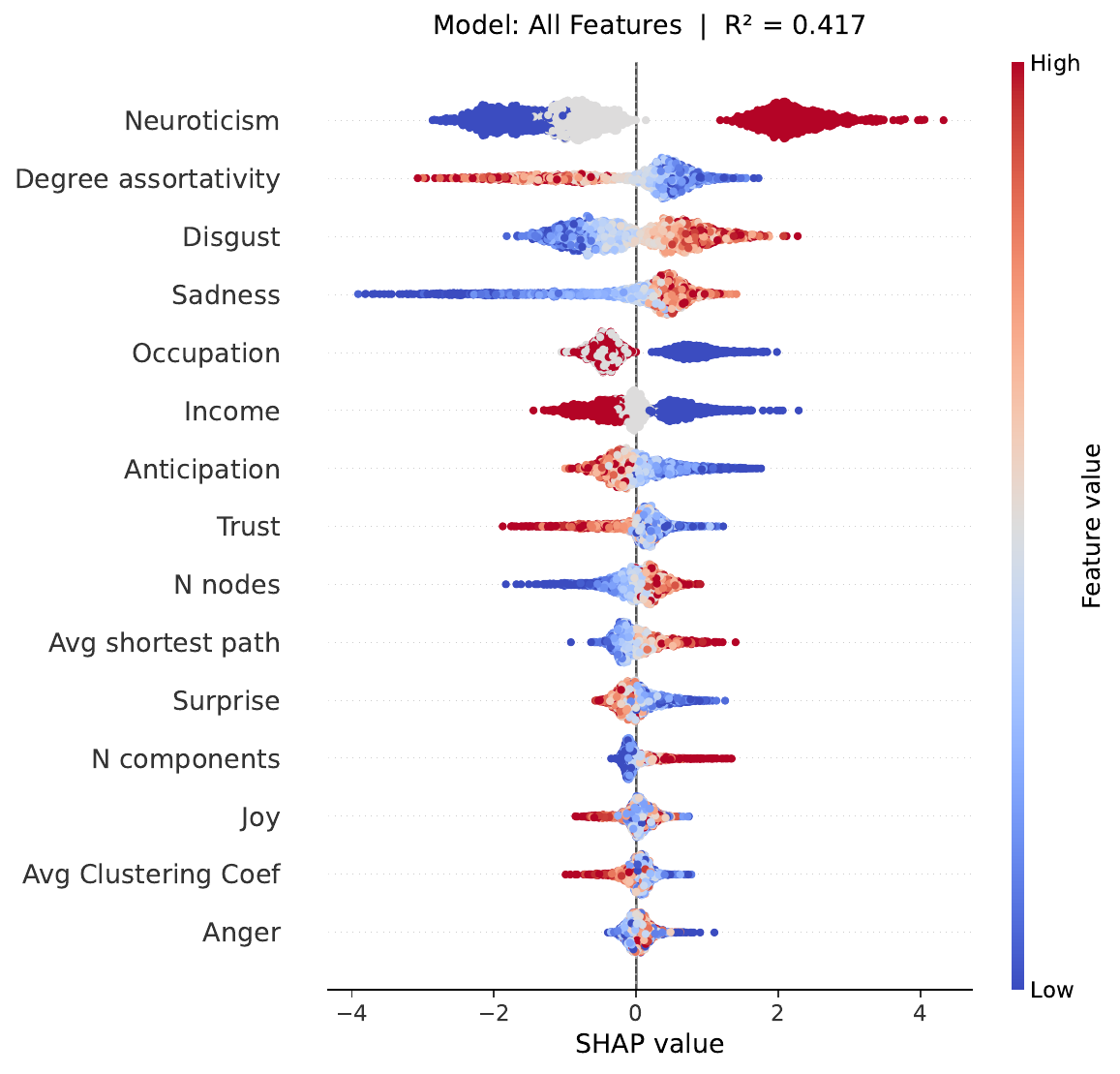} 
        \caption{Olmo-3-32B}
    \end{subfigure}
    \hfill
    \begin{subfigure}[b]{0.45\textwidth}
        \centering
        \includegraphics[width=\textwidth]{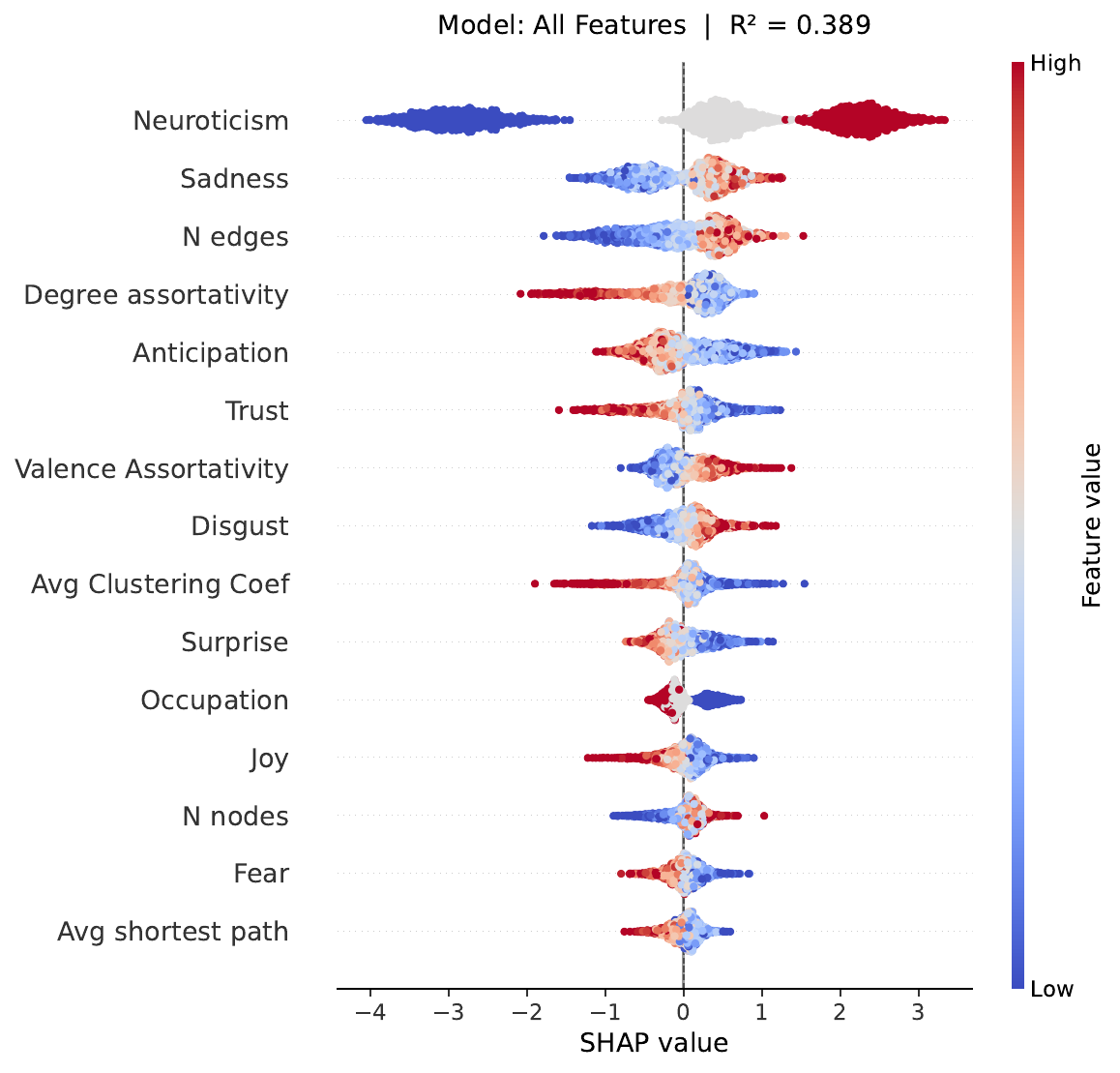}
        \caption{Nemotron-3-Nano-30B}
    \end{subfigure}
   \caption{SHAP summary plots for the four best-performing LLMs on PHQ-9, each showing the best Random Forest model. Features are ranked by mean absolute SHAP value (top to bottom); point colour encodes the feature's original value (red = high, blue = low), and horizontal position indicates the SHAP value's effect on the predicted PHQ-9 score.}
    \label{fig:shap_phq}
\end{figure} 

\paragraph{DASS-21: a shared core of neuroticism and different signatures for each factor.} 
Figure \ref{fig:shap_dass21} reports the SHAP summary plots for Mistral Small on the three DASS-21 factor subscales, all based on the All Features model (Depression, $R^2 = 0.685$; Anxiety, $R^2 = 0.760$; Stress, $R^2 = 0.724$). Neuroticism is the single most important predictor across all three subscales, with higher values consistently pushing predictions toward higher (more severe) depression, anxiety or stress scores, coherently with the conceptualisation of the DASS proposed by \citet{lovibond1995structure}. Network features are also consistently influential, with $N$ edges and $N$ nodes ranking among the top predictors for all three subscales. Interestingly, the role of degree assortativity is not uniform but shifts between constructs. For Depression, we retrieve the same pattern observed for PHQ-9: lower assortativity, i.e. star-like structures centred on a few hub concepts, corresponds to higher predicted severity, the rumination signature described above. For Anxiety, the pattern inverts: personas with higher anxiety scores tend to talk about more, different things, producing textual networks that are more integrated and distributed rather than centred on a few hubs, and higher anxiety is more generally associated with more connected networks across indices of syntactic-lexical complexity. This reflects a discursive mode opposite to depressive rumination \citep{watkins2008constructive}. Beyond this shared core, the contribution of emotion features varies by subscale: disgust and sadness are prominent for depression, fear and joy for anxiety, and surprise and fear for stress. Sociodemographic and personality features other than neuroticism (e.g., occupation, income) contribute only marginally across all three DASS-21 factors. Overall, unlike SWLS, where emotion features and income were the main drivers, and PHQ-9, where neuroticism and network connectivity dominated more uniformly, the Mistral Small DASS-21 subscales show a similar core (neuroticism, network features) but with each subscale drawing on a distinct emotional signature. Mistral Small is thus able to alter the syntactic and network structure of generated text according to the assigned Depression and Anxiety scores, with each construct shaping discourse organisation in a distinct direction.

\paragraph{Summary: a shared emotion-network core with construct-specific predictors} Taken together, the SHAP results for SWLS, PHQ-9, and DASS-21 reveal a consistent underlying structure beneath the LLM- and questionnaire-specific differences. This consistency extends to the five additional LLMs reported in Appendix \ref{app:shap_results} (Figures \ref{fig:app_shap_swls} and \ref{fig:app_shap_phq}): all reproduce the same emotion/income (SWLS) or neuroticism/network (PHQ-9) pattern identified for the four LLMs discussed above, except GPT-OSS-Uncensored, whose best-performing model relies on sociodemographics and personality alone, consistent with its markedly lower $R^2$ reported in Section \ref{sec:rf_results}. First, network features and personality or emotion features together form the most stable predictive core across all three questionnaires, even though which of the two dominates shifts by questionnaire: emotion features (mainly sadness, joy, fear) and family income are the strongest and most consistent predictors for SWLS, whereas neuroticism and network features (i.e., $N$ edges, degree assortativity) take over as the dominant predictors for PHQ-9 and, even more markedly, for all three DASS-21 subscales. Second, sociodemographic and Big Five features beyond neuroticism and income contribute only marginally and inconsistently across LLMs and questionnaires, mirroring the Random Forest finding that Sociodemographics model fail to explain meaningful variance. The DASS-21 results further show that, once neuroticism and network features are accounted for, the remaining predictive signal is subscale-specific: disgust and sadness are most relevant for depression, fear and joy for anxiety, and surprise and fear for stress. Overall, these results indicate that the same feature families identified as most important by the Random Forest models, primarily emotion and network features, also emerge as the most influential at the level of individual predictors, with their relative weight shifting depending on both the construct being predicted and the specific psychological or relational profile it captures.

\FloatBarrier 

\subsection{Out-of-Domain Transfer: LLM-generated diaries and transcribed human interviews}
Having established, in the Random Forest and SHAP analyses above, which feature combinations and individual predictors best explain SWLS, PHQ-9 and DASS-21 scores when models are trained and tested on the same type of text (questionnaire item explanations), we now turn to a stronger test of generalisation: whether these models can predict psychometric scores from a structurally different type of generated text, namely diary entries, and, further, from real transcribed human interviews. This analysis moves NLP Psychometrics across three progressively less controlled settings: a controlled setting, where the score is known and the text is scaffolded by questionnaire items; a free setting, where the persona is prompted with a score but produces unconstrained diary text; and real human data, where no psychometric ground truth exists and only a binary clinical label is available. Hence, for human transcripts, we evaluate transfer both as discrimination (whether predicted scores separate the clinical and control groups) and as classification (binary labels obtained from a data-estimated threshold). Full methodological details are provided in Sections \ref{sec:diaries_methods} and \ref{sec:human_data_methods}. 

For this analysis, we used only Mistral Small, one of the strongest performers in the Random Forest analysis. We initially attempted the same transfer using Qwen-4B-Instruct, the overall best-performing LLM in the Random Forest analysis; however, the transfer proved inconsistent across diary-prompting variants for that model, in particular failing for the PHQ-9 under the reference condition; the corresponding results are nonetheless reported in Appendix \ref{app:diaries_results}. In what follows, we present, for each questionnaire, the transfer performance on diary entries and, where available (PHQ-9 and DASS-21 only; cf.\ Section~\ref{sec:human_data_methods}), on human data. For each condition, Ground Truth and Transfer median scores (and their group difference, $\Delta$) are reported as defined in Section \ref{sec:diaries_methods}.

\paragraph{SWLS: transfer preserves scores distribution} 
As shown in Table \ref{tab:swls_transfer}, the ground-truth diary entries show a clear separation between the low and high SWLS score groups (median scores of 7.00 and 33.00, respectively, $\Delta = 26.00$). Applying the Random Forest model trained on questionnaire explanations to these diary entries, without any retraining, still recovers a highly significant separation between groups (median predicted scores of 16.89 and 20.92, $\Delta = 4.03$; Mann–Whitney $U = 9{,}693$, $p < .0001$). The transfer effect size ($r = +0.690$) is close in magnitude to the in-domain Spearman correlation for the same model ($\rho_s = 0.739$; Table \ref{tab: rf_shap_llms_swls}), indicating that the learned mapping remains largely intact when applied to unconstrained diary text. Predicted scores compress toward the centre of the SWLS range relative to the ground-truth medians, however, suggesting that the model preserves the ordering between low- and high-SWLS individuals more faithfully than the absolute magnitude of their scores.

\begin{table}[!htbp]
\centering
\caption{Transfer learning results for SWLS. Median predicted scores for participants low vs.\ high on SWLS, with Mann-Whitney $U$ test and effect size.}
\label{tab:swls_transfer}
\begin{tabular}{l rrrrrrl}
\toprule
 & \multicolumn{3}{c}{Median Score} & \multicolumn{2}{c}{Mann--Whitney} & \multicolumn{2}{c}{Effect Size} \\
\cmidrule(lr){2-4} \cmidrule(lr){5-6} \cmidrule(lr){7-8}
 & Low & High & $\Delta$ (H$-$L) & $U$ & $p$ & $r$ & \\
\midrule
Ground Truth & 7.00 & 33.00 & 26.00 & --- & --- & --- \\
Transfer & 16.89 & 20.92 & 4.03 & 9,693 & $<$\,.0001 & +0.690 \\
\bottomrule
\end{tabular}
\end{table}

\paragraph{PHQ-9: robust diary transfer, weaker but significant human transfer} 
Table \ref{tab:phq_transfer} shows an analogous pattern found for SWLS, also for PHQ-9: the ground-truth low and high groups are again clearly separated (medians of 2.00 and 25.00, $\Delta = 23.00$), and the transferred model preserves this separation with an even larger effect size (medians of 4.81 and 13.44, $\Delta = 8.63$; $U = 2{,}891$, $p < .0001$, $r = +0.907$), suggesting that the PHQ-9 Random Forest model generalises particularly well to the diary-entry format. When applied to real transcribed human interviews (Table \ref{tab:human_phq}), the model still separates the two clinically defined groups, though by a narrow margin: predicted scores are higher for the clinical than for the control group (medians of 6.40 vs.\ 5.10, $\Delta = 1.30$; $U = 1{,}000$, $p = .0003$), corresponding to an AUC of $.695$. Converting these scores into binary labels yields modest classification performance (accuracy $= .62$; macro-average F1 $= .62$), above chance but well below the separation obtained on LLM-generated diaries.

\begin{table}[!htbp]
\centering
\caption{Transfer learning results for PHQ-9. Median predicted scores for participants low vs.\ high on PHQ-9, with Mann-Whitney $U$ test and effect size.}
\label{tab:phq_transfer}
\begin{tabular}{l rrrrrrl}
\toprule
 & \multicolumn{3}{c}{Median Score} & \multicolumn{2}{c}{Mann--Whitney} & \multicolumn{2}{c}{Effect Size} \\
\cmidrule(lr){2-4} \cmidrule(lr){5-6} \cmidrule(lr){7-8}
 & Low & High & $\Delta$ (H$-$L) & $U$ & $p$ & $r$ & \\
\midrule
Ground Truth & 2.00 & 25.00 & 23.00 & --- & --- & --- \\
Transfer & 4.81 & 13.44 & 8.63 & 2,891 & $<$\,.0001 & +0.907 \\
\bottomrule
\end{tabular}
\end{table}

\begin{table}[!htbp]
\centering
\caption{Transfer of the random forest trained on Mistral Small PHQ-9  data to human-transcribed conversations ($N = 115$; 63 clinical, 52 control). At the top, we present the separation of predicted scores between clinically defined groups. At the bottom, the classification performance of a single-feature logistic regression on the predicted score, evaluated with leave-one-out cross-validation.}
\label{tab:human_phq}
\begin{tabular}{l rrr rr r}
\toprule
\multicolumn{7}{l}{\textit{Predicted score by clinical group}} \\
\midrule
 & \multicolumn{3}{c}{Median Score} & \multicolumn{2}{c}{Mann--Whitney} & \\
\cmidrule(lr){2-4} \cmidrule(lr){5-6}
 & Low & High & $\Delta$ (H$-$L) & $U$ & $p$ & AUC \\
\midrule
Transfer & 5.10 & 6.40 & 1.30 & 1,000 & .0003 & .695 \\
\bottomrule

\vspace{0.5em} \\

\toprule
\multicolumn{7}{l}{\textit{Classification (LOO-CV)}} \\
\midrule
 & Precision & Recall & F1 & $n$ & & \\
\midrule
High          & .66 & .62 & .64 & 63 & & \\
Low           & .57 & .62 & .59 & 52 & & \\
Macro avg.    & .62 & .62 & .62 &    & & \\
Weighted avg. & .62 & .62 & .62 &    & & \\
\midrule
Accuracy      & \multicolumn{4}{l}{.62} & & \\
\bottomrule
\end{tabular}
\end{table}

\paragraph{DASS-21: factor-specific transfer strength, with Depression leading} 
Table \ref{tab:dass21_transfer} reports transfer results for the two diary-generation conditions defined in Section \ref{sec:diaries_methods}. In the whole-scale condition, the Depression model transfers most robustly (median predicted scores of 26.39 vs.\ 39.20, $\Delta = 12.82$; $U = 6{,}148$, $p < .0001$, $r = +0.803$), followed by the Stress model ($\Delta = 10.90$, $r = +0.684$) and the Anxiety model ($\Delta = 7.53$, $r = +0.534$). In the factor-subscale condition, where each model is tested on diaries generated specifically to reflect its own subscale, the Depression model again shows the strongest and most significant separation ($\Delta = 14.69$, $r = +0.750$), while the Stress model shows a moderate effect ($\Delta = 6.94$, $r = +0.437$) and the Anxiety model the weakest ($\Delta = 2.54$, $r = +0.237$). The consistency of this ranking across both conditions, with Depression transferring best and Anxiety worst, regardless of the diaries' generation condition, suggests that this pattern reflects a genuine difference in how reliably each construct is expressed in diary-style text.

Applied to transcribed human interviews (Table \ref{tab:human_dass21}), the DASS-21 Depression model transfers more strongly than its PHQ-9 counterpart. Predicted scores separate the clinical from the control group by a substantial margin (medians of 32.40 vs.\ 21.32, $\Delta = 11.08$; $U = 720$, $p < .0001$), giving an AUC of $.780$; that is, a randomly chosen clinically depressed participant receives a higher predicted depression score than a randomly chosen control in roughly four cases out of five. The corresponding classification performance is likewise the strongest obtained on human data (accuracy $= .68$; macro-average F1 $= .67$), with balanced precision and recall across the two classes.

\begin{table}[!ht]
\centering
\caption{Transfer learning results for DASS-21, evaluated under two diary-generation conditions: whole-scale diaries, generated from low/high total DASS-21 scores (range 0-59) and factor-subscale diaries, generated separately for each subscale from low/high scores on that subscale alone (range 0-20). For each condition, the Depression, Anxiety and Stress Random Forest models are applied to the corresponding diary set (in the factor-subscale condition, each model is tested on diaries generated using its own matching subscale). Median predicted scores for the low and high groups are reported together with Mann-Whitney $U$ tests and effect sizes.}
\label{tab:dass21_transfer}
\begin{tabular}{l rrrrrr}
\toprule
& \multicolumn{3}{c}{Median Score} & \multicolumn{2}{c}{Mann--Whitney} & Effect Size \\
\cmidrule(lr){2-4} \cmidrule(lr){5-6} \cmidrule(lr){7-7}
& Low & High & $\Delta$ & $U$ & $p$ & $r$ \\
\midrule
\multicolumn{7}{l}{\textit{Whole DASS-21 score (range 0--59)}} \\
\quad Ground Truth & 4.00 & 59.00 & 55.00 & --- & --- & --- \\
\quad Depression RF & 26.39 & 39.20 & 12.82 & 6{,}148 & $<$.0001 &  0.803 \\
\quad Anxiety RF    & 30.54 & 38.07 & 7.53  & 14{,}559 & $<$.0001 &  0.534 \\
\quad Stress RF     & 28.64 & 39.55 & 10.90 & 9{,}865 & $<$.0001 &  0.684 \\
\midrule
\multicolumn{7}{l}{\textit{Factor subscale scores (range 0--20)}} \\
\quad Ground Truth  & 1.00  & 20.00 & 19.00 & --- & --- & --- \\
\quad Depression RF & 25.01 & 39.70 & 14.69 & 7{,}808 & $<$.0001 &  0.750 \\
\quad Anxiety RF    & 30.66 & 33.20 & 2.54  & 23{,}849 & $<$.0001 &  0.237 \\
\quad Stress RF     & 28.22 & 35.16 & 6.94  & 17{,}598 & $<$.0001 &  0.437 \\
\bottomrule
\end{tabular}
\end{table}


\begin{table}[!ht]
\centering
\caption{Transfer of the random forest trained on Mistral Small DASS-21  data to human-transcribed conversations ($N = 115$; 63 clinical, 52 control). At the top, we present the separation of predicted scores between clinically defined groups. At the bottom, the classification performance of a single-feature logistic regression on the predicted score, evaluated with leave-one-out cross-validation.}
\label{tab:human_dass21}

\begin{tabular}{l rrr rr r}
\toprule
\multicolumn{7}{l}{\textit{Predicted score by clinical group}} \\
\midrule
 & \multicolumn{3}{c}{Median Score} & \multicolumn{2}{c}{Mann--Whitney} & \\
\cmidrule(lr){2-4} \cmidrule(lr){5-6}
 & Low & High & $\Delta$ (H$-$L) & $U$ & $p$ & AUC \\
\midrule
Transfer & 21.32 & 32.40 & 11.08 & 720 & $<$\,.0001 & .780 \\
\bottomrule

\vspace{0.5em} \\

\toprule
\multicolumn{7}{l}{\textit{Classification (LOO-CV)}} \\
\midrule
 & Precision & Recall & F1 & $n$ & & \\
\midrule
High          & .70 & .73 & .71 & 63 & & \\
Low           & .65 & .62 & .63 & 52 & & \\
Macro avg.    & .68 & .67 & .67 &    & & \\
Weighted avg. & .68 & .68 & .68 &    & & \\
\midrule
Accuracy      & \multicolumn{4}{l}{.68} & & \\
\bottomrule
\end{tabular}
\end{table}

\paragraph{Summary: transfer generalises unevenly across constructs, genres, and text sources}
Taken together, these out-of-domain transfer results show that the predictive signal captured by the Random Forest models is not an artefact of the specific textual format used for training, but generalises, to varying degrees, across text genre, prompting condition, and text source. Within the LLM-generated diary entries, transfer is strongest for PHQ-9 and DASS-21 Depression, intermediate for SWLS and DASS-21 Stress, and weakest for DASS-21 Anxiety. This ranking mirrors the RF and SHAP findings above: Depression-related predictions relied heavily on emotion features (Sadness, Disgust) plausibly expressed similarly across questionnaire explanations and diaries, whereas Anxiety-related predictions relied more heavily on network features, which may be less directly reflected in short, free-form diary text. Moving from LLM-generated to transcribed human speech introduces a further, expected drop in performance, yet both models still separate the clinical from the control group above chance (AUC $= .695$ for PHQ-9 and $.780$ for DASS-21 Depression; accuracy = $.62$ and $.68$, respectively), with the DASS-21 Depression model generalising more robustly. Overall, synthetic training data captures enough of the linguistic signal present in authentic psychological language to support above-chance transfer to real clinical speech, particularly for depression-related content.

\FloatBarrier

\section{Discussion}
This study introduces NLP Psychometrics, a framework linking psychometric scores with language structure, emotional content and respondents' profiles. The framework is grounded in lexical psychometric questionnaires, i.e., questionnaires where respondents have to justify with a brief text their psychometric scoring. These questionnaires provide data for exploring the cognitive framework of the Deep Lexical Hypothesis \cite{cutler2023deep, carrillo2026llms, fatima2021dasentimental}, where language use can reflect psychological traces of those who produced it. In the absence of extensive human lexical psychometric questionnaires, we resorted to 9 LLMs, whose variance in responses was amplified via the cognitive digital shadow framework \cite{ardebili2026mapping, franchino2026digital}, i.e. a systematic personification of LLMs with sociodemographics, personality, and other individual characteristics (e.g., mental health status).

From our pioneering work with NLP Psychometrics, three findings stand out. First, language carried most of the recoverable psychometric signal: emotion and network features formed a predictive core of machine learning features across all tested scales (SWLS, PHQ-9 and DASS-21), whereas sociodemographics alone rarely explained meaningful variance. Second, the markers identified by SHAP scores were interpretable and construct-specific, ranging from family income and affect for life satisfaction to neuroticism and discourse topology for depression. Third, the machine learning "feature to psychometric score" mapping transferred, with reduced yet significant accuracy, to both out-of-genre LLM-generated diaries and to human clinically labelled data \citep{tao2023androids, borraccino2025modeling}, i.e., speech transcripts of clinically depressed patients and controls.

Interestingly, current results indicate that NLP Psychometrics can be used to infer psychometric scores from text even on occasions where users do not complete a psychometric questionnaire, e.g., on social media \cite{de2013predicting, harrigian2020models}. However, the quality of NLP text-to-psychometrics mappings varies across scales. Sociodemographic features predicted SWLS scores reasonably well, yet failed almost completely for depression, in both the PHQ-9 and the DASS-21. This suggests that LLMs anchor simulated life satisfaction partly in persona details, most notably family income. In contrast, simulated depression is expressed almost entirely through language itself, as captured by forma mentis networks and emotional profiling. Interestingly, this division of labour mirrors human evidence. Income is a robust, if bounded, correlate of life evaluation \citep{diener2002will, kahneman2010high, jebb2018happiness}, and the SWLS explicitly measures a cognitive judgement about one's circumstances \citep{diener1985satisfaction}. Depression, in contrast, is better detected in how people write and speak, e.g., through negative affect, absolutist terms and self-focused style, than in who they are demographically \citep{rude2004language, al2018absolute, eichstaedt2018facebook}. The cognitive digital shadows examined here thus reproduce a distinction between demographically grounded and linguistically grounded constructs that is well documented in human samples \citep{boyd2017language, diener2002will, rude2004language}.

A second result concerns the value of network structure as interpretable, predictive features of text-to-psychometrics mappings. For depression, three network features converged: the number of nodes, capturing lexical diversity; the number of edges, capturing syntactic complexity; and degree assortativity, capturing how hubs connect to peripheral concepts \citep{stella2020forma, stella2019forma}. Higher predicted depression corresponded to larger, denser networks with lower assortativity, that is, star-like structures in which a few central concepts are specified by constellations of syntactic associates. Such discourse revolves around a small set of ideas, elaborated at length without repeating the same words. We read this convergence as a topological signature of rumination, the repetitive, self-focused thinking strongly associated with major depression \citep{nolen2008rethinking, american2013diagnostic}. NLP Psychometrics thus suggests that a hallmark of depressive cognition may be reflected in network topology, in generated as well as authentic language.

The DASS-21 subscales sharpen this picture. Depression, anxiety and stress shared a predictive core of neuroticism and network structure, in line with the transdiagnostic role of negative emotionality \citep{lovibond1995structure, kotov2010linking}. Yet degree assortativity worked in opposite directions across constructs. Generated personas with higher depression scores produced hub-centred, ruminative structures, whilst more anxious ones produced more integrated and distributed networks, touching many different topics. This opposition echoes the psychological distinction between rumination and worry: rumination dwells repetitively on a narrow set of mostly past-focused concerns, whereas worry ranges across many anticipated threats \citep{borkovec1998worry, watkins2008constructive}. LLMs reorganise discourse topology in construct-specific, and even opposite, directions. This indicates that assigned psychometric profiles shape not only what these models say but how their discourse is structured, with depression and anxiety pulling network topology in opposite directions despite sharing a common core of neuroticism and connectivity.             

These results carry implications for computer science. NLP Psychometrics operates as an explainable AI methodology for auditing how LLMs build psychometric scores out of personifications: whether LLMs rely on autobiographical persona details, on emotional content, or on the structural organisation of text \citep{ardebili2026mapping, franchino2026digital}. Feature ablation combined with SHAP makes this attribution explicit and reveals stark differences between models, such as the near-complete failure of the abliterated GPT-OSS variant, whose generated texts carried little psychometric signal. This positions NLP Psychometrics within the growing effort to study LLMs with psychological instruments \citep{hagendorff2023machine, serapio2023personality}. Importantly, interpretability is here achieved by design rather than post hoc, following calls to prefer glass-box models in high-stakes domains such as mental health \citep{rudin2019stop, salih2025perspective}.

As mentioned above, for psychology, NLP Psychometrics opens up novel opportunities for extracting psychometric scores even when not available, e.g., from social media. Our transfer results are the most consequential. Models trained purely on synthetic questionnaire explanations separated high- from low-scoring diaries and classified clinically depressed speakers above chance. At least part of the language-score mapping learned from LLMs therefore corresponds to genuine markers present in human speech. This supports the agenda of text psychometrics \citep{low2024speech, lowtext}: language can serve as psychometric evidence when the text-construct mapping is tested, interpreted and stress-tested across contexts. At the same time, transfer was uneven. Anxiety travelled poorly to diaries, and several network features reversed sign across genres, consistent with the known fragility of mental-health language models under domain shift \citep{harrigian2020models}. Genre, and not only construct, shapes the linguistic trace, and any deployment of NLP Psychometrics must therefore validate features register by register.

\subsection{Limitations}\label{sec:limitations}

The main limitation of this work is that the machine-learning pipeline was trained on LLM-generated texts and only subsequently deployed on human data. We adopted this design in the absence, to the best of our knowledge, of datasets pairing psychometric questionnaires with item-level linguistic explanations from human respondents. As a consequence, NLP Psychometrics should currently be framed as an auditing and exploration tool, not as a psychometric measure validated on human populations. Relatedly, the human evaluation relied on a single Italian corpus of 115 transcribed speakers with binary clinical labels rather than questionnaire scores.

Further limitations concern the features and the simulated personas. Several textual forma mentis network features scale with raw text length rather than purely with discourse structure. $N$ nodes, a proxy for lexical diversity \citep{stella2020text}, and $N$ edges, a measure of syntactic complexity \citep{stella2020text}, both grow mechanically with longer texts, independently of how those texts are organised. $N$ components, which indexes discourse fragmentation \citep{carrillo2025textual}, is similarly sensitive to length, since longer texts have more opportunity to form disconnected clusters. By contrast, degree and valence assortativity are Pearson correlations computed over existing edges \citep{rossetti2026ysocial, stella2019forma} and are therefore largely normalised against network size. The rumination signature reported here, which combines both size-sensitive ($N$ nodes, $N$ edges) and size-independent (degree assortativity) features, thus requires explicit verbosity controls before clinical interpretation \citep{al2018absolute}. Persona attributes were simplified, with binary gender, coarse categorical levels, and prevalence weights drawn from pandemic-era Italian surveys \citep{rossi2020covid, bonati2021psychological}. Finally, LLM questionnaire responses are known to be sensitive to prompt formulation and to display lower variance than human respondents \citep{hu2024quantifying, wenger2026large, wang2025large}, and all textual analyses were conducted in Italian with a single emotion lexicon \citep{mohammad2013crowdsourcing, semeraro2025emoatlas}, which bounds the generality of our findings.

\subsection{Future directions}

The natural next step is empirical: collecting human datasets that merge psychometric experiments with linguistic explanations of individual item scores, following the NLP Psychometrics protocol introduced here. Such data would allow the explainable pipeline to be trained and validated end-to-end on human language, turning the present audit into a proper psychometric validation. Methodologically, future work should implement length-controlled and content-matched text generation to isolate structural markers such as assortativity from verbosity, extend the framework to further constructs, languages and larger model populations. It should also focus on how reasoning regimes and safety alignment shape the linguistic expression of psychological profiles. Longitudinal designs, in which diary-like entries are collected over time (cf. \cite{carrillo2026llms}), could test whether NLP Psychometrics tracks within-person change rather than between-persona differences. Any application to clinical screening should proceed under human oversight, with NLP Psychometrics supporting, and never replacing, validated assessment \citep{guo2024large,taylor2025users}.

\section{Conclusion}

NLP Psychometrics combines validated psychometric instruments, cognitive network science and explainable machine learning into a single framework for studying how psychological constructs become expressed in language. Across 9 LLMs and 3 scales, psychometric scores proved recoverable from generated text through interpretable emotion and network features, whose signatures matched well-established human patterns, from the income-satisfaction link to ruminative discourse in depression. These mappings partially transferred to unconstrained diaries and to authentic clinical speech. Taken together, these results establish cognitive digital shadows as controlled probes of machine psychology, and lay the groundwork for a language-extended psychometrics that remains transparent about what it measures and how.

\clearpage

\bibliographystyle{unsrtnat}
\bibliography{references}

\clearpage



\appendix

\section{Supplementary Results: Random Forest} \label{app:rf_results}
\renewcommand{\thetable}{A.\arabic{table}}
\setcounter{table}{0}

\begin{table}[!ht]
\centering
\scriptsize
\caption{Random Forest performance on SWLS scores across the nine feature-set configurations, for the remaining LLMs. $N$ is the number of features used; $\rho_s$ is the Spearman correlation between predicted and true scores.}
\label{tab: rf_no_shap_llms_swls}
\begin{tabular}{ll|ccccc|ccccc}
\toprule
\multicolumn{2}{c}{} & \multicolumn{5}{c}{GPT Oss Uncensored} & \multicolumn{5}{c}{GPT Oss 20B} \\
\cmidrule(lr){3-7} \cmidrule(lr){8-12}
Feature Set & $N$ & MSE & RMSE & MAE & $R^2$ & $\rho_s$ & MSE & RMSE & MAE & $R^2$ & $\rho_s$ \\
\midrule
Sociodemographics & 5 & 3.053 & 1.747 & 1.387 & 0.050 & 0.243*** & 29.560 & 5.437 & 4.408 & 0.219 & 0.501*** \\
Big5 & 5 & 3.162 & 1.778 & 1.385 & 0.016 & 0.223*** & 35.301 & 5.941 & 4.946 & 0.067 & 0.290*** \\
Network  & 8 & 3.200 & 1.789 & 1.400 & 0.005 & 0.075*** & 36.925 & 6.077 & 5.127 & 0.025 & 0.148*** \\
Emotion  & 8 & 3.259 & 1.805 & 1.403 & -0.014 & 0.033 & 20.133 & 4.487 & 3.579 & 0.468 & 0.690*** \\
Sociodemographics And Big5 & 10 & 2.598 & 1.612 & 1.271 & 0.192 & 0.418*** & 24.240 & 4.923 & 3.998 & 0.360 & 0.620*** \\
Sociodemographics And Network & 13 & 2.913 & 1.707 & 1.345 & 0.094 & 0.287*** & 27.605 & 5.254 & 4.331 & 0.271 & 0.542*** \\
Sociodemographics And Emotions & 13 & 2.917 & 1.708 & 1.341 & 0.093 & 0.291*** & 17.545 & 4.189 & 3.275 & 0.537 & 0.749*** \\
Network And Emotions & 16 & 3.206 & 1.791 & 1.398 & 0.003 & 0.084*** & 19.690 & 4.437 & 3.556 & 0.480 & 0.700*** \\
Full Model  & 26 & 2.665 & 1.632 & 1.285 & 0.171 & 0.405*** & 16.752 & 4.093 & 3.221 & 0.557 & 0.766*** \\
\bottomrule
\end{tabular}

\vspace{1em}

\begin{tabular}{ll|ccccc|ccccc}
\toprule
\multicolumn{2}{c}{} & \multicolumn{5}{c}{Anita Uncensored} & \multicolumn{5}{c}{Granite 3 8B} \\
\cmidrule(lr){3-7} \cmidrule(lr){8-12}
Feature Set & $N$ & MSE & RMSE & MAE & $R^2$ & $\rho_s$ & MSE & RMSE & MAE & $R^2$ & $\rho_s$ \\
\midrule
Sociodemographics & 5 & 22.744 & 4.769 & 3.910 & 0.107 & 0.345*** & 26.084 & 5.107 & 4.142 & 0.214 & 0.477*** \\
Big5 & 5 & 20.761 & 4.556 & 3.730 & 0.184 & 0.429*** & 32.938 & 5.739 & 4.700 & 0.007 & 0.189*** \\
Network  & 8 & 24.059 & 4.905 & 4.017 & 0.055 & 0.221*** & 33.067 & 5.750 & 4.746 & 0.003 & 0.095*** \\
Emotion  & 8 & 15.048 & 3.879 & 3.102 & 0.409 & 0.638*** & 18.565 & 4.309 & 3.431 & 0.440 & 0.664*** \\
Sociodemographics And Big5 & 10 & 15.984 & 3.998 & 3.250 & 0.372 & 0.601*** & 21.415 & 4.628 & 3.799 & 0.355 & 0.594*** \\
Sociodemographics And Network & 13 & 20.656 & 4.545 & 3.728 & 0.189 & 0.433*** & 23.573 & 4.855 & 3.950 & 0.290 & 0.543*** \\
Sociodemographics And Emotions & 13 & 14.038 & 3.747 & 2.987 & 0.449 & 0.669*** & 16.153 & 4.019 & 3.191 & 0.513 & 0.721*** \\
Network And Emotions & 16 & 14.320 & 3.784 & 3.021 & 0.437 & 0.662*** & 18.511 & 4.302 & 3.431 & 0.442 & 0.665*** \\
Full Model  & 26 & 11.830 & 3.440 & 2.734 & 0.535 & 0.739*** & 15.683 & 3.960 & 3.172 & 0.527 & 0.732*** \\
\bottomrule
\end{tabular}

\vspace{1em}

\begin{tabular}{ll|ccccc}
\toprule
\multicolumn{2}{c}{} & \multicolumn{5}{c}{Nemotron 3 Nano 30B} \\
\cmidrule(lr){3-7}
Feature Set & $N$ & MSE & RMSE & MAE & $R^2$ & $\rho_s$ \\
\midrule
Sociodemographics & 5 & 35.539 & 5.961 & 4.721 & 0.290 & 0.526*** \\
Big5 & 5 & 45.501 & 6.745 & 5.487 & 0.091 & 0.325*** \\
Network  & 8 & 49.953 & 7.068 & 5.935 & 0.002 & 0.108*** \\
Emotion  & 8 & 30.967 & 5.565 & 4.417 & 0.381 & 0.568*** \\
Sociodemographics And Big5 & 10 & 25.256 & 5.026 & 3.949 & 0.495 & 0.678*** \\
Sociodemographics And Network & 13 & 33.937 & 5.826 & 4.731 & 0.322 & 0.549*** \\
Sociodemographics And Emotions & 13 & 22.363 & 4.729 & 3.753 & 0.553 & 0.688*** \\
Network And Emotions & 16 & 30.878 & 5.557 & 4.435 & 0.383 & 0.570*** \\
Full Model  & 26 & 20.770 & 4.557 & 3.671 & 0.585 & 0.722*** \\
\bottomrule
\end{tabular}
\caption*{\footnotesize *** $p<.001$, ** $p<.01$, * $p<.05$}
\end{table}

\begin{table}[!ht]
\centering
\scriptsize
\caption{andom Forest performance on PHQ scores across the nine feature-set configurations, for the remaining LLMs. $N$ is the number of features used; $\rho_s$ is the Spearman correlation between predicted and true scores.}
\label{tab: rf_no_shap_llms_phq}
\begin{tabular}{ll|ccccc|ccccc}
\toprule
\multicolumn{2}{c}{} & \multicolumn{5}{c}{GPT Oss Uncensored} & \multicolumn{5}{c}{GPT Oss 20B} \\
\cmidrule(lr){3-7} \cmidrule(lr){8-12}
Feature Set & $N$ & MSE & RMSE & MAE & $R^2$ & $\rho_s$ & MSE & RMSE & MAE & $R^2$ & $\rho_s$ \\
\midrule
Sociodemographics & 5 & 5.058 & 2.249 & 1.816 & -0.057 & 0.039 & 49.850 & 7.060 & 6.024 & -0.054 & 0.029 \\
Big5 & 5 & 4.810 & 2.193 & 1.757 & -0.005 & 0.177*** & 40.809 & 6.388 & 5.295 & 0.137 & 0.390*** \\
Network  & 8 & 4.820 & 2.195 & 1.790 & -0.007 & 0.061** & 38.252 & 6.185 & 5.203 & 0.191 & 0.419*** \\
Emotion  & 8 & 4.778 & 2.186 & 1.781 & 0.001 & 0.072*** & 36.787 & 6.065 & 5.017 & 0.222 & 0.457*** \\
Sociodemographics And Big5 & 10 & 4.528 & 2.128 & 1.710 & 0.053 & 0.236*** & 38.258 & 6.185 & 5.186 & 0.191 & 0.430*** \\
Sociodemographics And Network & 13 & 4.746 & 2.179 & 1.777 & 0.008 & 0.107*** & 37.480 & 6.122 & 5.168 & 0.207 & 0.437*** \\
Sociodemographics And Emotions & 13 & 4.710 & 2.170 & 1.765 & 0.016 & 0.116*** & 36.648 & 6.054 & 5.037 & 0.225 & 0.456*** \\
Network And Emotions & 16 & 4.765 & 2.183 & 1.775 & 0.004 & 0.089*** & 29.987 & 5.476 & 4.529 & 0.366 & 0.583*** \\
Full Model  & 26 & 4.488 & 2.119 & 1.718 & 0.062 & 0.249*** & 26.209 & 5.119 & 4.213 & 0.446 & 0.664*** \\
\bottomrule
\end{tabular}

\vspace{1em}

\begin{tabular}{ll|ccccc|ccccc}
\toprule
\multicolumn{2}{c}{} & \multicolumn{5}{c}{Anita Uncensored} & \multicolumn{5}{c}{Qwen 4B Thinking} \\
\cmidrule(lr){3-7} \cmidrule(lr){8-12}
Feature Set & $N$ & MSE & RMSE & MAE & $R^2$ & $\rho_s$ & MSE & RMSE & MAE & $R^2$ & $\rho_s$ \\
\midrule
Sociodemographics & 5 & 28.677 & 5.355 & 4.407 & -0.052 & 0.058** & 65.559 & 8.097 & 7.006 & -0.053 & 0.054** \\
Big5 & 5 & 19.498 & 4.416 & 3.673 & 0.285 & 0.535*** & 49.750 & 7.053 & 5.817 & 0.201 & 0.415*** \\
Network  & 8 & 25.329 & 5.033 & 4.099 & 0.071 & 0.214*** & 59.260 & 7.698 & 6.662 & 0.048 & 0.215*** \\
Emotion  & 8 & 20.267 & 4.502 & 3.602 & 0.257 & 0.440*** & 53.611 & 7.322 & 6.249 & 0.139 & 0.274*** \\
Sociodemographics And Big5 & 10 & 18.303 & 4.278 & 3.593 & 0.329 & 0.560*** & 47.680 & 6.905 & 5.779 & 0.234 & 0.431*** \\
Sociodemographics And Network & 13 & 25.066 & 5.007 & 4.098 & 0.081 & 0.228*** & 58.474 & 7.647 & 6.638 & 0.061 & 0.229*** \\
Sociodemographics And Emotions & 13 & 20.226 & 4.497 & 3.610 & 0.258 & 0.447*** & 52.583 & 7.251 & 6.231 & 0.155 & 0.291*** \\
Network And Emotions & 16 & 18.814 & 4.337 & 3.494 & 0.310 & 0.486*** & 51.163 & 7.153 & 6.093 & 0.178 & 0.345*** \\
Full Model  & 26 & 13.746 & 3.708 & 3.024 & 0.496 & 0.673*** & 39.745 & 6.304 & 5.270 & 0.361 & 0.548*** \\
\bottomrule
\end{tabular}

\vspace{1em}

\begin{tabular}{ll|ccccc}
\toprule
\multicolumn{2}{c}{} & \multicolumn{5}{c}{Granite 3 8B} \\
\cmidrule(lr){3-7}
Feature Set & $N$ & MSE & RMSE & MAE & $R^2$ & $\rho_s$ \\
\midrule
Sociodemographics & 5 & 33.616 & 5.798 & 4.745 & -0.011 & 0.137*** \\
Big5 & 5 & 27.539 & 5.248 & 4.231 & 0.172 & 0.425*** \\
Network  & 8 & 29.645 & 5.445 & 4.430 & 0.109 & 0.322*** \\
Emotion  & 8 & 28.716 & 5.359 & 4.370 & 0.137 & 0.347*** \\
Sociodemographics And Big5 & 10 & 25.433 & 5.043 & 4.084 & 0.235 & 0.483*** \\
Sociodemographics And Network & 13 & 28.868 & 5.373 & 4.372 & 0.132 & 0.357*** \\
Sociodemographics And Emotions & 13 & 28.273 & 5.317 & 4.350 & 0.150 & 0.364*** \\
Network And Emotions & 16 & 26.257 & 5.124 & 4.177 & 0.210 & 0.434*** \\
Full Model  & 26 & 21.432 & 4.629 & 3.741 & 0.356 & 0.597*** \\
\bottomrule
\end{tabular}
\caption*{\footnotesize *** $p<.001$, ** $p<.01$, * $p<.05$}
\end{table}

\FloatBarrier

\section{Supplementary Results: Qwen-4B-Instruct Diary Transfer}\label{app:diaries_results}
\renewcommand{\thetable}{B.\arabic{table}}
\setcounter{table}{0}

As reported in Section~\ref{sec:diaries_methods}, we initially tested diary transfer using Qwen-4B-Instruct, the overall best-performing LLM in the Random Forest analysis (Section~\ref{sec:rf_results}). Despite its strong in-domain performance, transfer to diary entries proved inconsistent, motivating our decision to report Mistral Small in the main text.

For SWLS, transfer under the reference diary-prompting condition (100--150 words, matching the setup used for Mistral Small) was successful: the model recovered a significant separation between the low- and high-scoring groups in the expected direction (median predicted scores of 17.40 vs.\ 19.56, $\Delta = 2.16$; $U = 16{,}531$, $p < .0001$, $r = +0.471$; Table~\ref{tab:swls_original}).
    
For PHQ-9, in contrast, transfer under the same condition failed to reach significance: predicted medians for the low- and high-scoring groups were nearly indistinguishable and in the wrong direction (10.65 vs.\ 9.50, $p = .15$; Table~\ref{tab:phq_original}). We tested two additional prompt variants --- one with extreme injected scores (0/3 on PHQ-9 items) and one constraining the diary text to more closely resemble the surface form of questionnaire explanations --- with mixed results (see Table~\ref{tab:phq_original}). Only the surface-form-matched variants recovered a significant separation, and with small effect sizes ($r = 0.156$ and $r = 0.340$).

Overall, Qwen-4B-Instruct's transfer performance is markedly more sensitive to the diary-prompting condition than Mistral Small's, which is why the main text reports results for the latter.

\begin{table}[!ht]
\centering
\scriptsize
\caption{Transfer learning results for SWLS diary entries (Qwen-4B-Instruct, Original prompting condition, All Features model).}
\label{tab:swls_original}
\begin{tabular}{l rrrrrrl}
\toprule
 & \multicolumn{3}{c}{Median Score} & \multicolumn{2}{c}{Mann--Whitney} & \multicolumn{2}{c}{Effect Size} \\
\cmidrule(lr){2-4} \cmidrule(lr){5-6} \cmidrule(lr){7-8}
 & Low & High & $\Delta$ (H$-$L) & $U$ & $p$ & $r$ & \\
\midrule
Ground Truth & 6.00 & 34.00 & 28.00 & --- & --- & --- \\
Transfer & 17.40 & 19.56 & 2.16 & 16,531 & $<$\,.0001 & +0.471 \\
\bottomrule
\end{tabular}
\end{table}

\begin{table}[!ht]
\centering
\scriptsize
\caption{Transfer learning results for PHQ-9 diary entries (Qwen-4B-Instruct, All Features model), under the Original, Extreme scores, and Sentences prompt variants.}
\label{tab:phq_original}
\begin{tabular}{l rrrrrrl}
\toprule
 & \multicolumn{3}{c}{Median Score} & \multicolumn{2}{c}{Mann--Whitney} & \multicolumn{2}{c}{Effect Size} \\
\cmidrule(lr){2-4} \cmidrule(lr){5-6} \cmidrule(lr){7-8}
 & Low & High & $\Delta$ (H$-$L) & $U$ & $p$ & $r$ & \\
\midrule
\multicolumn{7}{l}{\textit{Original condition}} \\
\quad Ground Truth & 2.00 & 25.00 & 23.00 & --- & --- & --- \\
\quad Transfer & 5.02 & 6.12 & 1.10 & 26,390 & .0026 & +0.156 \\
\midrule
\multicolumn{7}{l}{\textit{Extreme scores condition}} \\
\quad Ground Truth & 0.00 & 27.00 & 27.00 & --- & --- & --- \\
\quad Transfer & 10.95 & 10.56 & -0.39 & 30,959 & .9182 & +0.005 \\
\midrule
\multicolumn{7}{l}{\textit{Sentences (strict, 10$\times$18 words)}} \\
\quad Ground Truth & 2.00 & 25.00 & 23.00 & --- & --- & --- \\
\quad Transfer & 4.95 & 6.94 & 1.99 & 20,628 & $<$\,.0001 & +0.340 \\
\bottomrule
\end{tabular}
\end{table}

\FloatBarrier

\section{Supplementary Results: SHAP Feature Importance} \label{app:shap_results}
\renewcommand{\thefigure}{C.\arabic{figure}}
\setcounter{figure}{0}

\begin{figure}[!ht]
    \centering
    \begin{subfigure}[b]{0.3\textwidth}
        \centering
        \includegraphics[width=\textwidth]{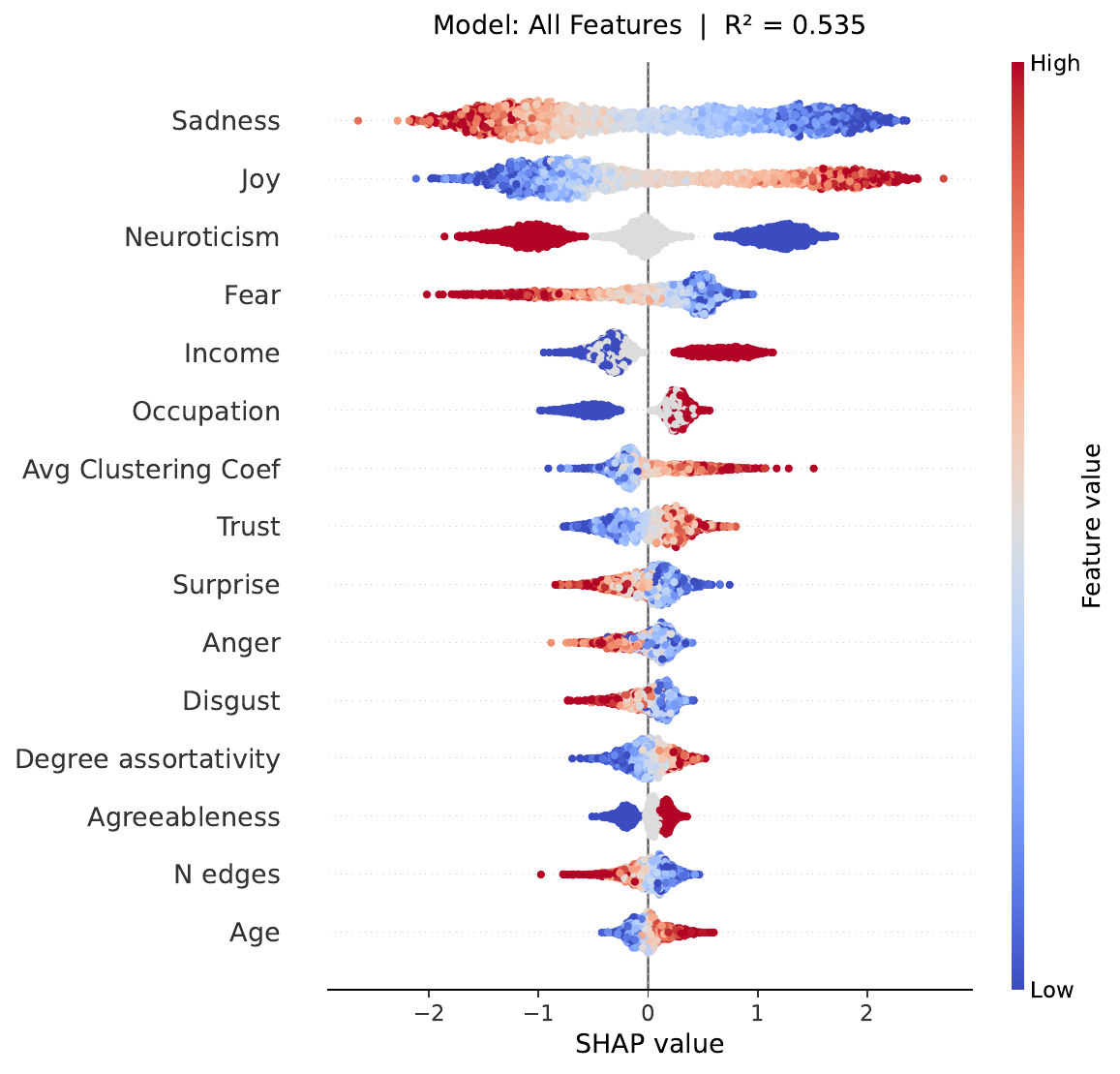}
        \caption{Anita Uncensored}
    \end{subfigure}
    \begin{subfigure}[b]{0.3\textwidth}
        \centering
        \includegraphics[width=\textwidth]{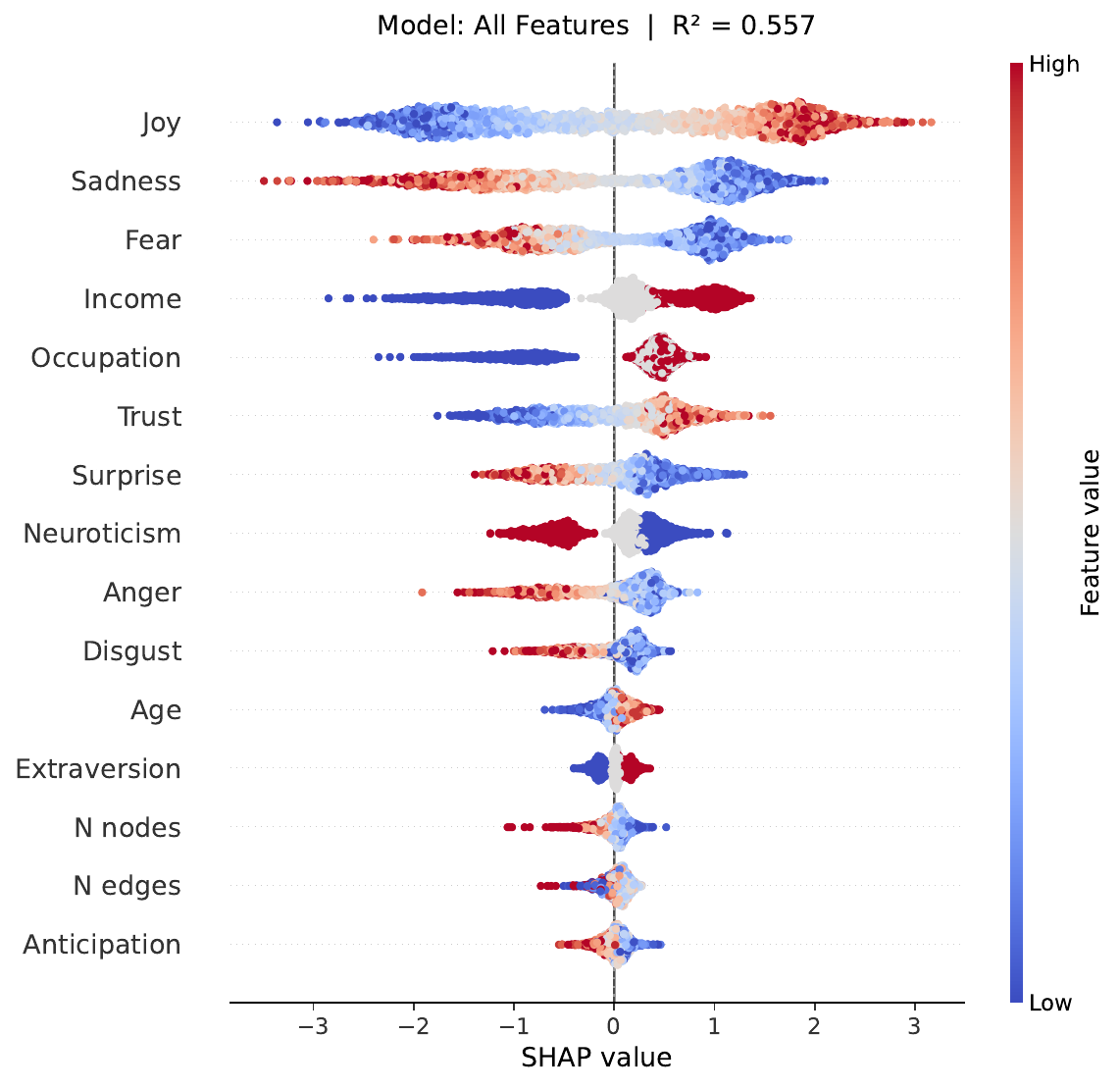}
        \caption{GPT-OSS-20B}
    \end{subfigure}   
    \begin{subfigure}[b]{0.3\textwidth}
        \centering
        \includegraphics[width=\textwidth]{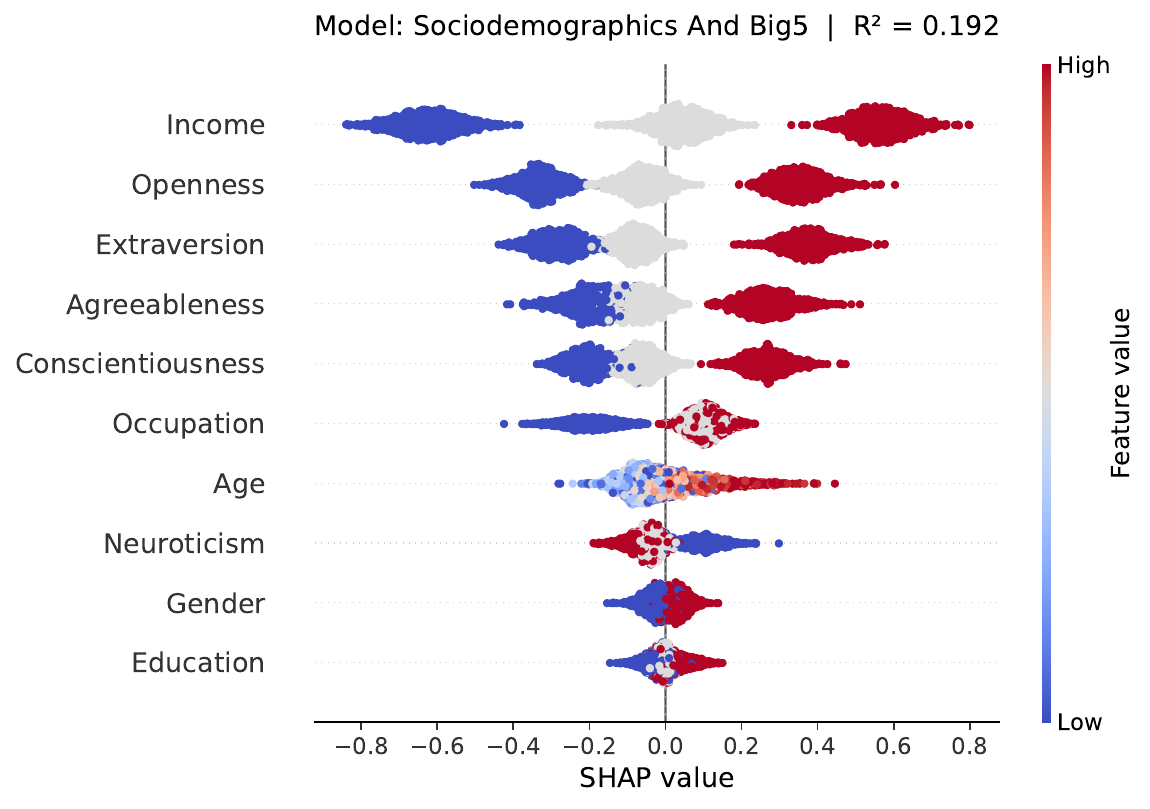} 
        \caption{GPT-OSS-Uncensored}
    \end{subfigure}
    \begin{subfigure}[b]{0.33\textwidth}
        \centering
        \includegraphics[width=\textwidth]{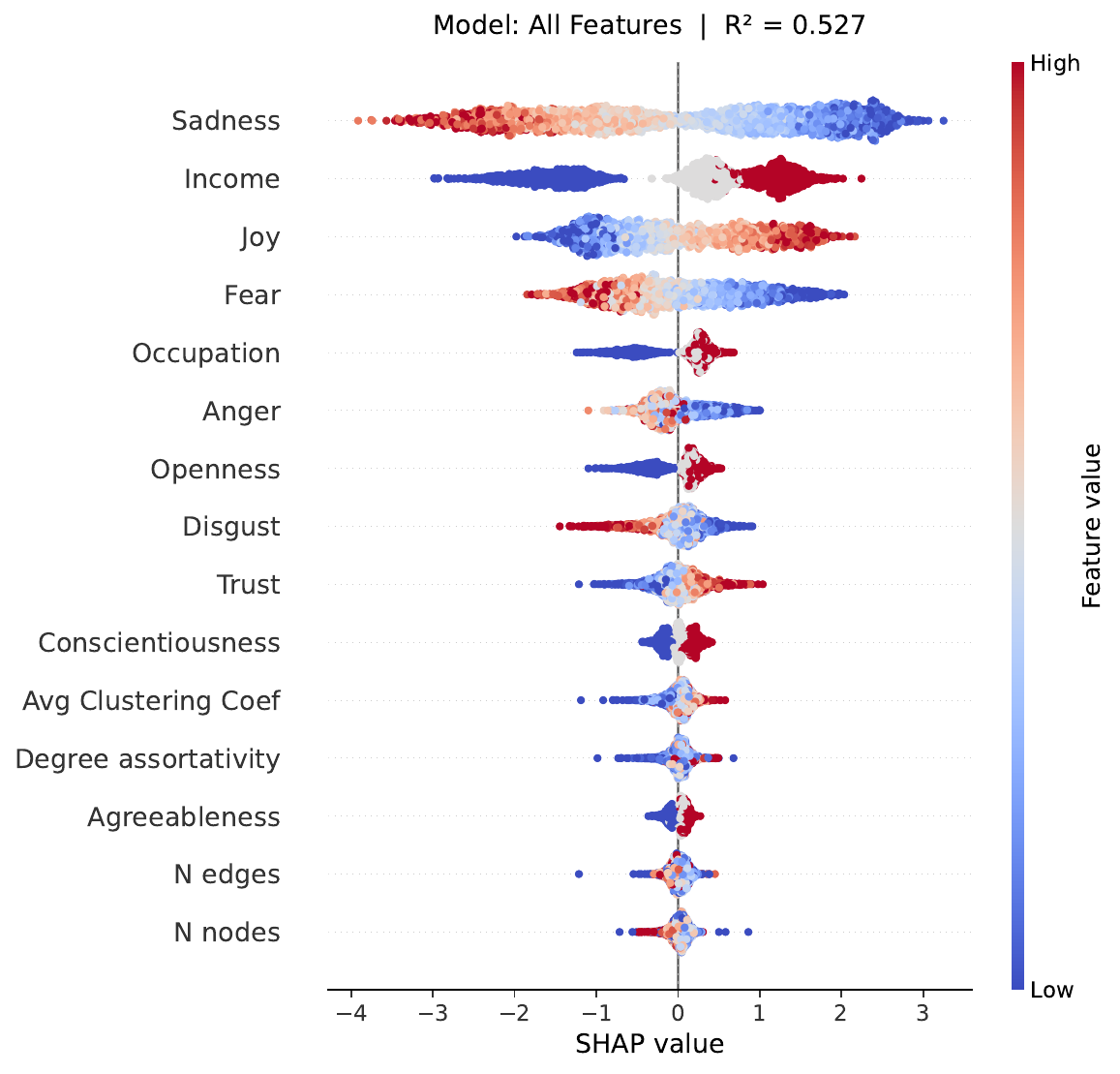}
        \caption{Granite-3-8B}
    \end{subfigure}
    \begin{subfigure}[b]{0.33\textwidth}
        \centering
        \includegraphics[width=\textwidth]{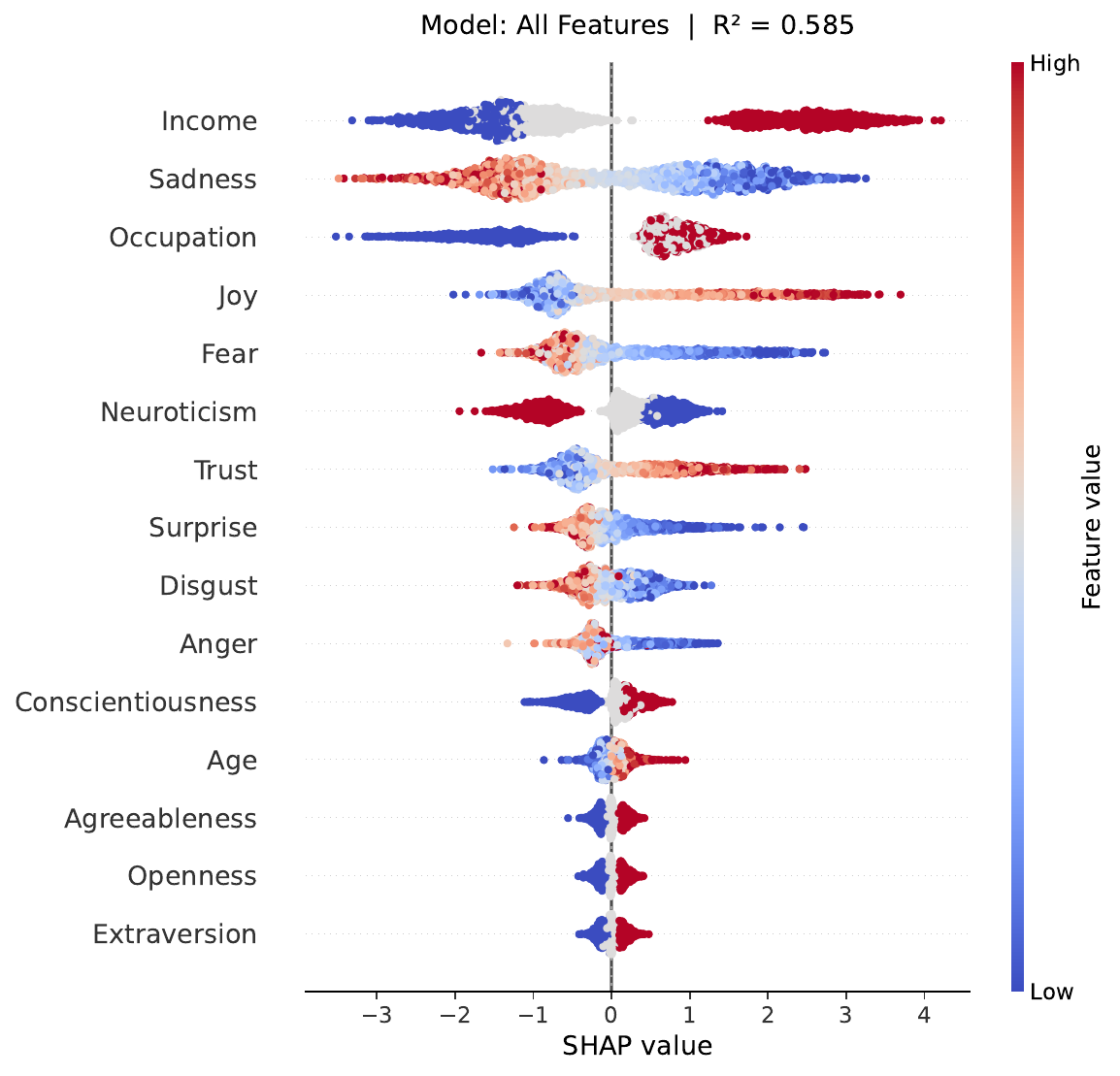}
        \caption{Nemotron-3-Nano-30B}
    \end{subfigure}
    
   \caption{SHAP summary plots for the remaining five LLMs on SWLS, each showing the best Random Forest model. Features are ranked by mean absolute SHAP value (top to bottom); point colour encodes the feature's original value (red = high, blue = low), and horizontal position indicates the SHAP value's effect on the predicted SWLS score.}
    \label{fig:app_shap_swls}
\end{figure}

\begin{figure}[!ht]
    \centering
    \begin{subfigure}[b]{0.3\textwidth}
        \includegraphics[width=\textwidth]{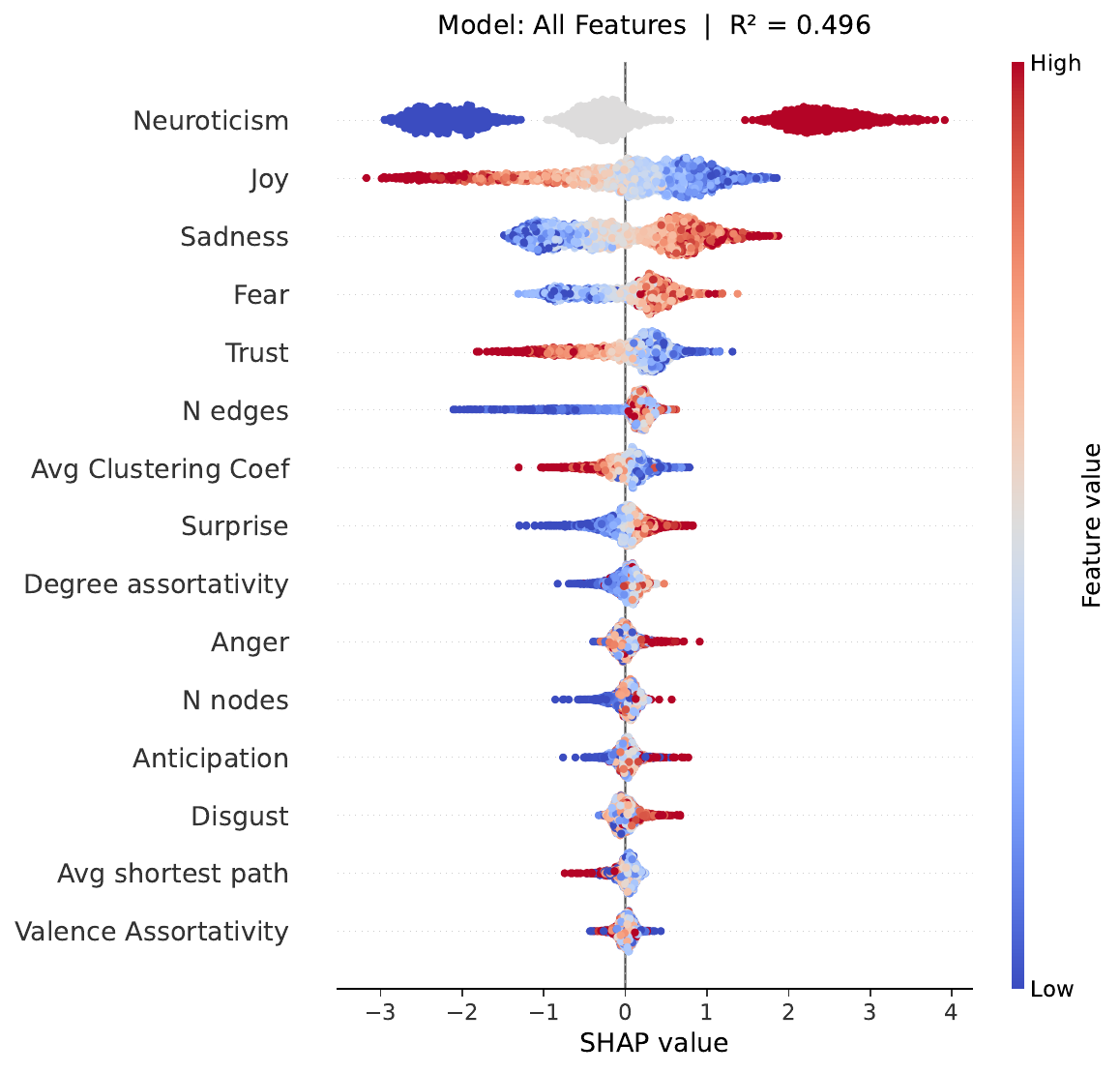}
        \caption{Anita Uncensored}
    \end{subfigure}
    \begin{subfigure}[b]{0.3\textwidth}
        \includegraphics[width=\textwidth]{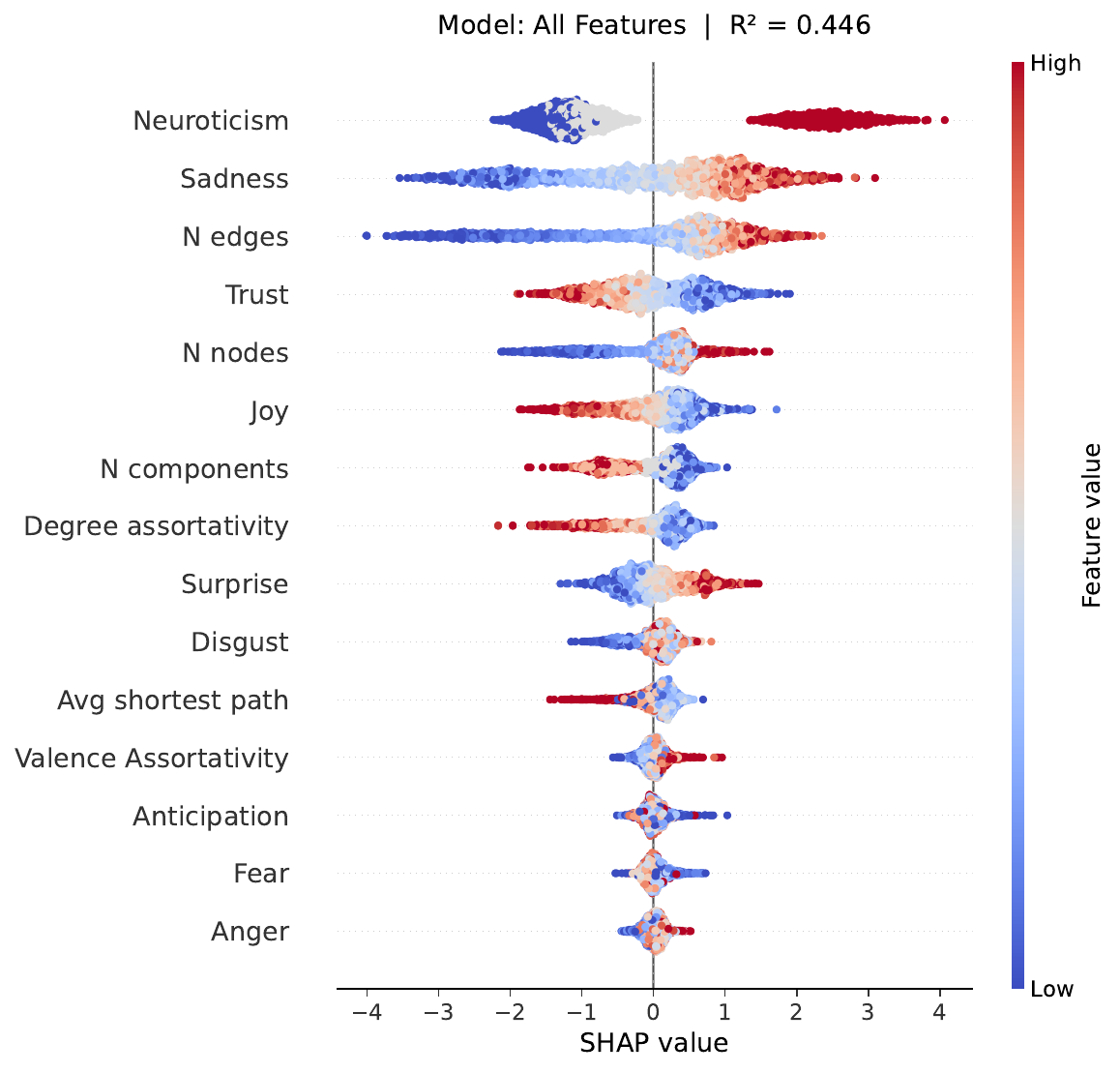}
        \caption{GPT-OSS-20B}
    \end{subfigure}
    \begin{subfigure}[b]{0.3\textwidth}
        \includegraphics[width=\textwidth]{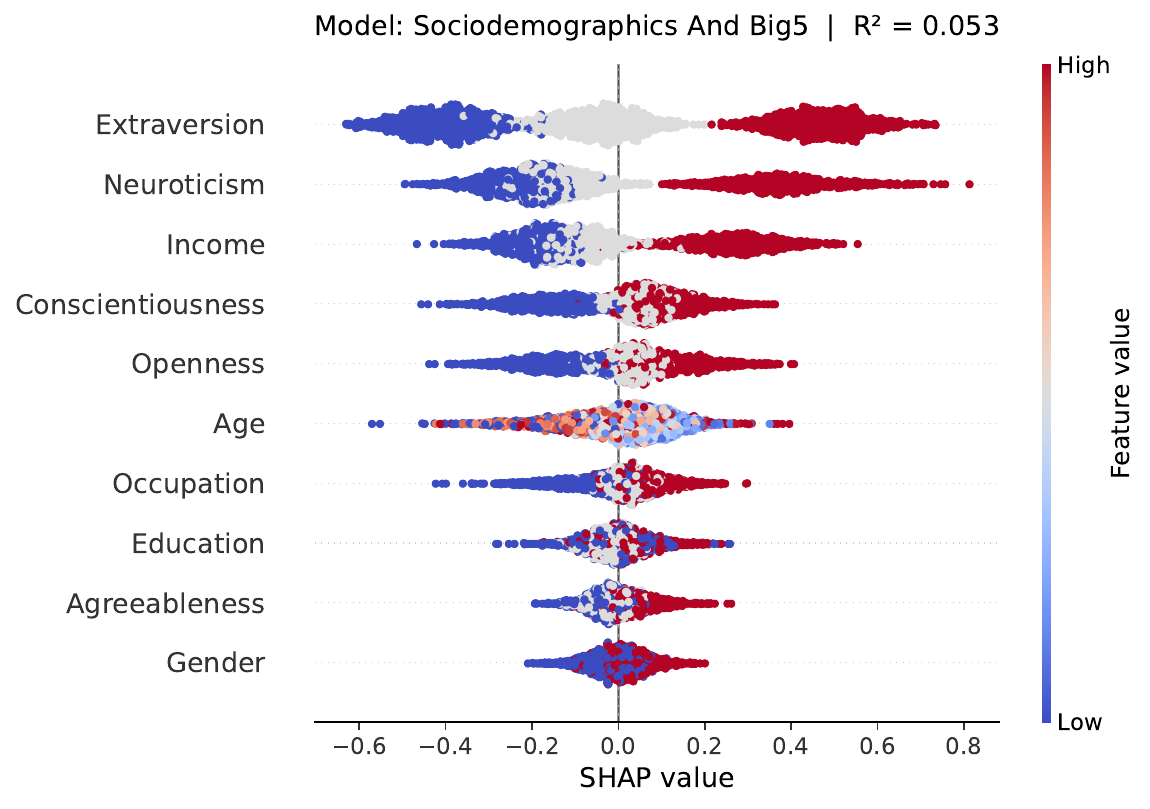} 
        \caption{GPT-OSS-Uncensored}
    \end{subfigure}
    \begin{subfigure}[b]{0.3\textwidth}
        \includegraphics[width=\textwidth]{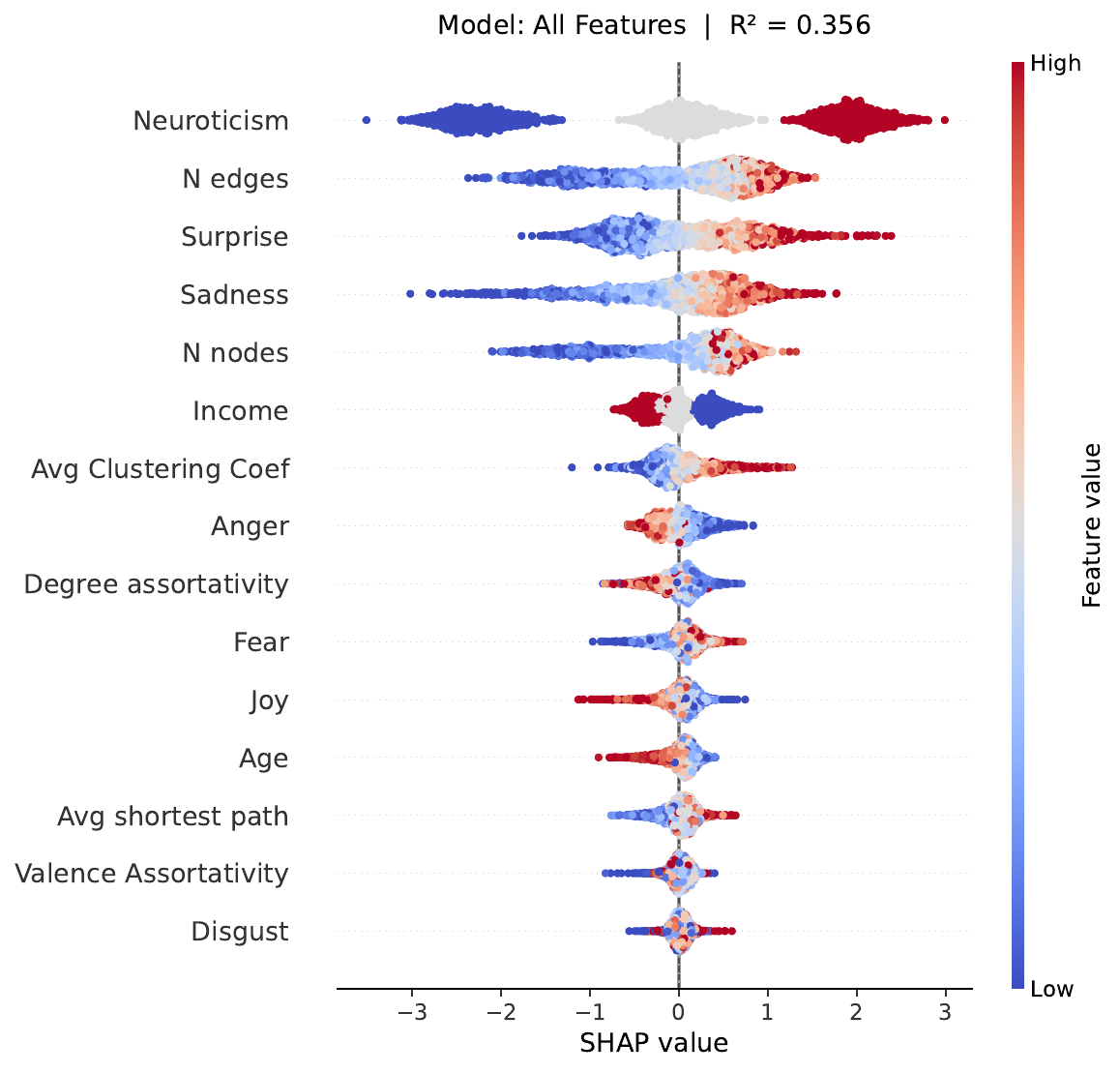}
        \caption{Granite-3-8B}
    \end{subfigure}
    \begin{subfigure}[b]{0.3\textwidth}
        \includegraphics[width=\textwidth]{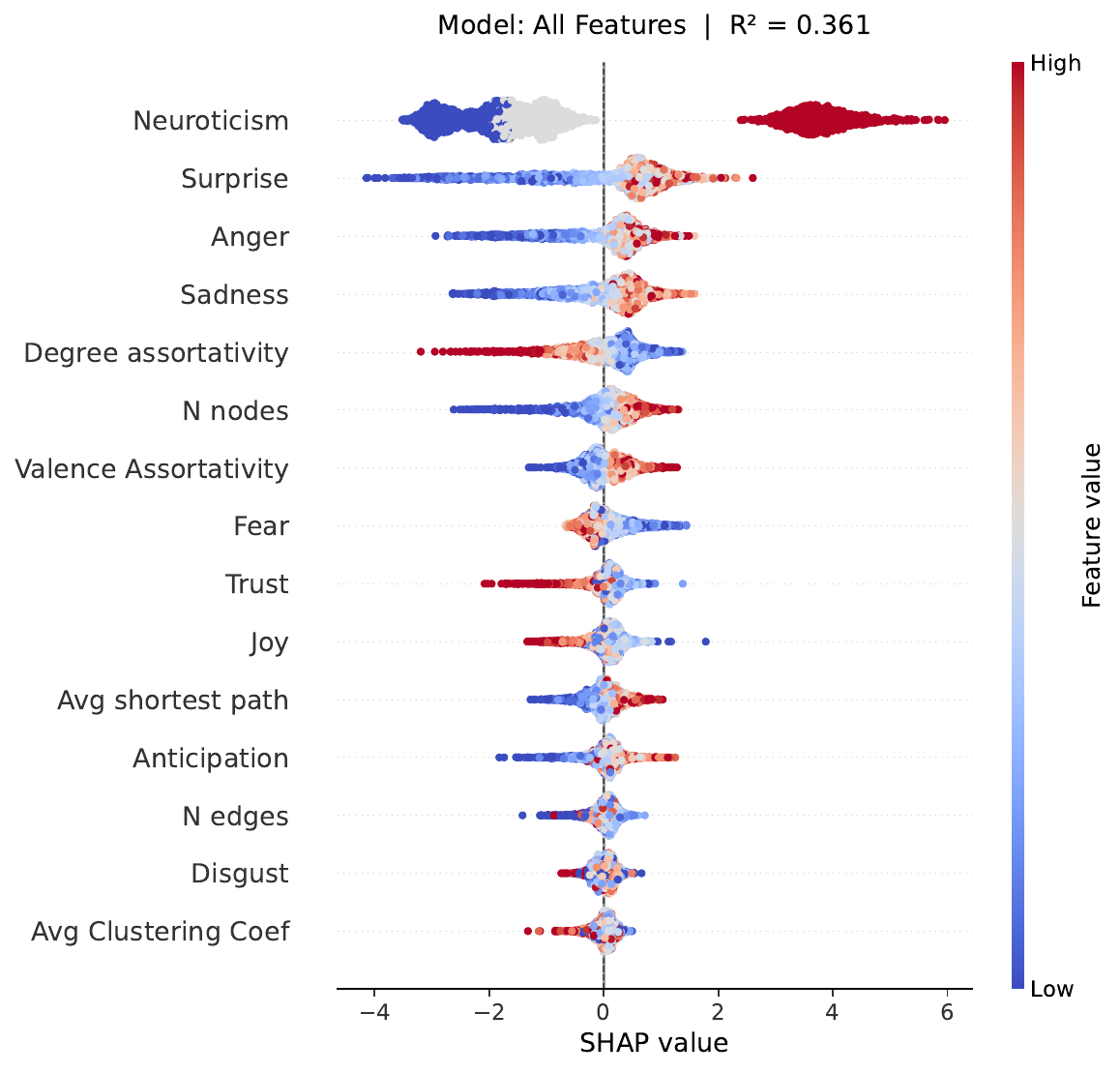}
        \caption{Qwen-4B-Thinking}
    \end{subfigure}
    
   \caption{SHAP summary plots for the remaining five LLMs on PHQ-9, each showing the best Random Forest model. Features are ranked by mean absolute SHAP value (top to bottom); point colour encodes the feature's original value (red = high, blue = low), and horizontal position indicates the SHAP value's effect on the predicted PHQ-9 score.}
    \label{fig:app_shap_phq}
\end{figure}

\end{document}